\documentclass{article}
\PassOptionsToPackage{numbers, compress}{natbib}
\usepackage[utf8]{inputenc} 
\usepackage[T1]{fontenc}    
\usepackage{url}            
\usepackage{booktabs}       
\usepackage{amsfonts}       
\usepackage{nicefrac}       
\usepackage{microtype}      
\usepackage[table]{xcolor}  
\usepackage{hyperref}       
\usepackage{graphicx}       
\usepackage{enumitem}       
\usepackage{xspace}         
\usepackage{caption}
\usepackage{subcaption}
\usepackage{multirow}
\usepackage{makecell}
\usepackage{amsmath}
\usepackage{pifont}
\usepackage[most]{tcolorbox}          
\tcbuselibrary{listings, skins, breakable}
\usepackage[capitalize]{cleveref}
\usepackage{algorithm}
\usepackage{algorithmicx}
\usepackage{algpseudocode}
\usepackage{adjustbox}          
\usepackage{mathtools}
\usepackage{fontawesome5}

\newcommand{\cmark}{\ding{51}}

\definecolor{table_ours}{HTML}{F5FFFA}   
\definecolor{titleblue}{HTML}{5AA1BF}     
\definecolor{lightblue}{HTML}{EAF4FB}     

\newcounter{promptcount}
\newtcolorbox{my_template_box}[2][]{
    enhanced,
    colback=lightblue, colframe=titleblue,
    colbacktitle=titleblue, coltitle=white,
    fonttitle=\bfseries\small,
    boxrule=1pt, arc=2mm, boxsep=4pt,
    left=6pt,right=6pt,top=4pt,bottom=4pt,
    code={\refstepcounter{promptcount}},
    title={#2}, #1
}

\newcommand{\myeg}{\textit{e.g.,}\xspace}
\newcommand{\mypara}[1]{\vspace{-0.2mm}\noindent\textbf{#1}\hspace{0.02cm}}

\newcommand{\logo}[1]{%
    \smash{%
        \raisebox{-\dp\strutbox}{%
            \includegraphics[height=\dimexpr\ht\strutbox+\dp\strutbox\relax]{figures/logo/org/#1}%
        }%
    }\,%
}

\AtEndPreamble{
    \crefname{section}{Sec.}{Secs.}         \Crefname{section}{Sec.}{Secs.}
    \crefname{equation}{Eq.}{Eqs.}          \Crefname{equation}{Eq.}{Eqs.}
    \crefname{table}{Tab.}{Tabs.}           \Crefname{table}{Tab.}{Tabs.}
    \crefname{figure}{Fig.}{Figs.}          \Crefname{figure}{Fig.}{Figs.}
    \crefname{promptcount}{prompt}{prompts} \Crefname{promptcount}{Prompt}{Prompts}
    \creflabelformat{equation}{#2(#1)#3}
    \creflabelformat{table}{#2#1#3}
    \creflabelformat{figure}{#2#1#3}

}

\let\titleold\title
\renewcommand{\title}[1]{\titleold{#1}\newcommand{\thetitle}{#1}}

\makeatletter
\newcommand{\safenolinenumbers}{\ifdefined\nolinenumbers\nolinenumbers\fi}
\newcommand{\safelinenumbers}{\ifdefined\linenumbers\linenumbers\fi}
\newcommand{\saferesetlinenumber}[1][]{\ifdefined\resetlinenumber\resetlinenumber[#1]\fi}
\makeatother

\def\maketitlesupplementary
{
    {
            \newpage
            \par\safenolinenumbers
            \centering
            \Large
            \textbf{\thetitle}\\
            \vspace{0.3em}
            Supplementary Material \\
            \safelinenumbers
        }
}

\newcommand{\nextline}{\\} 
\newcommand{\ul}[1]{\underline{#1}}
\newcommand{\sd}[2]{%
    \ensuremath{#1\mathrlap{_{\textcolor{red}{#2}}}}%
}

\usepackage[main, final]{neurips_2026}

\title{
    ChartArena: A Unified Benchmark with Atomic-Primitive Reasoning for Chart Parsing
}

\author{
    Shangpin Peng$^{1,\,4,\,5,\,\ast}$\qquad
    Gengluo Li$^{2,\,\ast}$\qquad
    Xingyu Wan$^{1}$\qquad
    Chengquan Zhang$^{1,\,\dagger}$
    \\[1pt]
    Hao Feng$^{1}$\qquad
    Binghong Wu$^{1}$\qquad
    Huawen Shen$^{1}$\qquad
    Weinong Wang$^{1}$\qquad
    Ziyi Cai$^{4}$
    \\[1pt]
    Zhuotao Tian$^{4,\,}$\textsuperscript{\scalebox{0.8}{\faEnvelope}}\qquad
    Han Hu$^{1}$\qquad
    Can Ma$^{2}$\qquad
    Yu Zhou$^{3,\,}$\textsuperscript{\scalebox{0.8}{\faEnvelope}}
    \\[5pt]
    $^{1}$Large Language Model Department, Tencent
    \\[1pt]
    $^{2}$Institute of Information Engineering, Chinese Academy of Sciences
    \\[1pt]
    $^{3}$Nankai University
    \qquad
    $^{4}$Shenzhen Loop Area Institute
    \\[1pt]
    $^{5}$Hong Kong University of Science and Technology
    \\[5pt]
    \centerline {
        \tt\small
        pspdada0808@gmail.com
        \quad
        zhuotaotian@slai.edu.cn
        \quad
        yzhou@nankai.edu.cn
    }\\[1pt]
    \centerline{
        Code:
        \raisebox{-0.15\height}{\includegraphics[height=12pt]{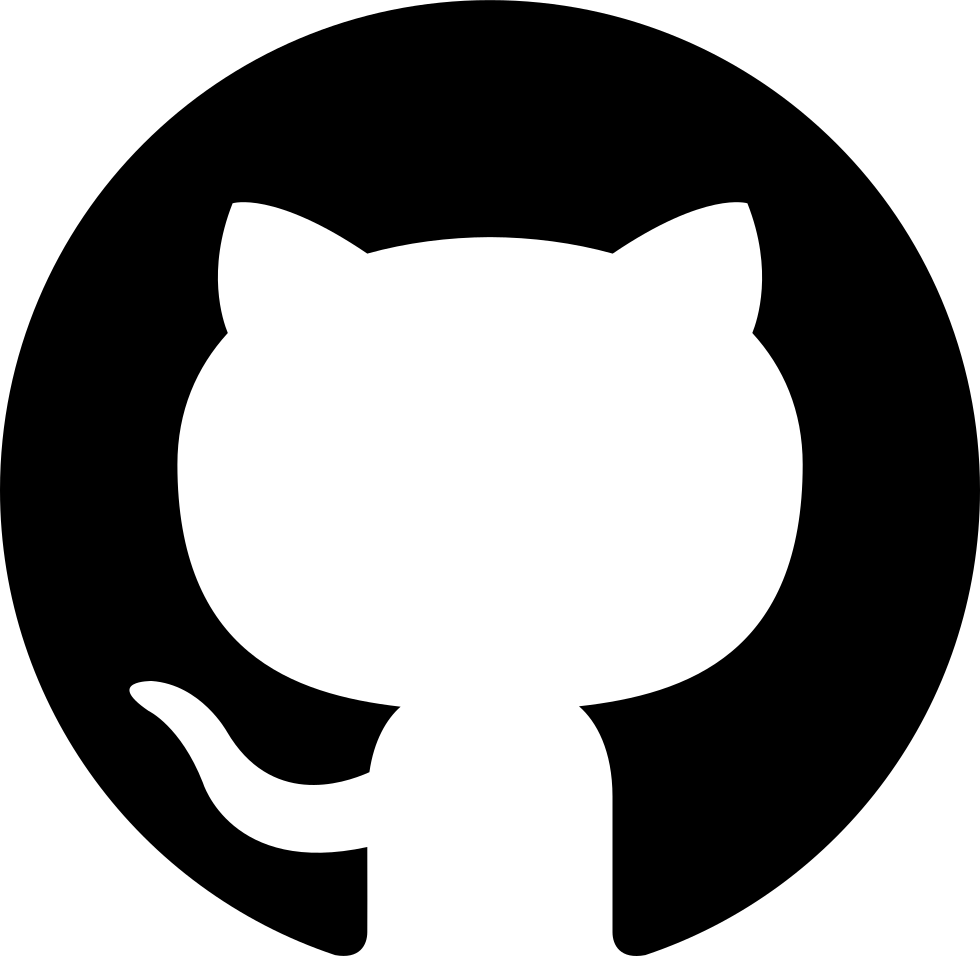}}
        \url{https://github.com/pspdada/ChartArena}
    }
}

\begin{document}

\maketitle

\makeatletter
\if@anonymous
\else
    \let\oldthefootnote\thefootnote
\let\thefootnote\relax\footnotetext{
    $^{\scalebox{1.0}{\hspace{-0.7em}$\ast$}}$Equal contribution.
    \hspace{1em}
    $^{\dagger}$Project leader.
    \hspace{1em}
    \textsuperscript{\scalebox{0.8}{\faEnvelope}}Corresponding author.
}
\let\thefootnote\oldthefootnote

\fi
\makeatother

\begin{abstract}
    Charts convey dense quantitative and relational information, yet current general chart parsing remains challenging due to fragmented evaluation and limited structural reasoning. Existing benchmarks focus on narrow sub-tasks with inconsistent output formats (e.g., Markdown, CSV, JSON, SVG, code), hindering fair comparison and often overlooking real-world scenarios such as printed or hand-drawn photos. Meanwhile, end-to-end multimodal models tend to learn shallow pixel-to-text mappings without explicit structure understanding, making them fragile under visual variations.
    To address these issues, we present a unified framework that connects evaluation and modeling. On the evaluation side, we introduce \textbf{ChartArena}, a comprehensive bilingual benchmark covering eight chart families, including both numeric charts (bar, line, pie, radar, box plot, combination) and diagrammatic structures (flowchart, mind map). The dataset is constructed via a human-agent collaborative annotation pipeline, where model-assisted drafts are refined through multi-stage human verification to ensure structural consistency and annotation reliability.
    Furthermore, we design a format-agnostic evaluation protocol that maps different outputs into two canonical semantic spaces: a normalized triple view and a directed graph view, and evaluates them with structure-aware metrics.
    On the modeling side, we propose \textbf{Chart Atomic Primitives (CAP)}, a lightweight semantic scaffold that introduces structural inductive bias during supervised fine-tuning. By guiding the model to infer structure before generating outputs, CAP leads to more consistent and robust parsing behavior. Extensive experiments show that models trained with CAP mechanism achieve reliable performance on ChartArena.
\end{abstract}
\section{Introduction}
\label{sec:introduction}

Charts serve as indispensable visual instruments for conveying quantitative and relational data. To unlock this information, chart parsing~\cite{ChartX_ChartVLM_2025, OneChart_2024} aims to convert these images into structured, machine-executable representations, serving as a foundation for downstream data analysis and reasoning. Recently, with the rise of multimodal large language models (MLLMs), the field has shifted from traditional modular pipelines~\cite{ChartSense_2017, Revision_2011} to end-to-end generation approaches~\cite{Qwen2_5_VL_2025, HunyuanOCR_2025}. Yet, despite this rapid progress, building a truly general chart parser remains challenging due to two fundamental issues: \textit{fragmented evaluation} and \textit{limited structural reasoning}.

\begin{figure}[t!]
    \centering
    \includegraphics[width=\linewidth]{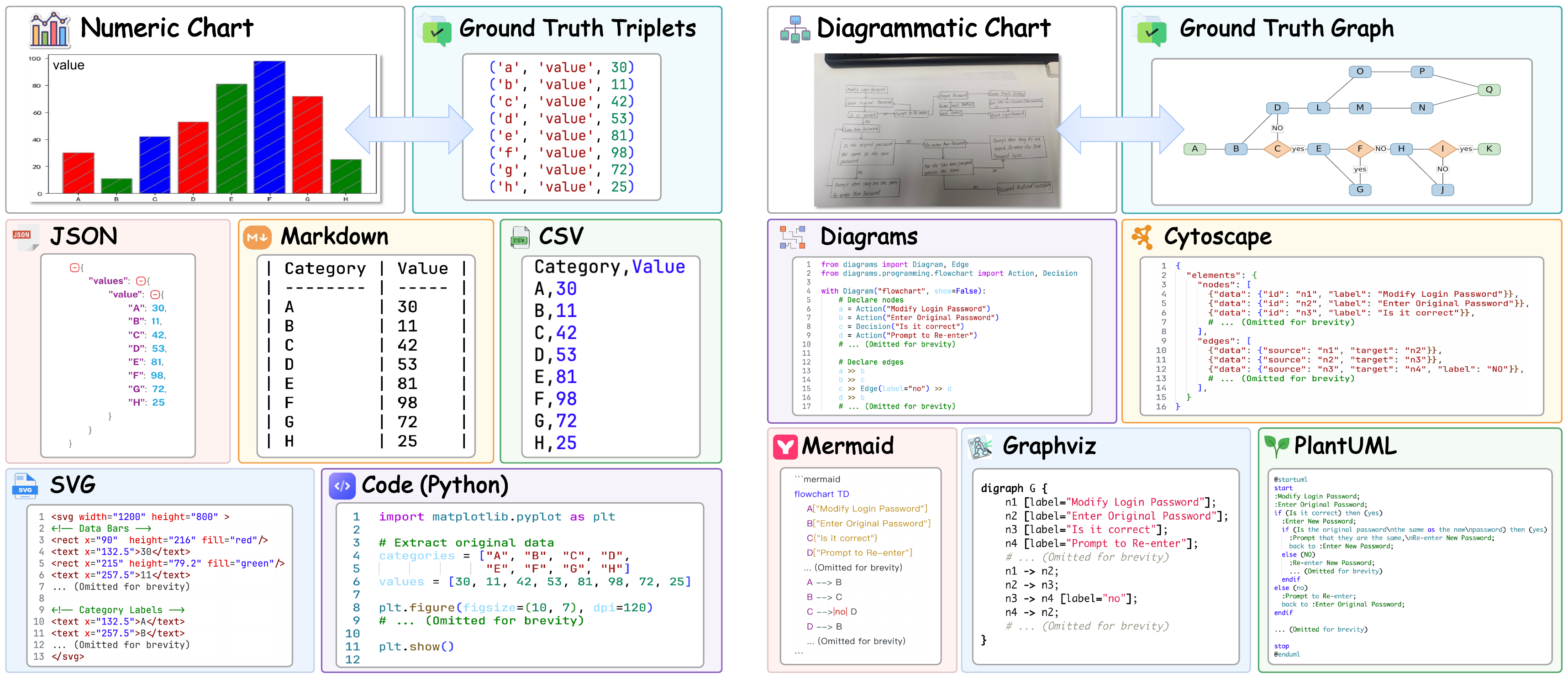}
    \vspace{-1.5em}
    \caption{
        \textbf{Heterogeneous output formats.} Existing models parse charts into disparate formats, making direct cross-model evaluation difficult and motivating a unified, format-agnostic protocol.
    }
    \label{fig:Chart_output_formats}
\end{figure}

Unlike tasks such as table~\cite{shen2023divide, data_fintabnet, data_pubtabnet, StrucTab_2026, CC_OCR_2025, TabPedia_2024} or formula~\cite{cdm_eval, data_hme, data_unimernet} parsing, which benefit from unified evaluation standards, chart understanding is hindered by severe fragmentation, presenting a major challenge for standardized assessment. This fragmentation manifests in three distinct ways. First, output formats are strictly siloed (\cref{fig:Chart_output_formats}); parsers emit results in disjoint syntactic forms (\myeg Markdown, JSON, CSV), rendering direct cross-model evaluation intractable. Second, existing benchmarks isolate narrow sub-domains, failing to provide a unifying taxonomy across numeric and diagrammatic charts. Third, datasets sample pristine synthetic images, overlooking real-world visual perturbations~\cite{wilddoc, li2026towardsrealworlddocument}.

Beyond the evaluation bottleneck, current end-to-end parsers~\cite{HunyuanOCR_2025, PaddleOCR_VL_2025, Dolphin_2025} also suffer from a critical architectural limitation. Most existing methods treat chart understanding as a black-box mapping from pixels directly to serialized text, relying on shallow sequence memorization instead of explicit structural reasoning. By bypassing intermediate semantics such as axis grounding and relational topology, these models lack a meaningful structural inductive bias and remain fragile under real-world presentation shifts. Taken together, these challenges raise a key question: \textit{How can we unify chart parsing evaluation while equipping models with stronger structural reasoning?}

To tackle these challenges, we present a holistic framework that unifies both the evaluation paradigm and the parsing architecture. For evaluation, we introduce \textbf{ChartArena}, the most comprehensive chart parsing benchmark to date (\cref{tab:chart_benchmark_comparison}). It spans eight families across numeric charts and diagrammatic structures, bilingual content, and three visual scenarios (digital renderings, printed photos, and hand-drawn photos). ChartArena is built through a human-agent collaborative annotation pipeline, where model-assisted drafts are iteratively refined through multi-stage human verification to ensure structural consistency and annotation reliability. We further design \textbf{a format-agnostic evaluation protocol} that deterministically maps heterogeneous model outputs into canonical semantic spaces. This enables direct cross-paradigm comparison with structure-aware metrics. With ChartArena, we evaluate a broad range of chart-specialized models as well as open-source and closed-source general MLLMs, revealing the current capability landscape and key failure modes in chart parsing.

To address the modeling gap, we explore reasoning-oriented architectures by proposing \textbf{Chart Atomic Primitives (CAP)}, which abstract the minimal semantic scaffold needed for robust chart comprehension. Integrated into the supervised fine-tuning stage, CAP encourages the model to first construct an explicit structural topology before generating the final sequence. Evaluated on a lightweight 1B-parameter model, our method delivers substantial parsing gains over direct modeling baselines, clearly demonstrating the benefit of explicit structural reasoning.

The primary contributions of this work are summarized as follows:
\vspace{-0.5em}
\begin{itemize}[
        label=\raisebox{0.4ex}{\tiny$\bullet$},
        leftmargin=1em,
        itemsep=2pt,
        parsep=0pt,
        topsep=0pt,
        partopsep=0pt
    ]
    \item \textbf{ChartArena benchmark.} We construct a unified benchmark spanning eight numeric and diagrammatic chart families across three visual scenarios. Featuring bilingual content and high-quality annotations, it faithfully captures the readability uncertainty inherent in real-world data.
    \item \textbf{Format-agnostic evaluation protocol.} We design a deterministic protocol that projects heterogeneous model outputs into shared canonical views. Leveraging this protocol, extensive evaluations across leading MLLMs reveal the current capability landscape and systemic vulnerabilities.
    \item \textbf{Reasoning-augmented supervision.} We introduce Chart Atomic Primitives to compel explicit structural reasoning before sequence generation. Evaluated on a 1B model, this approach outperforms direct modeling baselines, validating the fundamental advantage of structural abstraction.
\end{itemize}

\section{Related Work}
\label{sec:related_work}

\mypara{Evolution of chart parsing.}
The community initially approached chart parsing as a modular pipeline, combining optical character recognition with heuristic geometry~\cite{ChartSense_2017, Revision_2011}. These cascaded systems suffered from compounding errors and struggled with real-world visual noise~\cite{parsing_table_wild_2021, RealCQA_2023, EvoChart_2025, Chart_QA_real_2025}. The paradigm shifted dramatically with the introduction of MLLMs~\cite{Qwen2_5_VL_2025}. Recent literature reformulates chart extraction as an end-to-end sequence generation problem~\cite{HunyuanOCR_2025, PaddleOCR_VL_2025}. Despite remarkable progress, current architectures typically map raw pixels directly to final target strings (\myeg Markdown) while neglecting explicit intermediate structural reasoning~\cite{ChartX_ChartVLM_2025}. This direct mapping forces models to bypass the intermediate cognitive steps that humans instinctively use, such as anchoring axis semantics and localizing atomic data points. Consequently, existing parsers frequently fail to capture complex topological dependencies and lack interpretability, which severely impacts downstream analysis~\cite{OneChart_2024, ChartSense_2017}.

\mypara{Output paradigms and evaluation.}
Despite rapid advances in MLLMs for document understanding, chart parsing remains fragmented by a lack of unified standards. First, representational modalities are highly heterogeneous. Existing methods serialize extracted data from numeric charts into divergent formats, including Markdown~\cite{HunyuanOCR_2025, PaddleOCR_VL_2025, TinyChart_2024, ChartAssistant_2024}, SVG~\cite{Dots_mOCR_2026}, Python code~\cite{TinyChart_2024, MSRL_2025, RRVF_2025, ChartCoder_2025}, HTML table~\cite{Dots_mOCR_2026}, CSV~\cite{ChartX_ChartVLM_2025, ChartMoE_2024, ExChart_Bench_2026}, or JSON structures~\cite{OneChart_2024, ChartVSR_2026}. This structural diversity severely hinders cross-model comparisons and complicates the establishment of fair benchmarks. Second, current evaluation frameworks remain limited in scope and consistency. As shown in~\cref{tab:chart_benchmark_comparison}, recent evaluations typically rely on benchmarks that primarily target numeric charts~\cite{PaddleOCR_VL_2025}, systematically ignoring diagrammatic structures like flowcharts and mind maps. Consequently, there is a pressing need for a unified evaluation framework that covers diverse chart types and normalizes heterogeneous output representations, thereby enabling systematic progress in the field.

\mypara{Data synthesis and structural reasoning.}
The data engines driving recent chart parsers mainly rely on code-to-image synthesis. These pipelines inject semantic data into templating libraries to render millions of image and text pairs~\cite{ChartGalaxy_2025, ChartNet_2026, SynChart_2024}. While highly successful in scaling data volume, this methodology optimizes for visual variety at the expense of structural depth. It inadvertently trains models to memorize surface layouts rather than comprehend atomic semantic primitives, such as coordinate mappings and legend interactions. In text-centric domains like mathematical reasoning~\cite{DeepSeekMath_2024, Uni_DPO_2026, Dr_GRPO_2025} and code generation~\cite{SCoTs_2025}, chain-of-thought methodologies~\cite{CoT_2022} have proven highly effective at mitigating shallow memorization by enforcing intermediate logical steps. Chart parsing requires a similar structural inductive bias. Nevertheless, the systematic integration of granular, atomic reasoning steps to explicitly map the topological structure of visual data remains a largely underexplored frontier in chart parsing. Further discussions of related work are in~\cref{supp:sec:further_related_work}.

\section{Benchmark and Unified Evaluation}
\label{sec:benchmark_evaluation}

As discussed in~\cref{sec:related_work}, chart parsing remains fragmented in three ways: (i) models emit predictions in heterogeneous output formats, making direct cross-model evaluation difficult; (ii) existing benchmarks focus on isolated sub-domains and rarely unify numeric and diagrammatic charts under a common taxonomy; and (iii) most datasets rely on pristine synthetic renderings without covering real-world visual perturbations such as printed or hand-drawn photographs. To address these issues, we construct a unified benchmark together with a format-agnostic evaluation protocol that maps heterogeneous outputs into shared semantic views, enabling fair comparison across different parsing paradigms.

\begin{table*}[t]
    \centering
    \captionsetup{font={small}}
    \caption{
        \textbf{Comparison of Chart Parsing Benchmarks.}
        Our \textbf{ChartArena} provides the most comprehensive coverage across chart types, visual scenarios, and languages, enabling realistic evaluation of chart parsing.
    }
    \vspace{-0.6em}
    {
        \renewcommand{\arraystretch}{0.98}
        \setlength{\tabcolsep}{4pt}
        \resizebox{\textwidth}{!}{%
            \begin{tabular}{l c c cccccccc ccc cc}
                \toprule
                \multirow{2}{*}[-2.8mm]
                {\textbf{Benchmark}}                          &
                \multirow{2}{*}[-2.8mm]
                {\textbf{\makecell{Release \nextline Date}}}  &
                \multirow{2}{*}[-2.8mm]
                {\textbf{Size}}                               &
                \multicolumn{8}{c}{\textbf{Chart Types}}      &
                \multicolumn{3}{c}{\textbf{Image Styles}}     &
                \multicolumn{2}{c}{\textbf{Languages}}
                \\
                \cmidrule(lr){4-11}
                \cmidrule(lr){12-14}
                \cmidrule(lr){15-16}
                                                              &
                                                              &
                                                              &
                Bar                                           &
                Line                                          &
                Pie                                           &
                Radar                                         &
                \makecell{Box \nextline Plot}                 &
                \makecell{Comb. \nextline Chart}              &
                \makecell{Flow- \nextline chart}              &
                \makecell{Mind \nextline Map}                 &
                \makecell{Digital \nextline Rendering}        &
                \makecell{Printed \nextline Photo}            &
                \makecell{Hand-drawn \nextline Photo}         &
                English                                       &
                Chinese
                \\
                \midrule
                PlotQA-SE~\cite{OneChart_2024, PlotQA_2020}   & 2019.09 & 33,657 & \cmark & \cmark &        &        &        &        &        &        & \cmark &        &        & \cmark &        \\
                ChartQA-SE~\cite{OneChart_2024, ChartQA_2022} & 2022.03 & 1,509  & \cmark & \cmark & \cmark &        &        &        &        &        & \cmark &        &        & \cmark &        \\
                MMC-Bench~\cite{MMC_Bench_2024}               & 2023.11 & 1,063  & \cmark & \cmark & \cmark & \cmark &        &        &        &        & \cmark &        &        & \cmark &        \\
                ChartX-SE~\cite{ChartX_ChartVLM_2025}         & 2024.02 & 1,152  & \cmark & \cmark & \cmark & \cmark & \cmark &        &        &        & \cmark &        &        & \cmark &        \\
                ChartY~\cite{OneChart_2024}                   & 2024.04 & 6,048  & \cmark & \cmark & \cmark &        &        & \cmark &        &        & \cmark &        &        & \cmark & \cmark \\
                VG-DCU~\cite{VG_DCU_2024}                     & 2024.04 & 3,044  & \cmark & \cmark & \cmark &        & \cmark & \cmark &        &        & \cmark &        &        & \cmark &        \\
                ChartP-Bench~\cite{ChartVSR_2026}             & 2026.02 & 1,200  & \cmark & \cmark &        &        &        &        &        &        & \cmark &        &        & \cmark &        \\
                ParseBench~\cite{ParseBench_2026}             & 2026.04 & 1,039  & \cmark & \cmark & \cmark &        &        & \cmark &        &        & \cmark &        &        & \cmark &        \\
                ExChart-Bench~\cite{ExChart_Bench_2026}       & 2026.04 & 3,600  & \cmark & \cmark & \cmark & \cmark &        &        &        &        & \cmark &        &        & \cmark &        \\
                \midrule
                \rowcolor{table_ours}
                \textbf{ChartArena}                           & 2026.05 & 2,400  & \cmark & \cmark & \cmark & \cmark & \cmark & \cmark & \cmark & \cmark & \cmark & \cmark & \cmark & \cmark & \cmark \\
                \bottomrule
            \end{tabular}
        }
    }
    \label{tab:chart_benchmark_comparison}
\end{table*}

\subsection{Benchmark Construction}
\label{subsec:benchmark_construction}

\begin{figure}[t!]
    \centering
    \includegraphics[width=\linewidth]{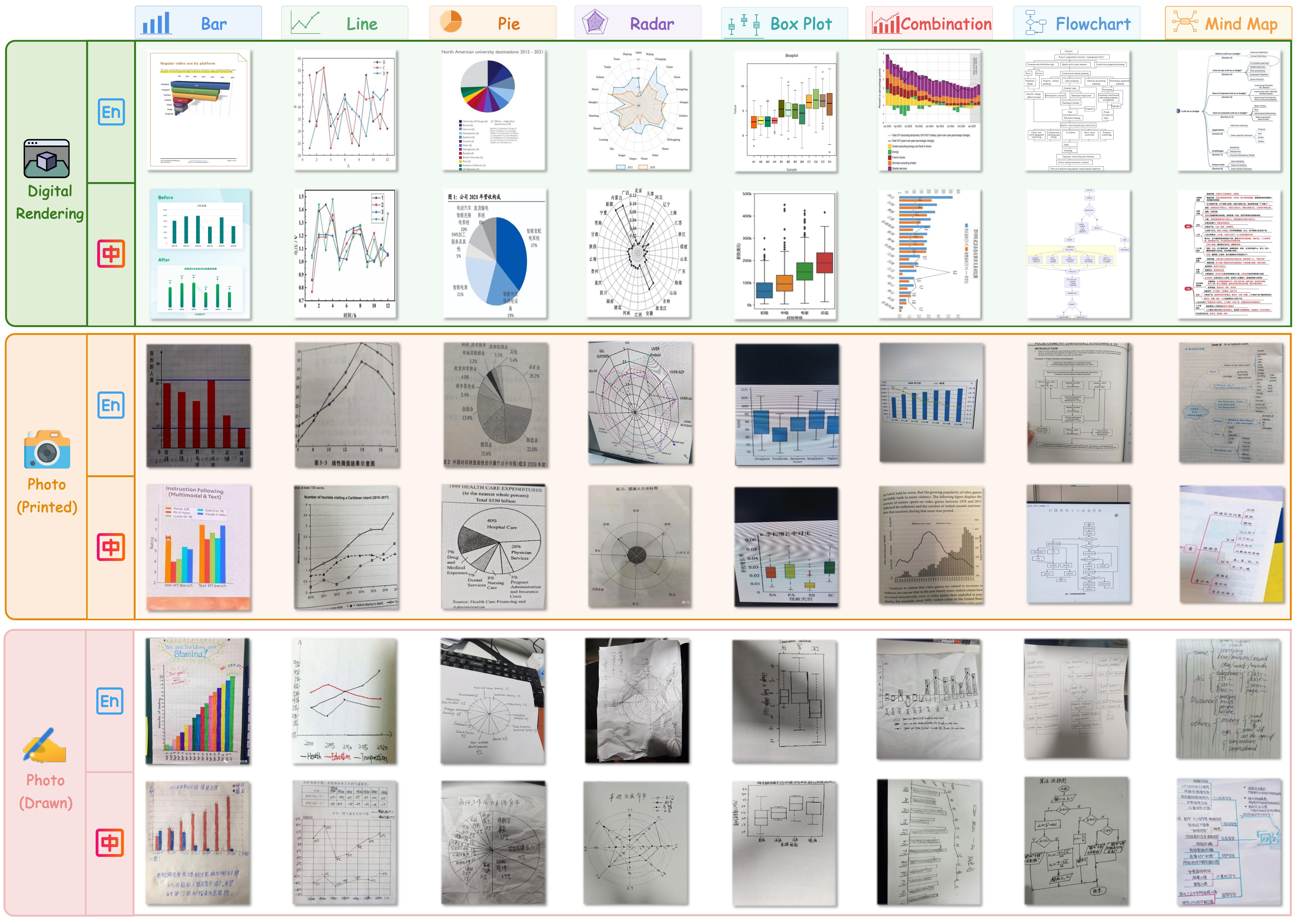}
    \vspace{-1.8em}
    \caption{
        \textbf{Benchmark overview.}
        Our benchmark covers eight chart types spanning both numeric and diagrammatic categories. For each type, we include three visual scenarios (digital rendering, printed photo, and hand-drawn photo) and two languages (English and Chinese), with 50 samples per setting, resulting in a total of 2,400 charts for comprehensive and unified evaluation of chart parsing.
    }
    \label{fig:ChartArena_overview}
    \vspace{-1.6em}
\end{figure}

\mypara{Task coverage.}
Our benchmark is organized along three explicit axes of diversity:
(a) \emph{chart family}: eight types spanning both \emph{numeric charts} (bar, line, pie, radar, box plot, and combination chart) and \emph{diagrammatic charts} (flowchart and mind map);
(b) \emph{visual scenario}: clean digital renderings as well as real-world sources such as printed and hand-drawn photos;
(c) \emph{language}: bilingual Chinese and English content.
As in~\cref{fig:ChartArena_overview}, ChartArena explicitly stress-tests chart parsers on the combinations most commonly encountered in real-world applications yet missing from prior benchmarks~\cite{PlotQA_2020, ChartQA_2022}.

\mypara{Data collection.}
Following recent studies~\cite{CC_OCR_2025, CCpdf_2023}, we curate chart images from public document corpora, web sources, and in-house collections. We explicitly over-sample under-represented scenarios (\myeg printed and hand-drawn charts) to avoid trivializing robustness. Details are shown in~\cref{supp:subsec:image_sources}.

\mypara{Annotation pipeline.}
To balance efficiency and quality, we adopt a hybrid annotation strategy that combines model-assisted drafting with multi-stage human refinement on a shared human--agent co-annotation platform. For each chart, MLLMs first generate coarse annotations aligned with the chart type (\myeg Markdown for numeric charts and Mermaid for diagrammatic charts), which substantially accelerates annotation. Human annotators then refine both structural elements, such as chart composition, node relations, and axis semantics, and semantic content, including labels and numerical values. When values are difficult to read, multiple annotators cross-check them to improve consistency and reliability. This collaborative pipeline produces high-quality annotations with strong structural consistency across diverse chart types and scenarios. Details are provided in~\cref{supp:subsec:annotation_protocol}.

\subsection{Unified Evaluation Protocol}
\label{subsec:unified_eval_protocol}

As discussed in~\cref{sec:related_work}, chart parsing models produce outputs in diverse formats and various diagram languages. These heterogeneous representations hinder direct comparison, as models often differ in \emph{how} results are expressed rather than \emph{what} is extracted. A unified evaluation protocol is therefore essential for fair and consistent assessment.

\begin{figure}[t!]
    \centering
    \includegraphics[width=\linewidth]{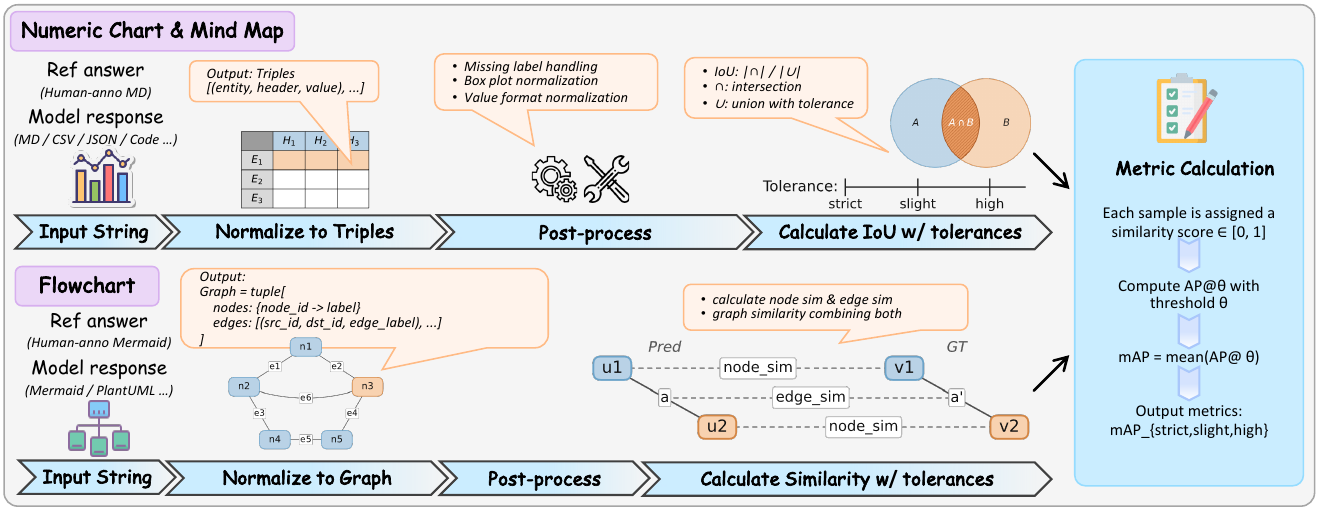}
    \vspace{-1.6em}
    \caption{
        \textbf{Evaluation protocol.}
        We first normalize predictions and references into structured representations (triples for numeric charts and mind maps, and graphs for flowcharts), followed by a format-agnostic post-processing stage. We then compute tolerance-aware similarity (IoU for triples and graph similarity via node and edge matching), and finally aggregate results into unified scores.
    }
    \label{fig:evaluation_protocol}
\end{figure}

\mypara{Format-agnostic normalization.}
As illustrated in~\cref{fig:evaluation_protocol}, we first normalize both predictions and references into structured representations. Specifically, numeric charts are mapped into a \emph{triple view} that captures values with associated entities and headers, following recent studies~\cite{ChartX_ChartVLM_2025, OneChart_2024}. Flowcharts are converted into a \emph{graph view} with nodes and labeled edges. This design abstracts away surface formats and ensures that semantically equivalent outputs become directly comparable.

\mypara{Structure-aware metrics with annotation-aware tolerance.}
For numeric charts, we follow StructChart's SCRM~\cite{StructChart_2026}, using tolerance-aware triple matching and IoU-based similarity. We retain its three tolerance levels (\emph{strict}, \emph{slight}, \emph{high}) and report Exact Match (EM) and mean Average Precision (mAP) over multiple thresholds. Our protocol adapts this scoring to heterogeneous output formats through canonical normalization and extends coverage to diagrammatic charts through graph and tree matching. For flowcharts, similarity combines node and edge matching. The inherited metric design and our format-specific adaptations are detailed in~\cref{supp:sec:evaluation_details}.

\section{Data Pipeline}
\label{sec:data_pipeline}

Our data construction framework consists of three components:
(i) a semantic-aligned synthesis pipeline that produces chart images with structured annotations (\cref{subsec:synthesis_pipeline}),
(ii) a CAP extractor that extracts foundational semantics from corpora (\cref{subsec:cap}), and
(iii) a reasoning construction module that converts CAP into unified, reasoning-augmented supervision targets (\cref{subsec:cap_reasoning}).

\subsection{Semantic-Aligned Chart Synthesis Pipeline}
\label{subsec:synthesis_pipeline}

To endow the model with robust structural priors, we build a large-scale data synthesis pipeline that guarantees explicit semantic grounding across diverse chart families.

\mypara{Numeric charts.}
For numeric domains, including bar, line, pie, radar, box plot, and combination charts, we use Qwen3-235B-A22B-Instruct~\cite{Qwen3_2025} to synthesize both semantic corpora and chart template libraries. The semantic module generates structured key--value configurations, such as category names, series definitions, and numerical fields with coherent internal logic. The template library covers diverse visual styles. Each template fixes both the rendering recipe and the Markdown serialization format. By pairing semantic instances with templates and geometric layouts, we obtain million-scale image--annotation pairs for each chart family.

\mypara{Diagrammatic charts.}
For diagrammatic domains (flowcharts and mind maps), the same LLM directly generates the structured target. For flowcharts, it emits executable Mermaid programs that explicitly encode node semantics and edge relations, which are then rendered into images; for mind maps, it produces hierarchical Markdown that captures the content organization. This guarantees that each chart image faithfully mirrors the underlying topology and the parsing target.

\begin{figure}[t!]
    \centering \includegraphics[width=\linewidth]{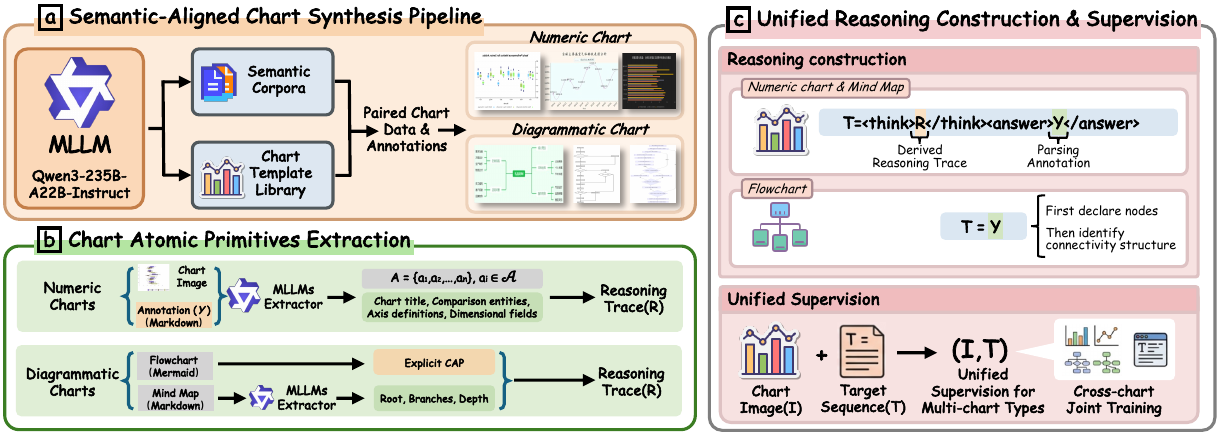}
    \vspace{-1.6em}
    \caption{
        \textbf{Overview of data pipeline.}
        (a) LLM-based synthesis stage generates chart images $I$ and annotations $Y$.
        (b) CAP extraction stage derives atomic semantics $A$ from $(I, Y)$.
        (c) Unified reasoning stage constructs targets $T$, enabling all chart types to share a single $(I, T)$ training interface.
    }
\end{figure}

\subsection{Chart Atomic Primitives}
\label{subsec:cap}

To bridge the semantic gap between raw pixels and serialized output, we introduce \textbf{Chart Atomic Primitives (CAP)} as a compact intermediate representation that isolates the minimal semantic scaffold needed for chart comprehension.
Formally, given a chart image $I$ and its target annotation $Y$, the corresponding CAP is defined as:

\vspace{-1.0em}
\begin{equation}
    A = \{a_1, a_2, \dots, a_n\}, \quad a_i \in \mathcal{A}\, ,
    \label{eq: cap}
\end{equation}
\par\nobreak\vspace{-0.5em}

where each primitive $a_i$ denotes a distinct, chart-level semantic component. By decoupling instance-level specificities from the underlying topology, CAP extracts structural semantics such as chart titles, comparison entities, axis definitions, dimensional fields, and core statistical attributes.

\mypara{Numeric charts.}
For numeric charts, CAP is materialized through schemas specifically tailored to the chart type. For instance, two-dimensional Cartesian charts inherently encode axis semantics and tick ranges, radar charts encapsulate dimensional nodes and radial axes, whereas combination charts capture intricate multi-axis alignments alongside dense statistical fields. Details are in~\cref{supp:subsec:training_data}.

\mypara{Diagrammatic charts.}
For diagrammatic charts, CAP aligns naturally with structured representations. In flowchart-like diagrams, Mermaid supports two common serialization styles: one that interleaves node definitions with edge relations, and another that separates node declarations from connectivity structure. To ensure consistency, we normalize all predictions into a unified form that explicitly separates nodes and edges. This normalized representation directly corresponds to atomic components and their interactions, and is used as the basis for CAP reasoning.
For mind maps, where structure is often implicit, we further abstract high-level primitives such as the root topic, primary branches, hierarchical depth, and semantic domains, enabling consistent structural reasoning across formats.

\subsection{CAP Extraction and Reasoning Construction}
\label{subsec:cap_reasoning}

To convert CAP into supervision signals, we first extract atomic primitives $A$ from $(I, Y)$ for numeric charts and mind maps by prompting Qwen3-VL-235B-A22B-Instruct~\cite{Qwen3_VL_2025}, and then serialize them into a structured reasoning trace. The extracted primitives focus on chart-level semantics such as titles, entities, axes, dimensions, and key statistics, while discarding low-level visual details. The reasoning trace is constructed as:

\vspace{-1.0em}
\begin{equation}
    R = \mathcal{G}(A)\, ,
    \label{eq: reasoning_construction}
\end{equation}
\par\nobreak\vspace{-0.5em}

where $\mathcal{G}(\cdot)$ acts as a templating function that serializes the isolated atomic fields into an ordered sequence of semantic analysis steps. Ultimately, the unified supervision target is formulated as:

\vspace{-1.0em}
\begin{equation}
    T = \texttt{<think>}~R~\texttt{</think>}~\texttt{<answer>}~Y~\texttt{</answer>}\, .
    \label{eq: unified_supervision}
\end{equation}
\par\nobreak\vspace{-0.5em}

This architectural decision injects a profound structural inductive bias into the training loop, forcing the model to articulate a topological scaffold before committing to a final parse.

For flowchart-like diagrams, we define $T = Y$ after normalizing the Mermaid representation into a canonical form that explicitly separates node definitions from edge relations. Only this structure-first representation aligns with CAP, as it exposes atomic components before their interactions.

Although the construction differs by chart type, all samples share the same $(I, T)$ interface, enabling joint optimization across heterogeneous chart types within a single training pipeline.
\section{Training Pipeline}
\label{sec:training_pipeline}

Our training pipeline follows two stages: pretraining learns direct parsing ability, and supervised fine-tuning (SFT) introduces CAP reasoning while adapting the model to broader visual diversity.

\subsection{Pretraining}
\label{subsec:curriculum_pretraining}

A naive strategy that trains directly on reasoning-augmented targets forces the model to learn extraction and reasoning simultaneously, which may create competing supervision signals and hurt optimization efficiency and generalization~\cite{Curriculum_2009, Small_model_learn_2025, Task_Curriculum_2026}. We therefore use pretraining only for direct parsing supervision, so that the model learns the basic mapping from chart images to structured outputs and develops a stable initialization for later reasoning-based adaptation.

\subsection{Supervised Fine-Tuning with CAP Reasoning}
\label{subsec:sft_diversity}

Building on this initialization, SFT introduces CAP reasoning. We augment the training targets with CAP-formatted supervision, encouraging the model to infer structure before generating the final output, thereby improving structural consistency during training.

At the same time, SFT broadens the data manifold along three axes: source heterogeneity, stylistic diversity, and semantic difficulty. We mix public chart understanding datasets with our human-annotated samples (\cref{sec:benchmark_evaluation}) to cover more chart types, layout patterns, and semantic topics. We also apply rendering perturbations such as font variation and geometric warping to improve robustness against presentation shifts. As a result, the model retains the structural priors learned during pretraining while becoming more reliable on rendered charts, printed photos, hand-drawn photos, and bilingual inputs.
\section{Experiments}
\label{sec:experiments}

We organize our experiments on \textbf{ChartArena} around four aspects:
(i) experimental settings, including backbone, training recipe and baselines (\cref{subsec:experimental_settings});
(ii) main comparison against expert chart parsers, general-purpose MLLMs, and document parsing MLLMs (\cref{subsec:main_results});
(iii) the adaptability of our unified evaluation protocol to diverse output formats (\cref{subsec:multi_format});
and (iv) ablations on the training pipeline that isolate the effect of pretraining, SFT, and CAP reasoning (\cref{subsec:ablation}). Further analysis under different visual scenarios is shown in~\cref{supp:subsec:appendix_visual_scenarios}.

\subsection{Experimental Settings}
\label{subsec:experimental_settings}

\mypara{Backbone and training recipe.}
We mainly adopt HunyuanOCR~\cite{HunyuanOCR_2025} as the vision--language backbone.
Full-parameter fine-tuning is used throughout, and all experiments are conducted on 16 NVIDIA H20 GPUs.
For \emph{pretraining}, we combine 2M synthesized chart--target pairs from our pipeline with 2M samples from ChartGalaxy~\cite{ChartGalaxy_2025} and ChartNet~\cite{ChartNet_2026}, and train for 1 epoch with a learning rate of $4\times10^{-5}$.
For \emph{SFT}, we use the 60K human-annotated examples released by ChartNet together with our additional 20K in-house annotations, and train for 1 epoch with a learning rate of $2\times10^{-5}$.
Detailed training settings are provided in~\cref{supp:subsec:training_setup}.

\mypara{Baselines.}
We compare our method against representative baselines across three categories:
(a) \textit{General-purpose MLLMs}, capable of chart parsing through unified multimodal modeling;
(b) \textit{Document parsing MLLMs}, optimized for holistic document parsing;
and (c) \textit{Expert chart understanding models}, dedicated parsers explicitly designed for chart structure recognition and parsing.
For each category, we select strong and representative models to ensure a comprehensive evaluation.
Each baseline is run in its native output format, and outputs are normalized before scoring under the unified protocol of~\cref{subsec:unified_eval_protocol}.
Slices outside the declared capability of a model are reported as ``--'', yielding an per-model all-task \emph{Average} columns over all eight chart families.

\subsection{Main Results}
\label{subsec:main_results}

\begin{table*}[t!]
  \centering
  \captionsetup{font={small}}
  \caption{
    \textbf{Main results on ChartArena.}
    We report detailed performance for each chart type as well as the overall average.
    Each type includes two scores: \textbf{EN} (English) and \textbf{ZH} (Chinese),
    both averaged over three visual styles (digital renderings, printed photos, and hand-drawn photos).
    The \textbf{Average} columns are all-task means over all eight chart families.
    For each metric, we report \textbf{mAP$_{\text{high}}$}.
    Within each model category, \textbf{bold} and \underline{underline} denote the best and second-best results within each chart type.
  }
  \vspace{-0.4em}
  {
    \renewcommand{\arraystretch}{1.1}
    \setlength{\tabcolsep}{4pt}
    \resizebox{\textwidth}{!}{
      \begin{tabular}{clc cc cc cc cc cc cc cc cc cc}
        \toprule
        \multirow{2}{*}[-0.8mm]
        {\textbf{\makecell{Model \nextline Type}}}   &
        \multirow{2}{*}[-0.8mm]
        {\textbf{Model}}                             &
        \multirow{2}{*}[-0.8mm]
        {\textbf{\makecell{Release \nextline Date}}} &
        \multicolumn{2}{c}{\textbf{bar}}             &
        \multicolumn{2}{c}{\textbf{line}}            &
        \multicolumn{2}{c}{\textbf{pie}}             &
        \multicolumn{2}{c}{\textbf{radar}}           &
        \multicolumn{2}{c}{\textbf{box plot}}        &
        \multicolumn{2}{c}{\textbf{comb. chart}}     &
        \multicolumn{2}{c}{\textbf{flowchart}}       &
        \multicolumn{2}{c}{\textbf{mind map}}        &
        \multicolumn{2}{c}{\textbf{Average}}
        \\
        \cmidrule(lr){4-5}
        \cmidrule(lr){6-7}
        \cmidrule(lr){8-9}
        \cmidrule(lr){10-11}
        \cmidrule(lr){12-13}
        \cmidrule(lr){14-15}
        \cmidrule(lr){16-17}
        \cmidrule(lr){18-19}
        \cmidrule(lr){20-21}
                                                     &
                                                     &
                                                     &
        EN                                           & ZH                                                              &
        EN                                           & ZH                                                              &
        EN                                           & ZH                                                              &
        EN                                           & ZH                                                              &
        EN                                           & ZH                                                              &
        EN                                           & ZH                                                              &
        EN                                           & ZH                                                              &
        EN                                           & ZH                                                              &
        EN                                           & ZH
        \\
        \midrule
        \multirow{16}{*}
        {\makecell{General \nextline Purpose \nextline MLLMs}}
                                                     & \logo{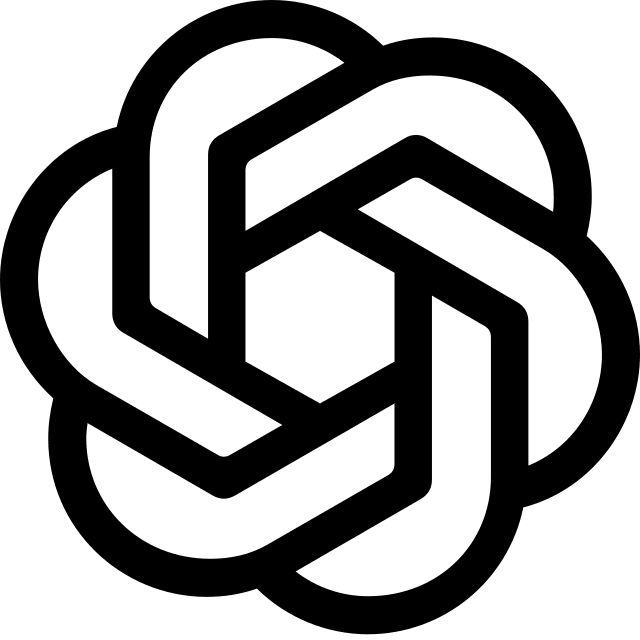}GPT-4o~\cite{GPT_4o_2024}                          & 2024.05 & 21.6          & 36.3          & 27.5          & 52.9          & 76.7          & 74.2          & 9.7           & 24.9          & 19.1          & 9.6           & 9.9           & 40.7          & 49.8          & 27.1          & 64.0          & 24.8          & 34.8          & 36.3          \\
                                                     & \logo{openai}GPT-5~\cite{GPT_5_2025}                            & 2025.08 & 35.1          & 52.3          & 48.1          & 65.1          & 81.1          & 78.9          & \textbf{32.0} & 41.5          & 19.8          & 12.8          & 14.2          & 46.5          & 58.1          & 35.3          & 76.6          & 33.5          & 45.6          & 45.8          \\
                                                     & \logo{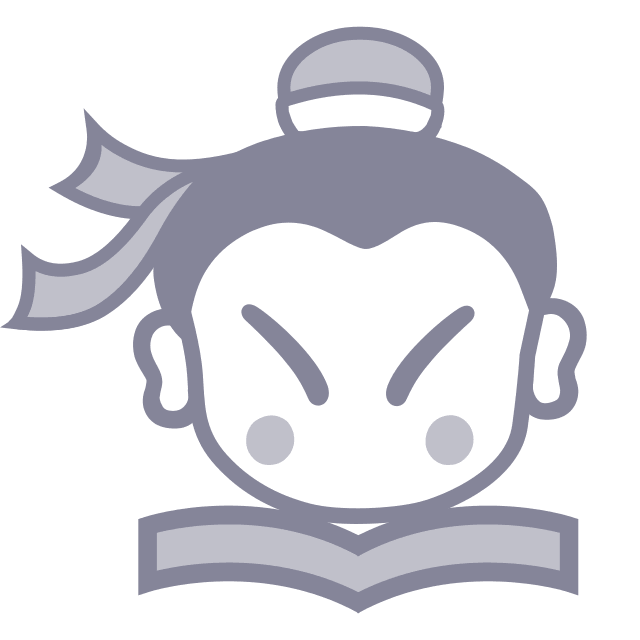}InternVL3.5-8B~\cite{InternVL3_5_2025}             & 2025.08 & 22.7          & 52.6          & 34.4          & 53.7          & 65.8          & 73.8          & 14.0          & 34.7          & 5.6           & 9.5           & 11.3          & 42.1          & 32.6          & 23.8          & 48.3          & 31.8          & 29.3          & 40.2          \\
                                                     & \logo{intern}InternVL3.5-241B-A28B~\cite{InternVL3_5_2025}      & 2025.08 & 27.5          & 57.2          & 41.3          & 55.7          & 77.7          & 83.3          & 15.2          & 41.4          & 18.7          & 21.6          & 17.7          & 47.8          & 43.8          & 36.6          & 62.6          & 45.5          & 38.0          & 48.6          \\
                                                     & \logo{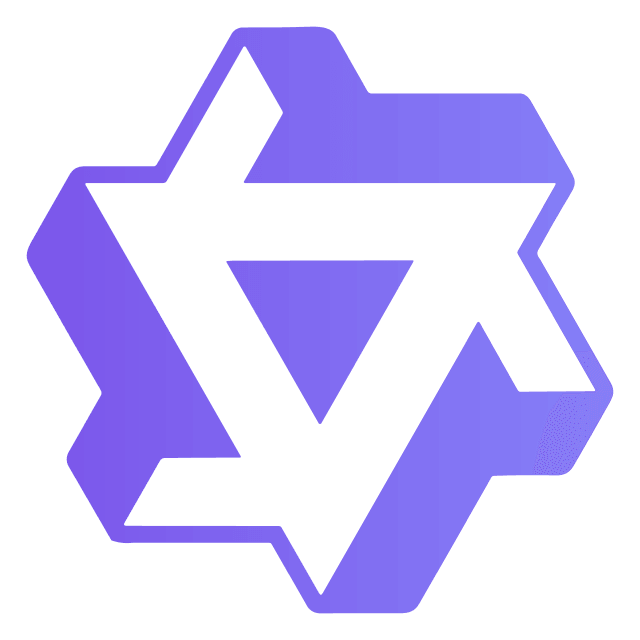}Qwen2.5-VL-7B-Ins.~\cite{Qwen2_5_VL_2025}            & 2025.01 & 15.2          & 36.9          & 17.9          & 39.9          & 63.4          & 73.1          & 8.3           & 19.1          & 0.9           & 2.8           & 6.0           & 40.6          & 29.7          & 23.2          & 45.4          & 29.9          & 23.3          & 33.2          \\
                                                     & \logo{qwen}Qwen2.5-VL-72B-Ins.~\cite{Qwen2_5_VL_2025}           & 2025.01 & 27.1          & 53.3          & 38.2          & 66.7          & 73.5          & 77.0          & 10.9          & 38.5          & 15.0          & 15.3          & 14.3          & 50.5          & 50.1          & 43.6          & 63.8          & 55.0          & 36.6          & 50.0          \\
                                                     & \logo{qwen}Qwen3-VL-8B-Ins.~\cite{Qwen3_VL_2025}                & 2025.10 & 27.5          & 58.6          & 35.5          & 61.1          & 77.3          & 84.7          & 16.8          & 42.6          & 11.6          & 12.1          & 13.2          & 47.9          & 50.0          & 41.5          & 66.4          & 54.6          & 37.3          & 50.4          \\
                                                     & \logo{qwen}Qwen3-VL-235B-A22B-Ins.~\cite{Qwen3_VL_2025}         & 2025.09 & 38.4          & 67.9          & 52.3          & 73.8          & 82.6          & 85.5          & 23.2          & 52.4          & 14.1          & 14.1          & 29.1          & 58.2          & 57.9          & 49.8          & 70.8          & 65.2          & 46.0          & 58.4          \\
                                                     & \logo{qwen}Qwen3.5-35B-A3B (think)~\cite{Qwen3_5_2026}          & 2026.02 & \ul{46.2}     & 65.3          & 60.3          & 77.6          & 89.7          & 88.4          & 25.2          & 57.8          & \ul{42.2}     & \ul{50.6}     & 31.5          & 56.9          & 62.5          & 56.5          & 75.1          & 70.9          & 54.1          & 65.5          \\
                                                     & \logo{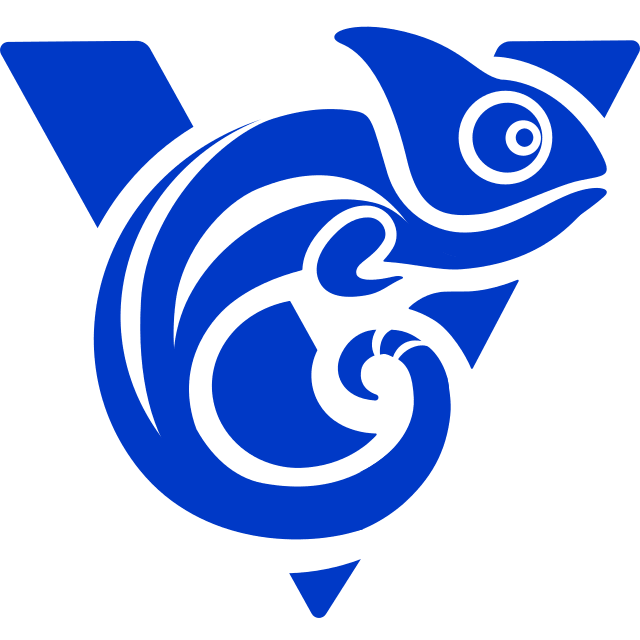}GLM-4.5V~\cite{GLM_4_5V_2025}                        & 2025.08 & 33.5          & 61.4          & 51.7          & 70.5          & 81.2          & 83.1          & 19.7          & 43.1          & 32.4          & 37.4          & 21.2          & 52.5          & 44.7          & 39.6          & 66.2          & 43.7          & 43.8          & 53.9          \\
                                                     & \logo{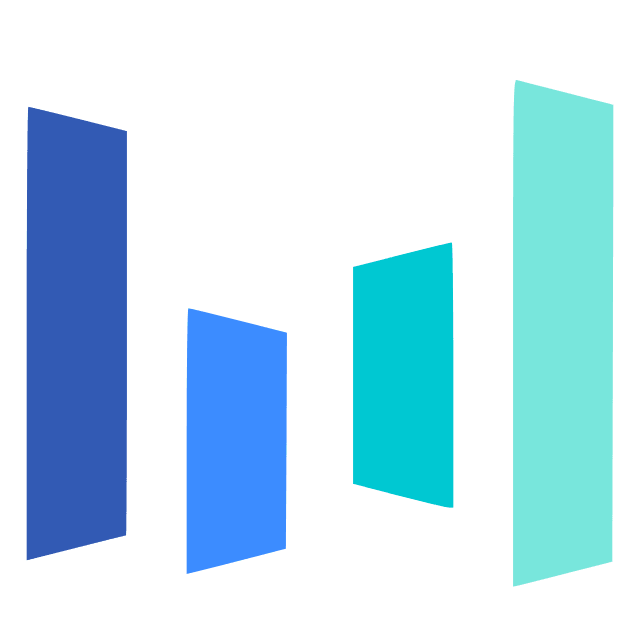}Seed1.8 (no-think)~\cite{Seed_1_8_2026}         & 2025.12 & 29.1          & 59.7          & 46.0          & 72.5          & 84.7          & 88.0          & 22.0          & 45.9          & 16.1          & 17.5          & 15.0          & 59.7          & 47.8          & 50.3          & 76.5          & 69.1          & 42.2          & 57.8          \\
                                                     & \logo{bytedance}Seed2.0 Pro (no-think)~\cite{Seed_2_0_2026}     & 2026.02 & 40.3          & 73.3          & 56.5          & \ul{80.7}     & \ul{91.5}     & \ul{90.5}     & 21.3          & 54.7          & \textbf{44.5} & \textbf{55.2} & 32.4          & 62.2          & \ul{62.6}     & \ul{61.3}     & \ul{83.1}     & \textbf{85.8} & 54.0          & \ul{70.5}     \\
                                                     & \logo{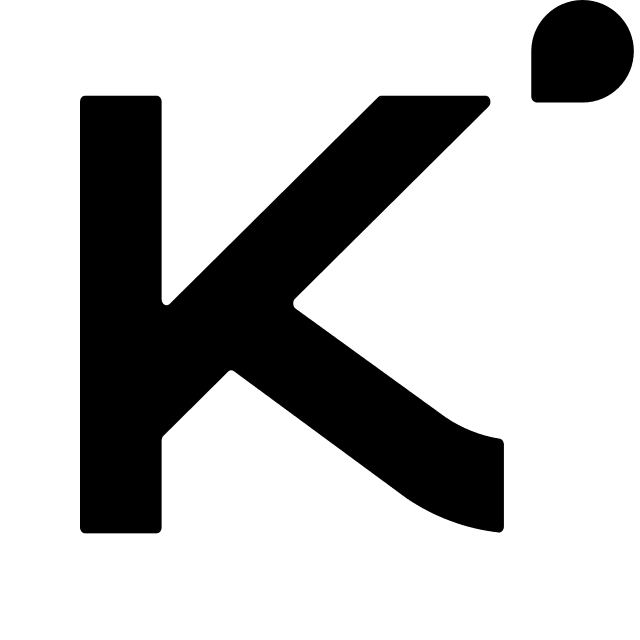}Kimi K2.5 (no-think)~\cite{Kimi_K2_5_2026}           & 2026.01 & 45.2          & 70.3          & \ul{60.9}     & 79.8          & 87.2          & 86.7          & 30.2          & \ul{59.7}     & 40.6          & 47.6          & \ul{33.6}     & \ul{63.6}     & 59.9          & 57.9          & 80.8          & 79.4          & \ul{54.8}     & 68.1          \\
                                                     & \logo{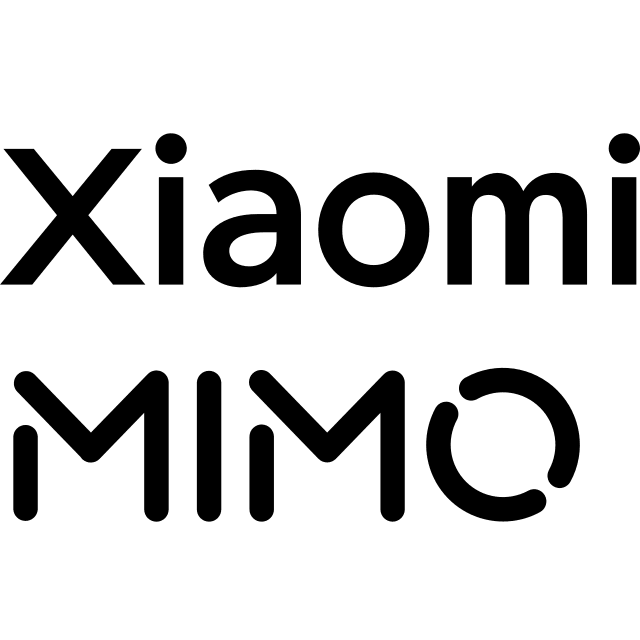}MiMo-V2-Omni~\cite{Mimo_V2_Omni_2026}          & 2026.03 & 31.1          & 56.9          & 41.5          & 66.4          & 87.0          & 85.8          & 19.7          & 46.1          & 19.1          & 30.3          & 19.4          & 54.7          & 57.1          & 51.0          & 76.6          & 64.6          & 43.9          & 57.0          \\
                                                     & \logo{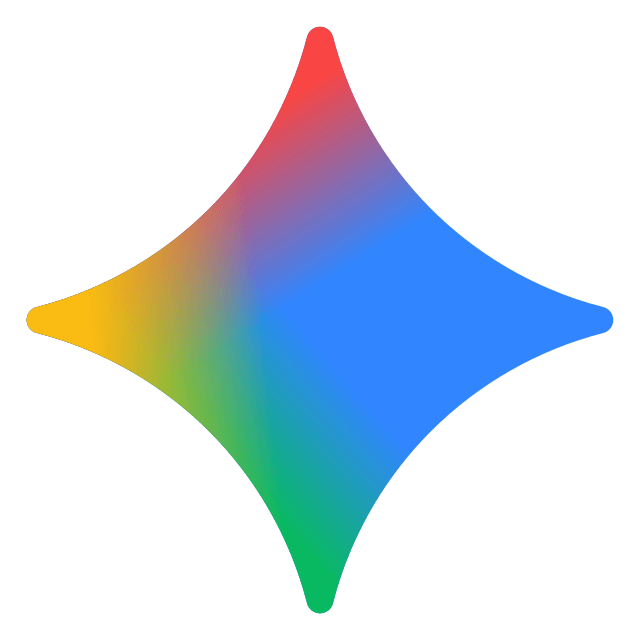}Gemini 2.5 Pro~\cite{Gemini_2_5_2025}              & 2025.03 & 46.0          & \ul{76.5}     & 56.5          & 77.6          & 88.6          & 87.3          & 17.5          & 53.0          & 10.2          & 22.1          & 28.7          & 57.6          & 62.1          & 57.8          & 71.7          & 67.1          & 47.7          & 62.4          \\
                                                     & \logo{gemini}Gemini 3.1 Pro~\cite{Gemini_3_1_2026}              & 2026.02 & \textbf{57.9} & \textbf{78.7} & \textbf{67.0} & \textbf{85.3} & \textbf{92.5} & \textbf{95.1} & \ul{31.8}     & \textbf{62.7} & 32.5          & 45.2          & \textbf{39.7} & \textbf{70.3} & \textbf{65.6} & \textbf{63.1} & \textbf{86.8} & \ul{85.2}     & \textbf{59.2} & \textbf{73.2} \\
        \midrule
        \multirow{3}{*}
        {\makecell{Document \nextline Parsing \nextline MLLMs}}
                                                     & \logo{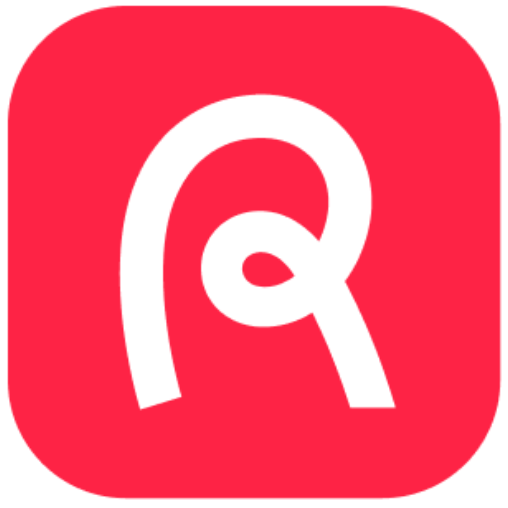}dots.mocr (3B)~\cite{Dots_mOCR_2026}          & 2026.03 & 28.3          & 40.9          & 41.8          & \ul{60.1}     & \ul{68.8}     & \textbf{78.3} & \textbf{20.3} & \textbf{43.1} & \ul{24.1}     & 16.0          & \textbf{26.9} & 47.1          & \ul{26.2}     & \ul{20.6}     & \ul{28.7}     & \ul{19.6}     & \ul{33.1}     & \ul{40.7}     \\
                                                     & \logo{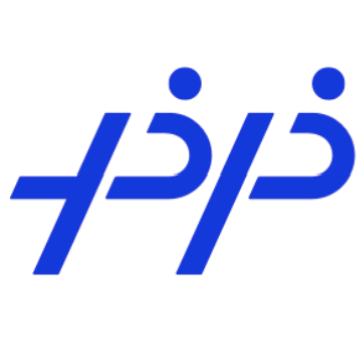}PaddleOCR-VL (1B)~\cite{PaddleOCR_VL_2025}         & 2025.10 & \ul{31.8}     & \ul{49.3}     & \ul{43.0}     & 51.6          & 57.5          & \ul{75.2}     & 14.4          & 29.0          & 11.7          & \ul{20.7}     & \ul{21.3}     & \textbf{54.0} & --            & --            & --            & --            & 22.5          & 35.0          \\
                                                     & \logo{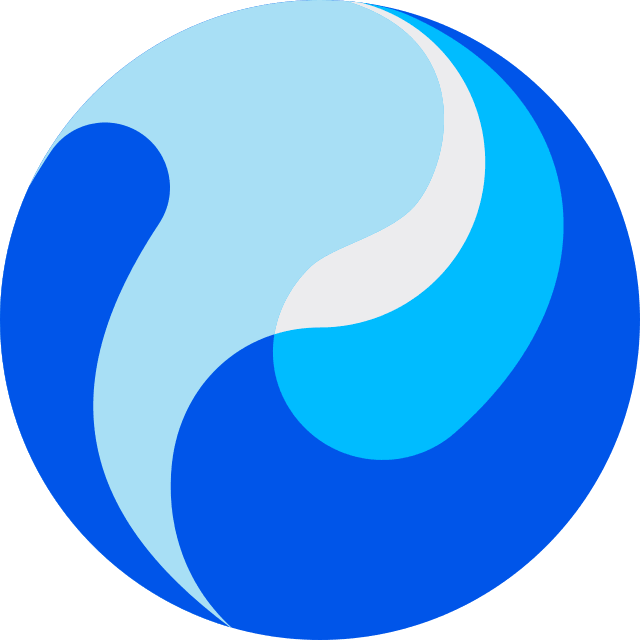}HunyuanOCR (1B)~\cite{HunyuanOCR_2025}            & 2025.11 & \textbf{32.4} & \textbf{58.9} & \textbf{48.6} & \textbf{68.0} & \textbf{70.8} & 72.5          & \ul{19.0}     & \ul{40.9}     & \textbf{43.9} & \textbf{43.9} & 20.1          & \ul{48.3}     & \textbf{38.3} & \textbf{35.5} & \textbf{55.0} & \textbf{46.6} & \textbf{41.0} & \textbf{51.8} \\
        \midrule
        \multirow{8}{*}
        {\makecell{Expert Chart \nextline Understanding \nextline Models}}
                                                     & \logo{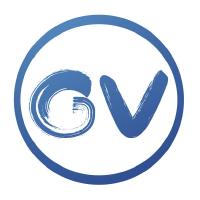}ChartAst (13B)~\cite{ChartAssistant_2024}       & 2024.01 & 5.2           & --            & 4.2           & --            & 0.3           & --            & 1.5           & --            & 0.3           & --            & 0.0           & --            & --            & --            & --            & --            & 1.4           & --            \\
                                                     & \logo{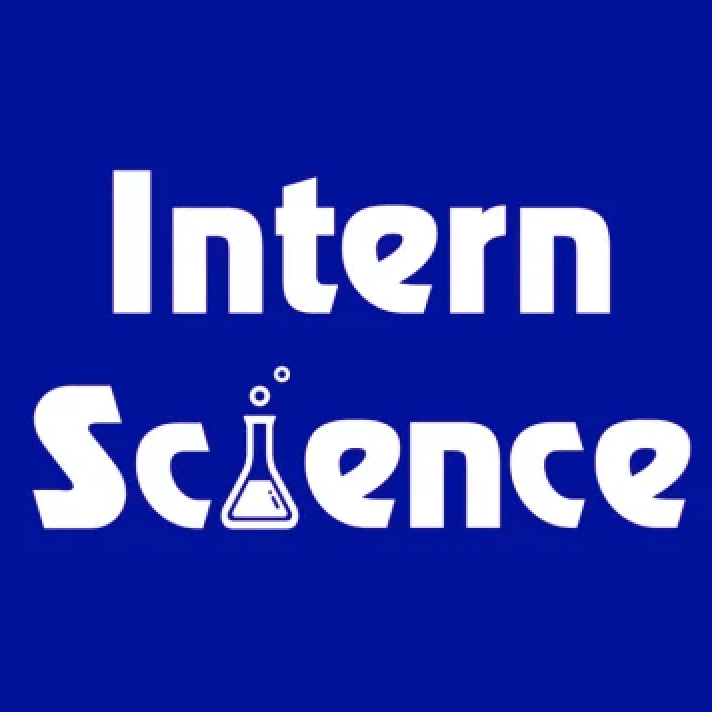}ChartVLM (8.3B)~\cite{ChartX_ChartVLM_2025} & 2024.02 & 11.2          & 5.3           & 11.5          & 4.3           & 12.9          & 8.2           & 2.1           & 5.0           & 0.7           & 0.4           & 4.1           & 4.4           & --            & --            & --            & --            & 5.3           & 3.5           \\
                                                     & \logo{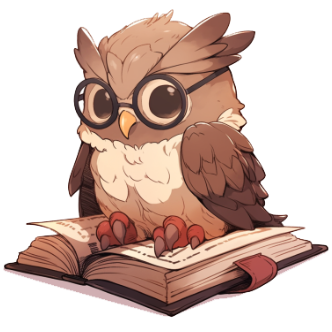}TinyChart (3B)~\cite{TinyChart_2024}                & 2024.04 & 6.1           & 6.3           & 9.7           & 3.2           & 5.7           & 5.4           & 0.5           & 3.4           & 0.2           & 1.3           & 0.7           & 4.2           & --            & --            & --            & --            & 2.9           & 3.0           \\
                                                     & \logo{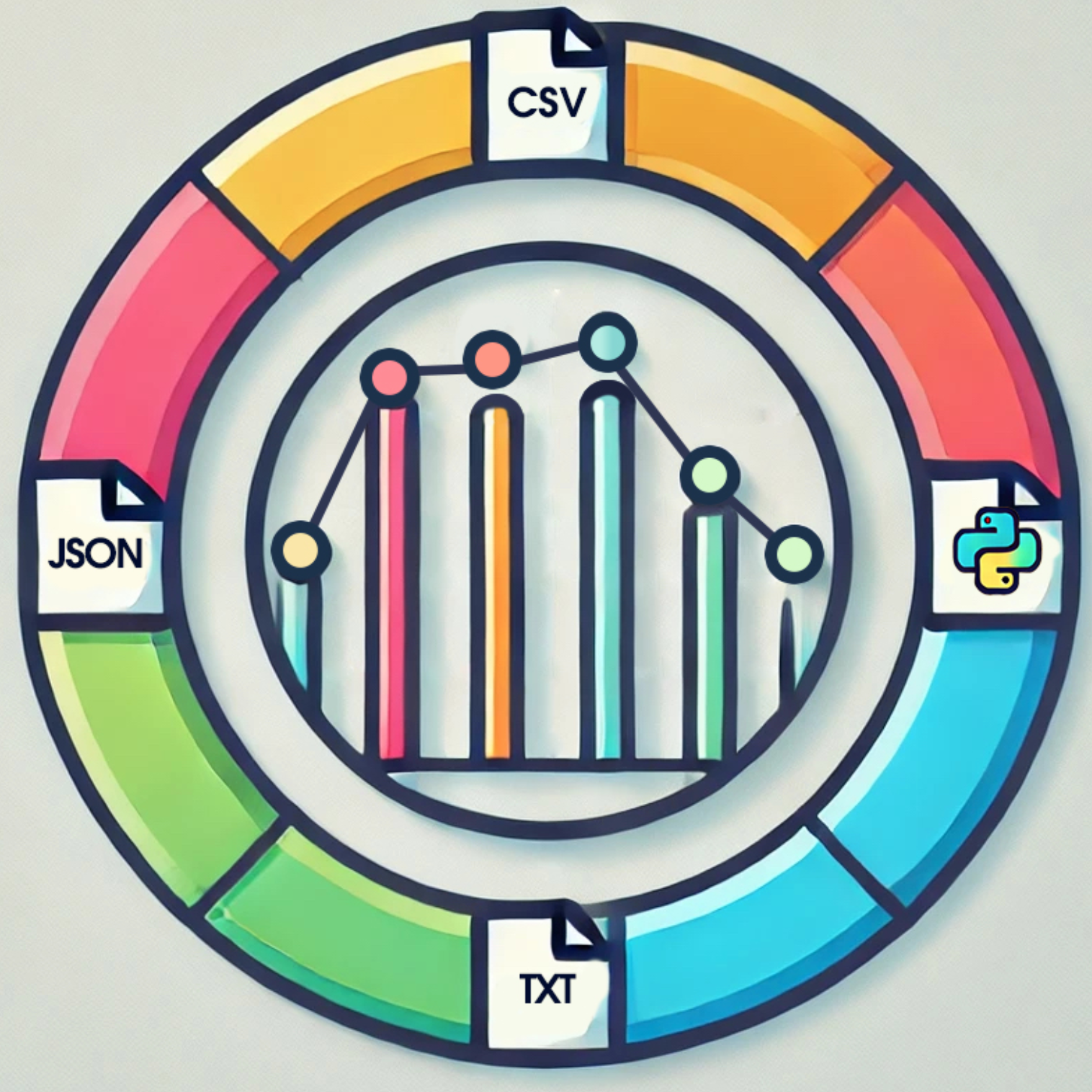}ChartMoE (8B)~\cite{ChartMoE_2024}               & 2024.09 & 18.7          & 24.4          & 14.7          & 22.3          & 15.0          & 48.5          & 3.7           & 16.1          & 2.7           & 1.6           & 5.1           & 19.5          & 4.0           & --            & 4.1           & --            & 8.5           & 16.6          \\
                                                     & \logo{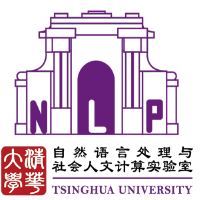}ChartCoder (7B)~\cite{ChartCoder_2025}             & 2025.01 & 23.2          & 12.6          & 22.0          & 19.6          & 34.3          & 16.7          & 5.5           & 13.9          & 5.4           & 11.4          & 3.7           & 5.1           & 5.6           & --            & 1.0           & --            & 12.6          & 9.9           \\
                                                     & \logo{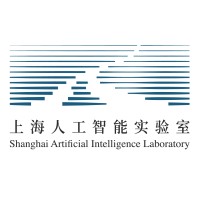}RRVF (7B)~\cite{RRVF_2025}                  & 2025.07 & \ul{35.8}     & \textbf{66.5} & \ul{41.5}     & \ul{54.3}     & \ul{51.6}     & \ul{75.3}     & 16.6          & 40.3          & \ul{14.7}     & \ul{14.1}     & \ul{23.5}     & \textbf{61.2} & \ul{36.4}     & \ul{32.4}     & \textbf{68.4} & \textbf{63.8} & \ul{36.0}     & \ul{51.0}     \\
                                                     & \logo{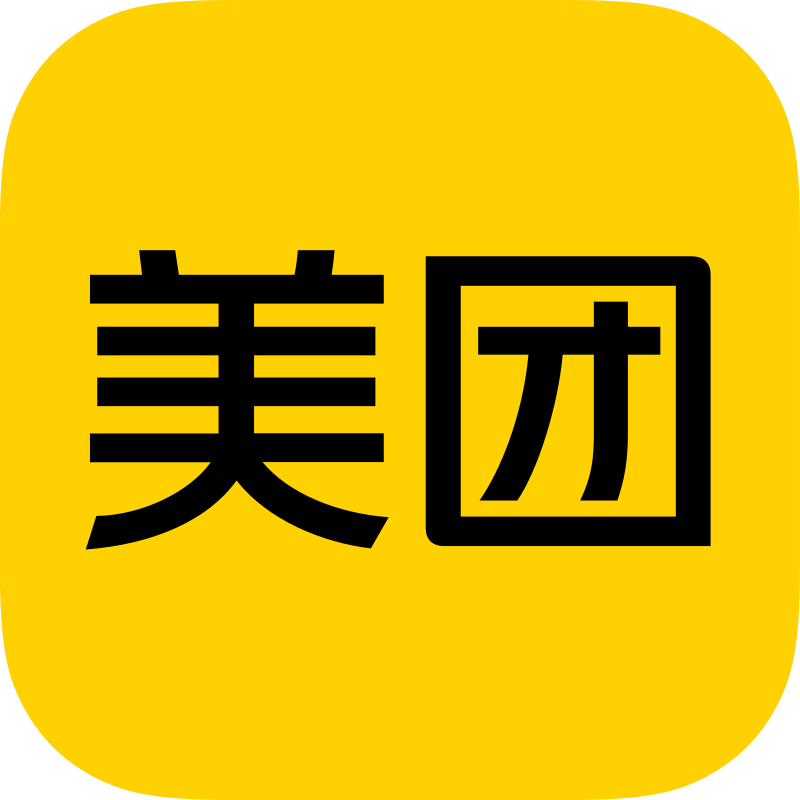}MSRL (7B)~\cite{MSRL_2025}                        & 2025.08 & 32.7          & 45.2          & 35.2          & 34.3          & 41.2          & 67.9          & \ul{25.9}     & \ul{48.0}     & 11.2          & 13.0          & 16.7          & 35.2          & 23.2          & 12.4          & 31.0          & 18.8          & 27.1          & 34.3          \\
        \rowcolor{table_ours}
                                                     & \logo{hunyuan}\textbf{Ours} (1B)                                & 2026.05 & \textbf{43.2} & \ul{63.8}     & \textbf{57.8} & \textbf{73.5} & \textbf{74.3} & \textbf{82.0} & \textbf{28.1} & \textbf{54.7} & \textbf{54.3} & \textbf{59.1} & \textbf{26.3} & \ul{55.1}     & \textbf{50.8} & \textbf{51.3} & \ul{61.4}     & \ul{50.9}     & \textbf{49.5} & \textbf{61.3} \\
        \bottomrule
      \end{tabular}
    }
  }
  \label{tab:ChartArena_main_results}
\end{table*}

\cref{tab:ChartArena_main_results} summarizes the main comparison on \textbf{ChartArena}, where we highlight two observations.

\mypara{Best performance among expert chart understanding models.}
Our parser supports the broadest range of chart types and language settings while remaining the most lightweight model (only 1B parameters) in this category. Existing expert parsers are typically limited to a few numeric chart families or English-only data. Despite using 7--13 times fewer parameters, our model achieves the highest average score (49.5 EN / 61.3 ZH) and outperforms prior expert models on most chart families, including structurally complex cases such as box plots, radar charts, and flowcharts.

\mypara{Competitive against general-purpose MLLMs.}
Compared with general-purpose MLLMs, our 1B parser already outperforms strong 8B baselines, including InternVL3.5-8B~\cite{InternVL3_5_2025} and Qwen3-VL-8B-Instruct~\cite{Qwen3_VL_2025} on the overall average, and is competitive with much larger open-source systems on most chart families.
A noticeable gap remains compared to frontier-scale models such as the open-source Kimi K2.5~\cite{Kimi_K2_5_2026} and the proprietary Gemini 3.1 Pro~\cite{Gemini_3_1_2026}. These models benefit from larger parameter scales and broader world knowledge. The advantage is most evident on the hardest ChartArena slices, which require inferring implicit semantics, units, or domain conventions from context.
This gap also varies by chart type. General-purpose MLLMs struggle on box plots, where richer visual cues must be identified. In contrast, our 1B parser lags on mind maps, which require broader world knowledge. Both models perform poorly on radar charts, highlighting the difficulty of visual estimation.

\subsection{Adaptability to Diverse Output Formats}
\label{subsec:multi_format}

\begin{table*}[t!]
    \centering
    \captionsetup{font={small}}
    \caption{
        \textbf{Adaptability to diverse output formats.}
        Our evaluation framework assesses model outputs across multiple structured formats.
        The left reports results on \textbf{numeric charts}, while the right reports \textbf{flowcharts}.
    }
    \vspace{-0.4em}
    {
        \renewcommand{\arraystretch}{1.0}
        \setlength{\tabcolsep}{4pt}
        \resizebox{\textwidth}{!}{
            \begin{tabular}{l l cccc l cccc}
                \toprule
                \multirow{2}{*}[-0.8mm]{\textbf{Model}}  &
                \multicolumn{5}{c|}
                {\textbf{Numeric Charts}}              &
                \multicolumn{5}{c}
                {\textbf{Flowcharts}}
                \\
                \cmidrule(lr){2-6}
                \cmidrule(lr){7-11}
                                                         &
                Output Format                            &
                EM                                       &
                mAP$_{\text{strict}}$                    &
                mAP$_{\text{slight}}$                    &
                \multicolumn{1}{c|}{mAP$_{\text{high}}$} &
                Output Format                            &
                EM                                       &
                mAP$_{\text{strict}}$                    &
                mAP$_{\text{slight}}$                    &
                mAP$_{\text{high}}$
                \\
                \midrule
                \multirow{5}{*}
                {\makecell{Seed2.0 Pro \nextline (no-think)}}
                                                         & Markdown & 16.3 & 22.0 & 38.7 & 54.9 & Mermaid   & 4.0 & 32.4 & 51.1 & 58.3 \\
                                                         & JSON     & 17.4 & 23.2 & 40.7 & 59.1 & Cytoscape & 5.3 & 35.7 & 55.5 & 62.0 \\
                                                         & CSV      & 14.4 & 20.0 & 37.2 & 55.0 & Diagrams  & 4.0 & 30.4 & 52.5 & 59.8 \\
                                                         & Code     & 17.0 & 22.2 & 37.3 & 53.9 & Graphviz  & 5.0 & 33.9 & 54.5 & 61.7 \\
                                                         & SVG      & 8.0  & 14.0 & 28.0 & 40.0 & PlantUML  & 1.0 & 12.9 & 24.3 & 33.8 \\

                \midrule
                \multirow{5}{*}
                {\makecell{Qwen3.5 \nextline 35B-A3B \nextline (think)}}
                                                         & Markdown & 15.8 & 21.4 & 37.0 & 56.0 & Mermaid   & 3.7 & 28.1 & 48.2 & 57.0 \\
                                                         & JSON     & 5.9  & 9.4  & 23.0 & 46.9 & Cytoscape & 4.7 & 31.0 & 50.6 & 59.5 \\
                                                         & CSV      & 15.0 & 20.0 & 38.5 & 56.6 & Diagrams  & 3.7 & 28.9 & 49.0 & 57.1 \\
                                                         & Code     & 14.3 & 19.3 & 32.2 & 46.2 & Graphviz  & 5.0 & 28.3 & 48.0 & 57.0 \\
                                                         & SVG      & 6.3  & 13.4 & 27.4 & 39.9 & PlantUML  & 0.3 & 9.4  & 16.5 & 29.0 \\

                \bottomrule
            \end{tabular}
        }
    }
    \label{tab:multi_format_eval}
\end{table*}

A central design choice of ChartArena is that the proposed unified evaluation protocol (\cref{subsec:unified_eval_protocol}) is format-agnostic, outputs are normalized into canonical triples/graph views before scoring, and we can evaluate the same model under any of its supported output paradigms.
We use two representative models, the proprietary Seed2.0 Pro~\cite{Seed_2_0_2026} and the open-source Qwen3.5-35B-A3B~\cite{Qwen3_5_2026}, and report their performance under five numeric-chart formats and five flowchart formats in~\cref{tab:multi_format_eval}.

For numeric charts, Seed2.0 Pro is relatively stable across Markdown, JSON, CSV, and Code. Qwen3.5-35B-A3B drops from 56.0 mAP$_{\text{high}}$ in Markdown to 46.9 in JSON and 46.2 in Code. Format-agnostic scoring therefore does not imply format-invariant model performance. SVG is weakest for both, suggesting difficulty recovering exact values from visual representations.

For flowcharts, Mermaid, Cytoscape, Diagrams, and Graphviz are all competent and roughly comparable, whereas PlantUML performs significantly worse: its code-like control statements struggle to encode complex graph topologies (e.g., cycles, multi-source/multi-sink subgraphs), which are frequent in the flowchart slice of ChartArena.

\subsection{Ablation Study}
\label{subsec:ablation}

\cref{tab:ablation_training_stage} presents ablation results for pretraining, SFT, and CAP reasoning, further grouped by visual style.
Pretraining (row (a)) mainly improves digital-rendering samples, with limited gains on printed and hand-drawn charts, reflecting the synthetic nature of the pretraining corpus.
SFT (row (b)) substantially improves robustness on printed and hand-drawn samples for both numeric and diagrammatic charts.
Adding CAP reasoning to either numeric or diagrammatic alone (rows (c)--(d)) consistently improves the corresponding subset without degrading the other. Enabling CAP on both subsets (row (e)) achieves the best overall performance and is used in~\cref{tab:ChartArena_main_results}.

\begin{table}[h!]
    \centering
    \captionsetup{font={small}}
    \caption{
        \textbf{Ablation study on training strategies.}
        We analyze the impact of pretraining and different SFT settings with or without CAP reasoning on numeric charts (1,800 samples)
        and diagrammatic charts (600 samples).
    }
    {
        \renewcommand{\arraystretch}{1.0}
        \setlength{\tabcolsep}{4pt}
        \resizebox{\columnwidth}{!}{
            \begin{tabular}{l c cc ccc ccc}
                \toprule
                \multirow{2}{*}[-0.8mm]{\textbf{Setting}}              &
                \multirow{2}{*}[-0.8mm]{\textbf{Pretrain}}             &
                \multicolumn{2}{c}
                {\textbf{SFT}}                                         &
                \multicolumn{3}{c}
                {\textbf{Numeric Charts}}                              &
                \multicolumn{3}{c}
                {\textbf{Diagrammatic Charts}}
                \\
                \cmidrule(lr){3-4}
                \cmidrule(lr){5-7}
                \cmidrule(lr){8-10}
                                                                       &
                                                                       &
                Numeric                                                &
                Flow                                                   &
                \makecell{Digital}                                     &
                \makecell{Printed}                                     &
                \makecell{Drawn}                                       &
                \makecell{Digital}                                     &
                \makecell{Printed}                                     &
                \makecell{Drawn}
                \\
                \midrule

                MLLM Backbone                                          & --         & --     & --     & 56.3 & 49.7 & 37.4 & 55.0 & 50.4 & 26.0 \\

                (a) Pretrain                                           & \checkmark & --     & --     & 62.5 & 50.2 & 38.1 & 58.9 & 51.8 & 27.7 \\

                (b) + SFT (w/o CAP Reasoning)                          & \checkmark & direct & direct & 63.1 & 57.8 & 44.9 & 59.6 & 55.8 & 37.9 \\

                (c) + SFT (Numeric w/ CAP Reasoning, Diagrammatic w/o) & \checkmark & CAP    & direct & 64.2 & 60.8 & 47.8 & 60.1 & 56.3 & 38.5 \\

                (d) + SFT (Numeric w/o CAP Reasoning, Diagrammatic w)  & \checkmark & direct & CAP    & 62.9 & 58.2 & 44.3 & 62.3 & 57.9 & 41.2 \\
                \rowcolor{table_ours}
                (e) + SFT (w CAP Reasoning)                            & \checkmark & CAP    & CAP    & 64.1 & 60.4 & 48.6 & 62.5 & 58.1 & 40.9 \\
                \bottomrule
            \end{tabular}
        }
    }
    \vspace{-1.0em}
    \label{tab:ablation_training_stage}
\end{table}

\begin{figure}[h!]
    \centering
    \includegraphics[width=\linewidth]{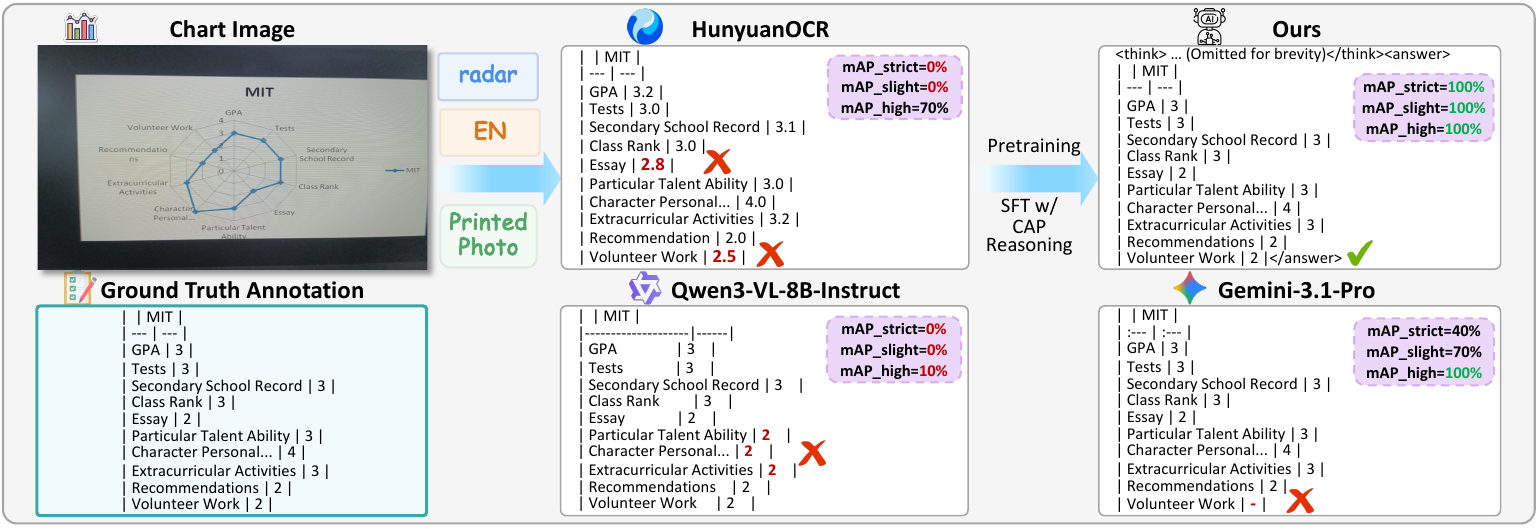}
    \vspace{-1.8em}
    \caption{
        \textbf{Qualitative Comparisons on ChartArena.}
        Photograph samples are challenging due to visual noise. CAP training enables more accurate and complete recovery of structure and values.
    }
    \label{fig:single_case}
    \vspace{-1.6em}
\end{figure}

\subsection{Qualitative Result}
\label{subsec:qualitative_result}

As shown in~\cref{fig:single_case}, photograph-based charts are challenging due to visual distortions. After CAP training, our model recovers all values correctly and achieves the best scores. Gemini performs relatively well overall, but replaces uncertain entries with ``--'', leading to incomplete predictions. Further case studies are provided in~\cref{supp:sec:further_case_study}.
\section{Conclusion}
\label{sec:conclusion}

We introduced ChartArena, a unified benchmark and format-agnostic evaluation framework for general chart parsing across eight numeric and diagrammatic chart families. By covering bilingual content, real-world visual scenarios, and heterogeneous output formats, ChartArena reveals the limitations of existing parsers under realistic structural and visual variation. We further proposed CAP-based reasoning, which injects explicit structural priors into chart parsers and consistently improves robustness over direct generation baselines. Together, ChartArena and CAP provide a unified foundation for evaluating and building robust general-purpose chart parsers.

\mypara{Limitations.}
Despite the broad coverage of ChartArena, two limitations remain. First, the current benchmark focuses on single-page chart images and does not yet cover multi-page charts, where the parser must aggregate visual elements, legends, or continuation tables across pages. Second, ChartArena does not include certain chart families that are particularly challenging for chart parsing models, such as scatter plots. These charts often require precise point localization and dense coordinate recovery, which introduce difficulties beyond the structural extraction tasks considered in this work. We leave these directions to future extensions of the benchmark.

\clearpage
{
    \small
    \bibliographystyle{unsrtnat}
    \bibliography{main}
}

\clearpage
\appendix
\saferesetlinenumber[1]
\counterwithin{figure}{section}
\counterwithin{table}{section}
\maketitlesupplementary
\setcounter{page}{1}

\section*{Overview}
This material provides supplementary details to the main paper, including the following sections:
\vspace{-0.5em}
\begin{itemize}[
        label=\raisebox{0.5ex}{\tiny$\bullet$},
        leftmargin=1em,
        itemsep=0pt, 
        parsep=2pt, 
        partopsep=0pt 
    ]
    \item (\ref{supp:sec:benchmark_details}) \textbf{Benchmark Details}
          \begin{itemize}[label=-]
              \item (\ref{supp:subsec:image_sources}) Image Sources and Scenarios
              \item (\ref{supp:subsec:annotation_protocol}) Annotation Protocol and Human Effort
              \item (\ref{supp:subsec:data_statistics}) Fine-grained Data Statistics
              \item (\ref{supp:subsec:benchmark_samples}) Benchmark Samples
          \end{itemize}
    \item (\ref{supp:sec:training_details}) \textbf{Training Details}
          \begin{itemize}[label=-]
              \item (\ref{supp:subsec:training_setup}) Training Setup
              \item (\ref{supp:subsec:training_data}) Training Data
          \end{itemize}
    \item (\ref{supp:sec:cap_analysis}) \textbf{Further Analysis of CAP}
          \begin{itemize}[label=-]
              \item (\ref{supp:subsec:cap_verification}) Human Verification of CAP Targets
              \item (\ref{supp:subsec:cap_isolation}) Isolating the Contribution of CAP
              \item (\ref{supp:subsec:data_attribution}) Attribution of Real-World Robustness to Training Data
              \item (\ref{supp:subsec:cap_scaling}) Scaling CAP to a Larger Backbone
          \end{itemize}
    \item (\ref{supp:sec:evaluation_details}) \textbf{Evaluation Details}
          \begin{itemize}[label=-]
              \item (\ref{supp:subsec:eval_routing}) Format-Agnostic Routing and Normalization
              \item (\ref{supp:subsec:triple_scoring}) Triple-Based Scoring for Numeric Charts
              \item (\ref{supp:subsec:graph_scoring}) Graph- and Tree-Based Scoring for Diagrammatic Charts
              \item (\ref{supp:subsec:aggregation}) Aggregation into Exact Match and mAP
              \item (\ref{supp:subsec:evaluation_setup}) Evaluation Setup
          \end{itemize}
    \item (\ref{supp:sec:further_evaluation_results}) \textbf{Further Evaluation Results}
          \begin{itemize}[label=-]
              \item (\ref{supp:subsec:appendix_visual_scenarios}) Detailed Analysis under Different Visual Scenarios
              \item (\ref{supp:subsec:language_gap}) Analysis of the English--Chinese Performance Gap
              \item (\ref{supp:subsec:hard_samples}) Fine-grained Analysis on Hard Samples
              \item (\ref{supp:subsec:metric_consistency}) Consistency Across Evaluation Metrics
              \item (\ref{supp:subsec:weight_sensitivity}) Node/Edge Weight Sensitivity in Flowchart Scoring
              \item (\ref{supp:subsec:bootstrap}) Statistical Significance
              \item (\ref{supp:subsec:additional_results}) Additional Model Results
          \end{itemize}
    \item (\ref{supp:sec:further_related_work}) \textbf{Further Related Work}
    \item (\ref{supp:sec:broader_impact}) \textbf{Broader Impact}
    \item (\ref{supp:sec:llm_usage}) \textbf{LLM Usage Statement}
    \item (\ref{supp:sec:further_case_study}) \textbf{Further Case Study}
\end{itemize}

\clearpage
\section{Benchmark Details}
\label{supp:sec:benchmark_details}
\vspace{-0.5em}

This section complements~\cref{subsec:benchmark_construction} with additional details on how ChartArena is collected and annotated. We describe the image sources under three real-world scenarios  in~\cref{supp:subsec:image_sources}, the human--agent collaborative annotation protocol and the corresponding human-effort budget in~\cref{supp:subsec:annotation_protocol}.

\subsection{Image Sources and Scenarios}
\label{supp:subsec:image_sources}
\vspace{-0.5em}

As outlined in~\cref{subsec:benchmark_construction}, the design goal of ChartArena is to cover as broad a distribution of real-world chart images as possible, so that the benchmark presents a more challenging test bed for chart parsing models rather than favoring synthetic or templated image inputs.
Images are primarily collected through web image search. When certain categories are under-represented, we further supplement the corpus by asking in-house annotators to \emph{scan, photograph, or hand-draw} additional samples so that each category reaches the target size and the benchmark coverage remains complete.

All collected images must satisfy the following quality requirements: (i) all textual content must be human-readable; (ii) the chart must reflect \emph{real-world data} rather than template-style placeholder data; and (iii) the image must be a single complete chart without heavy occlusion or cropping.

We organize the benchmark around three complementary visual scenarios, detailed below.

\begin{itemize}[
        label=\raisebox{0.4ex}{\tiny$\bullet$},
        leftmargin=1em,
        itemsep=2pt,
        parsep=0pt,
        topsep=0pt,
        partopsep=0pt
    ]
    \item \textbf{Digital rendering.} Digitally rendered charts taken directly from real-world documents, without any physical-capture distortion. Existing chart parsing benchmarks may be dominated by synthetically rendered images~\cite{ChartX_ChartVLM_2025}, and seldom stress-test chart parsers on the complex digital-native charts that appear in practice. To close this gap, we harvest charts from several sources such as academic papers, product launches, industry reports, financial disclosures, and analytical whitepapers, capturing the diversity of layouts, legends, and annotations found in real digital-native materials.
    \item \textbf{Printed photo.} Charts captured by camera from a printed page or an electronic screen. We first try to collect web-native photographs of printed charts; when such samples are scarce, annotators print or screen-display collected documents and re-capture them with a camera. We explicitly require the capture to reproduce real-world interference such as uneven illumination, reflections, and moir\'e patterns on screens. Annotators are instructed to keep all text visually legible, to focus on the chart region, and are allowed to introduce moderate perspective tilt to reflect casual handheld capture.
    \item \textbf{Hand-drawn photo.} Charts that are hand-drawn and photographed. On top of the real-world lighting conditions of the previous scenario, this category additionally introduces handwriting-specific variations: irregular fonts, non-uniform layout, strike-throughs, and edits. It is by far the hardest and most long-tailed scenario to collect. We first crawl high-quality handwritten charts that meet our quality bar via image search; for the remaining, annotators \emph{re-draw} a subset of the complex digital-rendering and printed-photographed charts found during search, so that different annotators contribute diverse handwriting styles onto the same underlying chart semantics.
\end{itemize}

\begin{my_template_box}[breakable]{CAP Extraction Prompt Template}
    \small
    \setlist[itemize]{
        label=\raisebox{0.5ex}{\tiny$\bullet$},
        leftmargin=1.2em,
        itemsep=2pt,
        parsep=0pt,
        topsep=2pt,
        partopsep=0pt
    }
    \setlist[enumerate]{
        leftmargin=1em,
        itemsep=2pt,
        parsep=0pt,
        topsep=2pt,
        partopsep=0pt
    }
    \texttt{<|im\_start|>user} \\
    \texttt{<image>} \\
    \texttt{Based on the chart image and markdown annotation, extract the chart's atomic semantic information.}
    \vspace{2pt}
    Only extract structural semantics. No trend analysis, no data reconstruction.
    \vspace{4pt}
    \textbf{[Input Information]}
    \begin{itemize}
        \item Chart image: title, axis labels, tick ranges
        \item Markdown: groups, series, fields
    \end{itemize}
    \vspace{4pt}
    \textbf{[Extraction Targets]}
    \begin{enumerate}
        \item \textbf{chart\_title}
              \begin{itemize}
                  \item Main title
              \end{itemize}

        \item \textbf{comparison\_items}
              \begin{itemize}
                  \item Series or group names
                  \item Prefer markdown:
                        \begin{itemize}
                            \item Multi-column $\rightarrow$ column names (except first)
                            \item Single metric $\rightarrow$ first column (groups)
                        \end{itemize}
                  \item No numerical values
              \end{itemize}

        \item \textbf{x\_axis}
              \begin{itemize}
                  \item semantic (e.g., year, category)
                  \item tick\_range: type, min/max, values
              \end{itemize}

        \item \textbf{y\_axis}
              \begin{itemize}
                  \item semantic (e.g., value, index)
                  \item tick\_range: type, min/max, values
              \end{itemize}

        \item \textbf{statistics\_fields}
              \begin{itemize}
                  \item If present: min, Q1, median, Q3, max, outliers
                  \item Else: null
              \end{itemize}

        \item \textbf{caption}
              \begin{itemize}
                  \item Include additional global descriptive text if present
              \end{itemize}
    \end{enumerate}
    \vspace{4pt}
    \textbf{[Output Format]} \\
    \texttt{\{} \\
    \texttt{"chart\_title": string | null,} \\
    \texttt{"comparison\_items": [string],} \\
    \texttt{"x\_axis": \{ "semantic": string | null, "tick\_range": \{...\} \},} \\
    \texttt{"y\_axis": \{ "semantic": string | null, "tick\_range": \{...\} \},} \\
    \texttt{"statistics\_fields": [string] | null,} \\
    \texttt{"caption": string | null} \\
    \texttt{\}}

    \vspace{4pt}
    \textbf{[Input]} \\
    \texttt{markdown: \{markdown\}} \\
    \texttt{<|im\_end|>} \\
    \texttt{<|im\_start|>assistant}

    \label{prompt:chart_2D_extract}
\end{my_template_box}

\subsection{Annotation Protocol and Human Effort}
\label{supp:subsec:annotation_protocol}
\vspace{-0.5em}

We rely on a human--agent collaborative annotation pipeline with multi-round verification to guarantee high-quality ground truth of ChartArena. In the pre-annotation stage, multiple MLLMs independently generate candidate annotations, which are then cross-checked and merged into a higher-quality draft before human refinement. Human annotators subsequently verify and correct both the structural semantics and fine-grained content details. To further improve annotation reliability, two complementary cross-checking mechanisms are applied throughout: (i) \emph{value-level cross-check}, where uncertain numeric entries identified during annotation are independently re-labeled by multiple annotators; and (ii) \emph{batch-level review}, where, after one full annotation pass, a separate reviewer performs a unified quality check over the whole batch.

The pipeline organizes the $2{,}400$ images into two annotation streams according to the target format: a Markdown stream that covers the $1{,}800$ numeric charts together with the $300$ mind maps, and a Mermaid stream that covers the $300$ flowcharts. We describe the two streams in turn.

\mypara{Numeric charts and mind maps.}
For these chart types, the ground truth is serialized as Markdown: numeric charts use a Markdown table, while mind maps use a nested Markdown unordered list. We define chart-type-specific annotation guidelines (column ordering, unit normalization, nesting rules, etc.) that annotators must strictly follow.
During annotation, annotators use auxiliary aids such as reference grids and guidelines to read off values accurately. Since hand-drawn charts and low-resolution photographs often contain values that cannot be read unambiguously, multiple annotators \emph{cross-check} all uncertain cells before finalizing the label.

\mypara{Flowcharts.}
For flowcharts, the ground truth is serialized as Mermaid code. After annotation, annotators are required to render the produced Mermaid code with an online visualization tool and compare the rendered diagram against the source image, verifying that (i) the set of nodes, (ii) the connectivity of edges, and (iii) the overall logical flow all match. Flowchart annotation is substantially more demanding than the numeric stream, both because of the richer topology and because a single misrouted edge can break the semantics of the whole diagram.

\mypara{Human-effort budget.}
The resulting per-image time cost and total human-effort budget (in person-days, $8$ working hours per day) for both annotation and quality review are summarized in~\cref{tab:annotation_effort}.

\begin{table}[t]
    \centering
    \captionsetup{font={small}}
    \caption{
        \textbf{Human-effort budget for ChartArena annotation.}
        Rows are grouped by \emph{annotation stream}: numeric charts and mind maps share a Markdown-based labeling workflow, whereas flowcharts use Mermaid-based workflow. Together, they comprise the $2{,}400$-image benchmark.
    }
    \label{tab:annotation_effort}
    {
        \setlength{\tabcolsep}{10pt}
        \renewcommand{\arraystretch}{0.98}
        \resizebox{\columnwidth}{!}{%
            \begin{tabular}{l c c c c c}
                \toprule
                \multirow{2}{*}[-0.8mm]
                {\textbf{Annotation Stream}}            &
                \multirow{2}{*}[-0.8mm]
                {\textbf{\#Images}}                     &
                \multicolumn{2}{c}{\textbf{Annotation}} &
                \multicolumn{2}{c}{\textbf{Quality Review}}
                \\
                \cmidrule(lr){3-4} \cmidrule(lr){5-6}
                                                        &           & Per image (min) & Total (person-day) & Per image (min) & Total (person-day) \\
                \midrule
                Markdown stream                         & $2{,}100$ & $19.7$          & $86.2$             & $8.9$           & $38.9$             \\
                Mermaid stream                          & $300$     & $46.3$          & $28.9$             & $17.1$          & $10.7$             \\
                \midrule
                \textbf{Total}                          & $2{,}400$ & --              & $115.1$            & --              & $49.6$             \\
                \bottomrule
            \end{tabular}
        }
    }
    \vspace{-1.4em}
\end{table}

\subsection{Fine-grained Data Statistics}
\label{supp:subsec:data_statistics}
\vspace{-0.5em}

\begin{figure*}[t!]
    \centering
    \includegraphics[width=\textwidth]{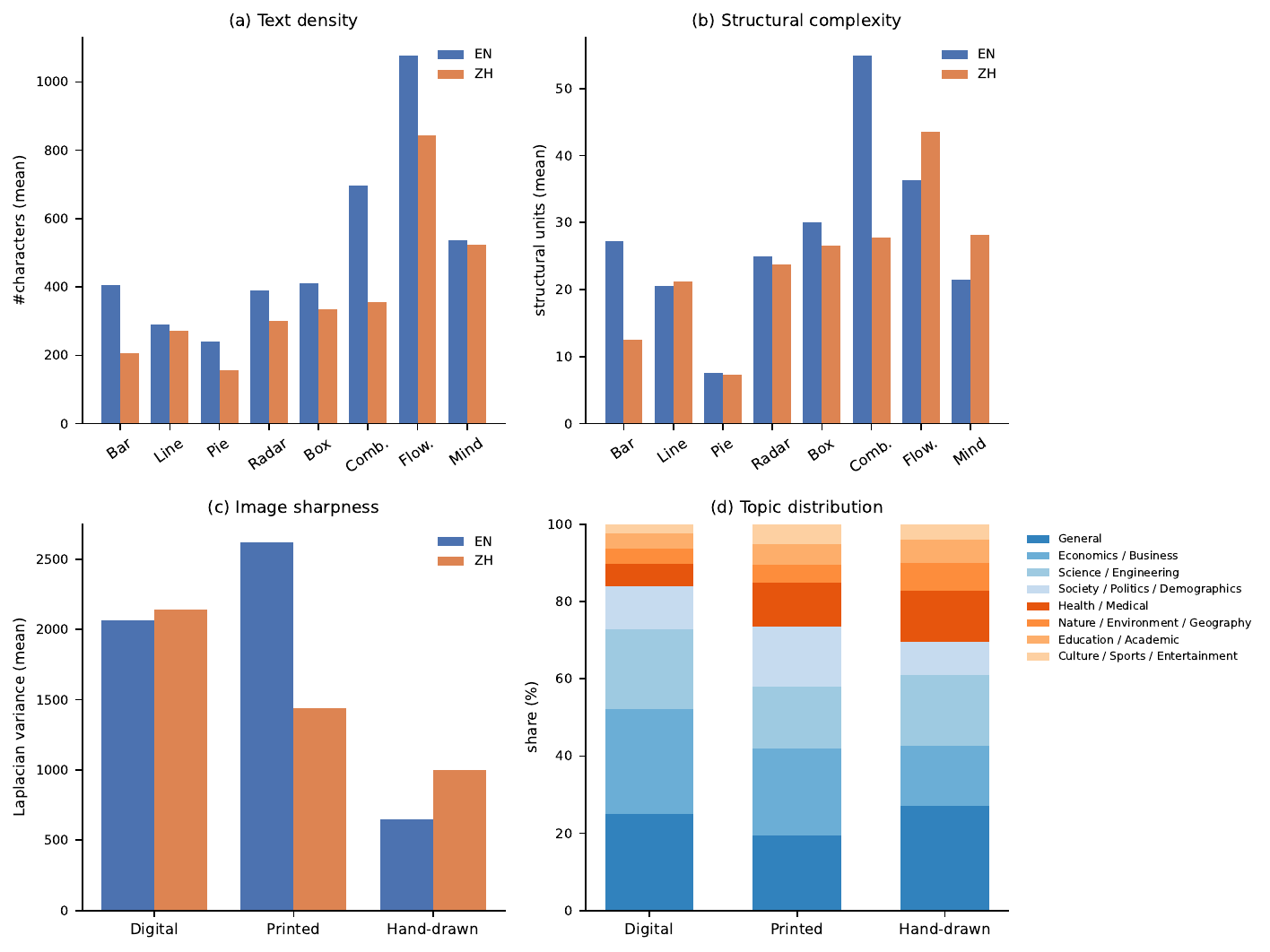}
    \vspace{-1.5em}
    \caption{
        \textbf{Fine-grained data distributions of ChartArena.}
        (a) Mean number of ground-truth characters per chart type and language. (b) Mean structural complexity per chart type and language, measured in triples, mind-map nodes, or flowchart nodes plus edges. (c) Mean Laplacian variance per visual scenario and language. (d) Topic distribution per visual scenario.
    }
    \label{fig:data_distribution}
\end{figure*}

ChartArena is balanced by chart type, visual scenario, and language by design. To characterize the benchmark beyond these three axes and provide meaningful comparisons, we report four objectively computable dimensions, each crossed with language or visual scenario, in~\cref{tab:data_distribution} and~\cref{fig:data_distribution}.

\mypara{Text density.}
We count the characters in each ground truth sample. Chinese and English samples are comparable in some chart types. For instance, $523$ versus $537$ characters on mind maps and $272$ versus $290$ on line charts.

\mypara{Structural complexity.}
We measure complexity with the same units as the evaluation backends, namely the number of triples for numeric charts, nodes for mind maps, and nodes plus edges for flowcharts. Averaged over chart families, the Chinese subset is simpler on numeric charts ($20.2\%$ fewer triples) and more complex on diagrammatic charts ($24.6\%$ more structural units).

\mypara{Image quality.}
We use two proxies, the compression ratio in bytes per pixel and the Laplacian variance as a sharpness measure. Averaged over both languages, the Laplacian variance decreases from $2106$ on digital renderings to $2024$ on printed photos and drops sharply to $827$ on hand-drawn photos. The English hand-drawn subset is the least sharp ($649$) even though it has the highest average resolution ($1.95$ megapixels), which shows that visual difficulty is driven mainly by the capture channel rather than by surface resolution or language.

\mypara{Topics.}
We assign each chart to one of seven named topics or a general bucket. The shares are general ($23.9\%$), economics and business ($21.7\%$), science and engineering ($18.4\%$), society, politics, and demographics ($11.7\%$), health and medicine ($10.2\%$), nature, environment, and geography ($5.2\%$), education and academia ($5.1\%$), and culture, sports, and entertainment ($3.8\%$). The distribution across visual scenarios is shown in~\cref{fig:data_distribution}(d).

\mypara{Coverage of difficult real-world charts.}
Images are screened with the quality requirements of~\cref{supp:subsec:image_sources}: every text must be human-readable, the chart must carry real-world data, and the image must be a single complete chart without heavy occlusion or cropping. Images that fail these requirements are discarded because their ground truth cannot be recovered reliably by human annotators. Difficult samples on which annotators can still reach agreement, such as low-contrast photographs or irregular handwriting, are kept, and uncertain values in them are resolved by the multi-annotator cross-check described above.

\begin{table}[t!]
    \centering
    \captionsetup{font={small}}
    \caption{
        \textbf{Fine-grained statistics of ChartArena.}
        \emph{Left}: per chart type and language, text density is mean ground truth characters, and structural complexity is mean triples (numeric charts), nodes (mind maps), and nodes plus edges (flowcharts).
        \emph{Right}: per visual scenario and language, image-quality proxies are mean resolution (megapixels), compression ratio (bytes/pixel), and Laplacian variance (higher = sharper).
    }
    \label{tab:data_distribution}
    {
        \renewcommand{\arraystretch}{1.0}
        \setlength{\tabcolsep}{4pt}
        \resizebox{\textwidth}{!}{
            \begin{tabular}{l cc cc c l cc cc cc}
                \toprule
                \multirow{2}{*}[-0.8mm]{\textbf{Chart type}} & \multicolumn{2}{c}{\textbf{Text density}} & \multicolumn{2}{c}{\textbf{Complexity}}    &                                        &
                \multirow{2}{*}[-0.8mm]{\textbf{Scenario}}   & \multicolumn{2}{c}{\textbf{Resolution}}   & \multicolumn{2}{c}{\textbf{Bytes / pixel}} & \multicolumn{2}{c}{\textbf{Sharpness}}                                                                         \\
                \cmidrule(lr){2-3} \cmidrule(lr){4-5} \cmidrule(lr){8-9} \cmidrule(lr){10-11} \cmidrule(lr){12-13}
                                                             & EN                                        & ZH                                         & EN                                     & ZH   &  &                   & EN   & ZH   & EN   & ZH   & EN   & ZH   \\
                \cmidrule(lr){1-5} \cmidrule(lr){7-13}
                Bar                                          & 404.9                                     & 207.4                                      & 27.3                                   & 12.6 &  & Digital rendering & 0.56 & 0.61 & 0.50 & 0.50 & 2067 & 2144 \\
                Line                                         & 290.0                                     & 271.8                                      & 20.5                                   & 21.2 &  & Printed photo     & 1.25 & 0.79 & 0.94 & 0.96 & 2617 & 1439 \\
                Pie                                          & 241.4                                     & 157.1                                      & 7.6                                    & 7.3  &  & Hand-drawn photo  & 1.95 & 0.86 & 0.56 & 0.92 & 649  & 1001 \\
                Radar                                        & 389.5                                     & 301.4                                      & 25.0                                   & 23.7 &  &                   &      &      &      &      &      &      \\
                Box plot                                     & 411.2                                     & 334.7                                      & 30.1                                   & 26.5 &  &                   &      &      &      &      &      &      \\
                Comb. chart                                  & 695.5                                     & 356.6                                      & 54.9                                   & 27.7 &  &                   &      &      &      &      &      &      \\
                Flowchart                                    & 1076.4                                    & 844.9                                      & 36.3                                   & 43.5 &  &                   &      &      &      &      &      &      \\
                Mind map                                     & 537.0                                     & 522.8                                      & 21.5                                   & 28.2 &  &                   &      &      &      &      &      &      \\
                \bottomrule
            \end{tabular}
        }
    }
    \vspace{-0.6em}
\end{table}

\subsection{Benchmark Samples}
\label{supp:subsec:benchmark_samples}

To make the three scenarios introduced in~\cref{supp:subsec:image_sources} more concrete, we show several representative samples in ChartArena. Each sample consists of the original image and its ground truth annotation.

\newtcolorbox{sample_anno_box}[1]{%
    enhanced,
    colback=gray!5, colframe=gray!55,
    boxrule=0.6pt, arc=1.5mm, boxsep=2.5pt,
    left=5pt, right=5pt, top=3pt, bottom=3pt,
    fontupper=\scriptsize\ttfamily,
    title={\normalfont\footnotesize\bfseries GT annotation -- #1},
    colbacktitle=gray!55, coltitle=white,
    fonttitle=\footnotesize\bfseries,
    valign=top,
}

\newcommand{\sampleimg}[2][4.0cm]{%
    \adjustbox{valign=t, minipage={\linewidth}, center}{%
        \includegraphics[width=\linewidth, height=#1, keepaspectratio]{#2}%
    }%
}

\begin{figure}[h]
    \centering
    \begin{minipage}[b]{0.42\textwidth}
        \vspace{0pt}%
        \sampleimg[3.8cm]{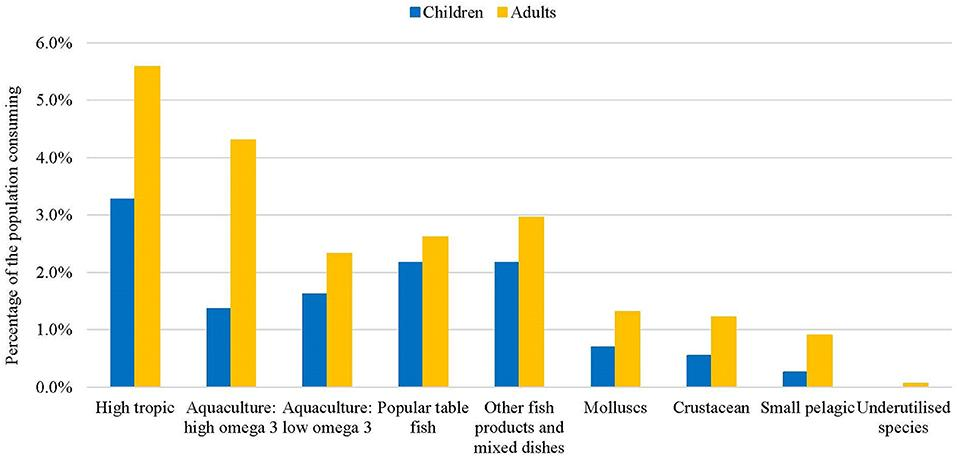}
        \par\vspace{0.3em}{\small (a) \textbf{Bar chart} -- digital.}
    \end{minipage}\hfill
    \begin{minipage}[b]{0.56\textwidth}
        \vspace{0pt}%
        \begin{sample_anno_box}{Markdown table}
            |                           | Children-Percentage | Adults-Percentage |\\
            | ---                       | --- | --- |\\
            | High tropic               | 3.3\% | 5.6\% |\\
            | Aquaculture: high omega 3 | 1.4\% | 4.3\% |\\
            | Aquaculture: low omega 3  | 1.6\% | 2.4\% |\\
            | Popular table fish        | 2.2\% | 2.6\% |\\
            | Other fish | mixed dishes | 2.2\% | 3.0\% |\\
            | Molluscs                  | 0.7\% | 1.3\% |\\
            | Crustacean                | 0.6\% | 1.2\% |\\
            | Small pelagic             | 0.3\% | 0.9\% |\\
            | Underutilised species     |       | 0.1\% |
        \end{sample_anno_box}
    \end{minipage}
    \caption{Representative sample: multi-series bar chart.}
    \label{fig:sample_bar}
\end{figure}

\begin{figure}[h]
    \centering
    \begin{minipage}[b]{0.44\textwidth}
        \vspace{0pt}%
        \sampleimg[3.8cm]{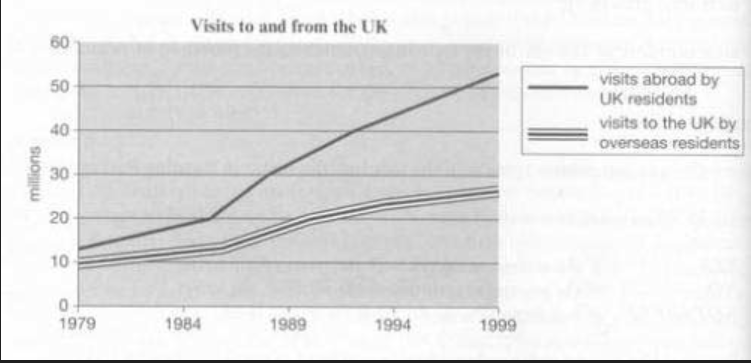}
        \par\vspace{0.3em}{\small (b) \textbf{Line chart} -- photo (printed).}
    \end{minipage}\hfill
    \begin{minipage}[b]{0.54\textwidth}
        \vspace{0pt}%
        \begin{sample_anno_box}{Markdown table}
            \#\#\# Visits to and from the UK\\
            |      | visits abroad by UK residents-millions | visits to the UK by overseas residents-millions |\\
            | ---  | --- | --- |\\
            | 1979 | 13.5 | 10   |\\
            | 1984 | 18.5 | 12.5 |\\
            | 1989 | 32.5 | 18.5 |\\
            | 1994 | 43.5 | 24   |\\
            | 1999 | 53   | 26.5 |
        \end{sample_anno_box}
    \end{minipage}
    \caption{Representative sample: line chart.}
    \label{fig:sample_line}
\end{figure}

\begin{figure}[t]
    \centering
    \begin{minipage}[b]{0.38\textwidth}
        \vspace{0pt}%
        \sampleimg[4.6cm]{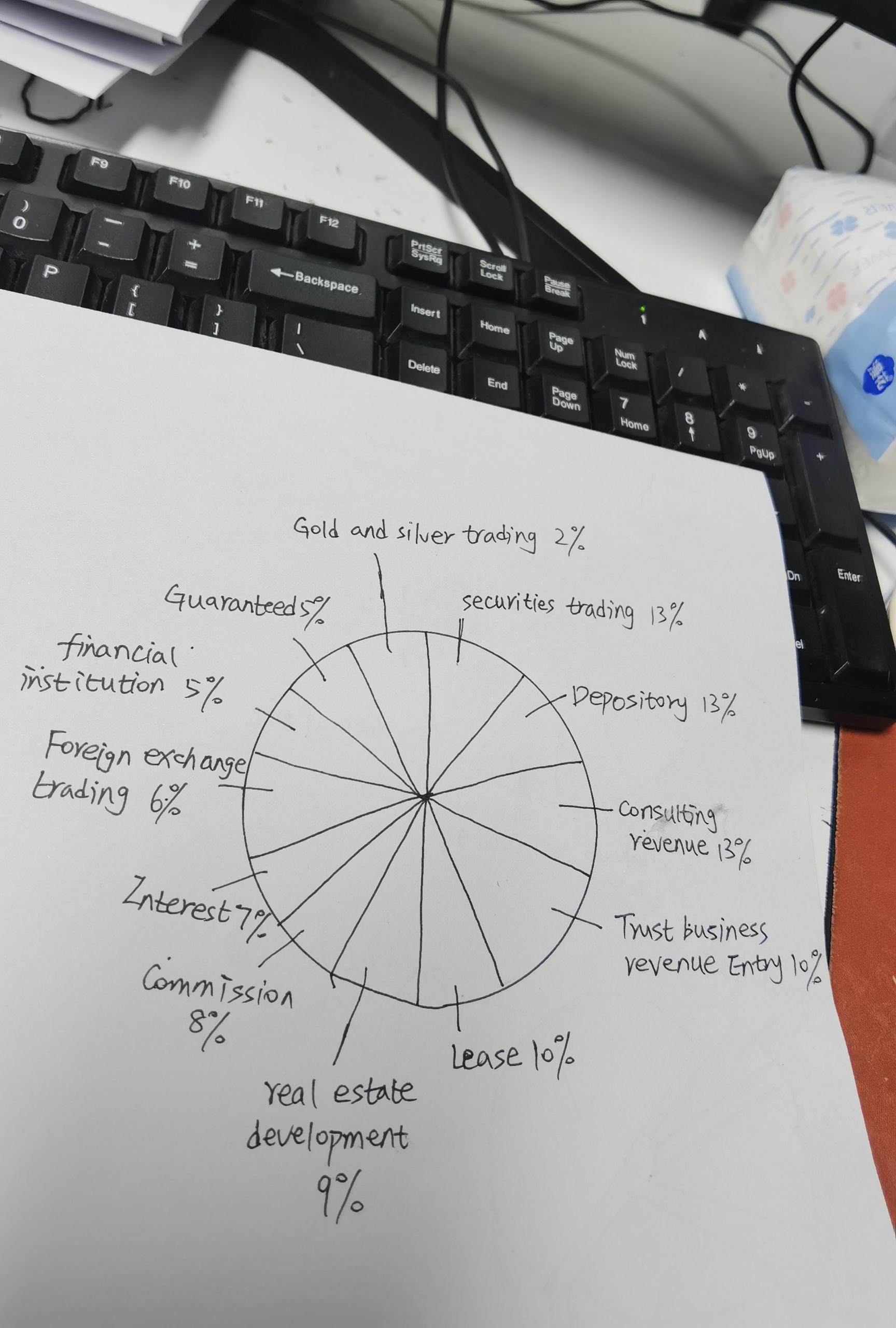}
        \par\vspace{0.3em}{\small (c) \textbf{Pie chart} -- hand-drawn.}
    \end{minipage}\hfill
    \begin{minipage}[b]{0.60\textwidth}
        \vspace{0pt}%
        \begin{sample_anno_box}{Markdown table}
            |                            |     |\\
            | ---                        | --- |\\
            | securities trading         | 13\% |\\
            | Depository                 | 13\% |\\
            | consulting revenue         | 13\% |\\
            | Trust business revenue Entry | 10\% |\\
            | Lease                      | 10\% |\\
            | real estate development    | 9\%  |\\
            | commission                 | 8\%  |\\
            | Interest                   | 7\%  |\\
            | Foreign exchange trading   | 6\%  |\\
            | financial institution      | 5\%  |\\
            | Guaranteed                 | 5\%  |\\
            | Gold and silver trading    | 2\%  |
        \end{sample_anno_box}
    \end{minipage}
    \caption{Representative sample: hand-drawn pie chart with 12 labelled slices.}
    \label{fig:sample_pie}
\end{figure}

\begin{figure}[t]
    \centering
    \begin{minipage}[b]{0.38\textwidth}
        \vspace{0pt}%
        \sampleimg[4.6cm]{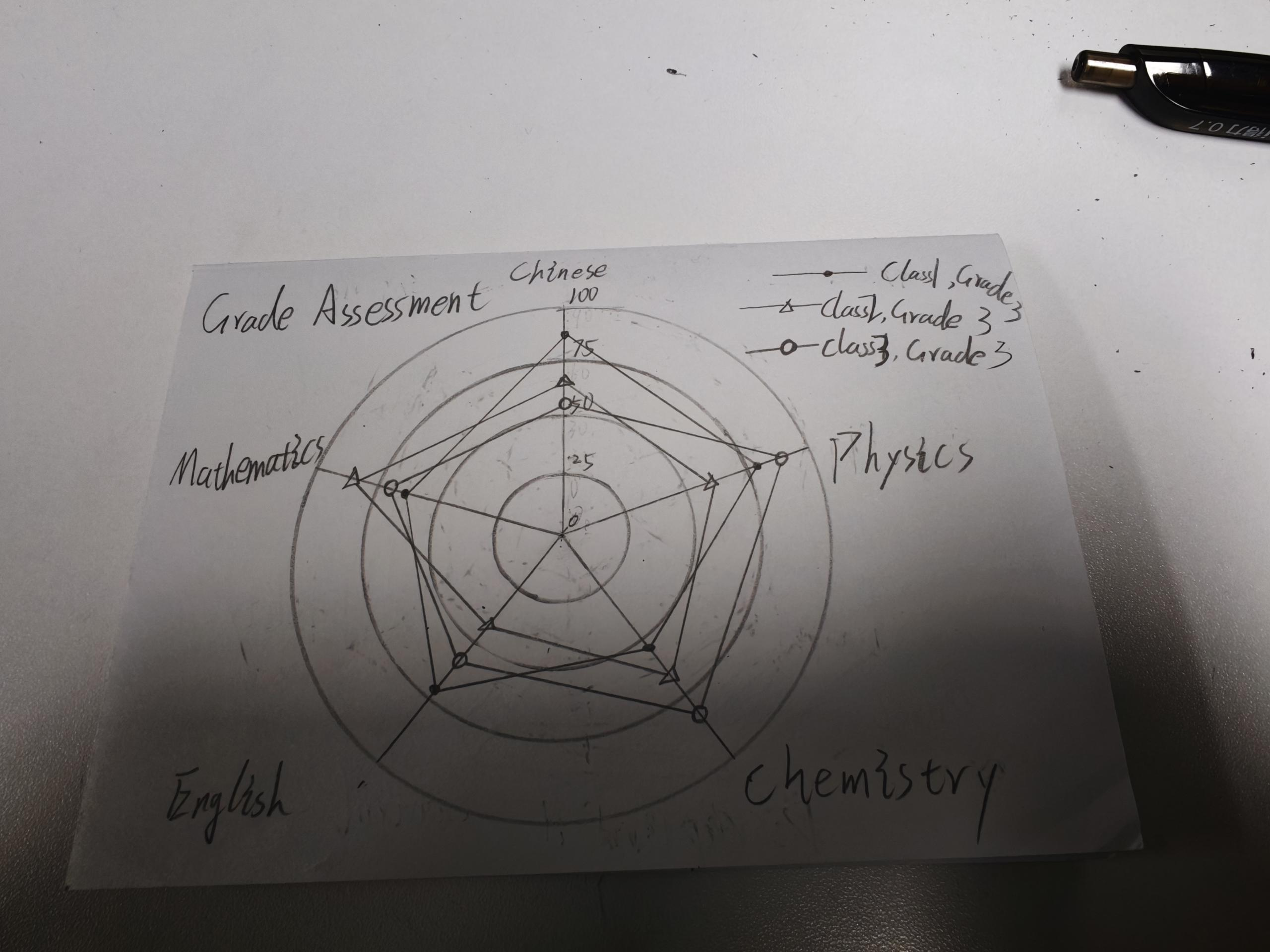}
        \par\vspace{0.3em}{\small (d) \textbf{Radar chart} -- hand-drawn, 3 series.}
    \end{minipage}\hfill
    \begin{minipage}[b]{0.60\textwidth}
        \vspace{0pt}%
        \begin{sample_anno_box}{Markdown table}
            \#\#\# Grade Assessment\\
            |             | Class1, Grade3 | Class2, Grade3 | Class3, Grade3 |\\
            | ---         | --- | --- | --- |\\
            | Chinese     | 87 | 65 | 55 |\\
            | Physics     | 79 | 60 | 90 |\\
            | Chemistry   | 54 | 67 | 83 |\\
            | English     | 72 | 42 | 58 |\\
            | Mathematics | 64 | 85 | 70 |
        \end{sample_anno_box}
    \end{minipage}
    \caption{Representative sample: hand-drawn radar chart with overlapping rings.}
    \label{fig:sample_radar}
\end{figure}

\begin{figure}[t]
    \centering
    \begin{minipage}[b]{0.38\textwidth}
        \vspace{0pt}%
        \sampleimg[3.8cm]{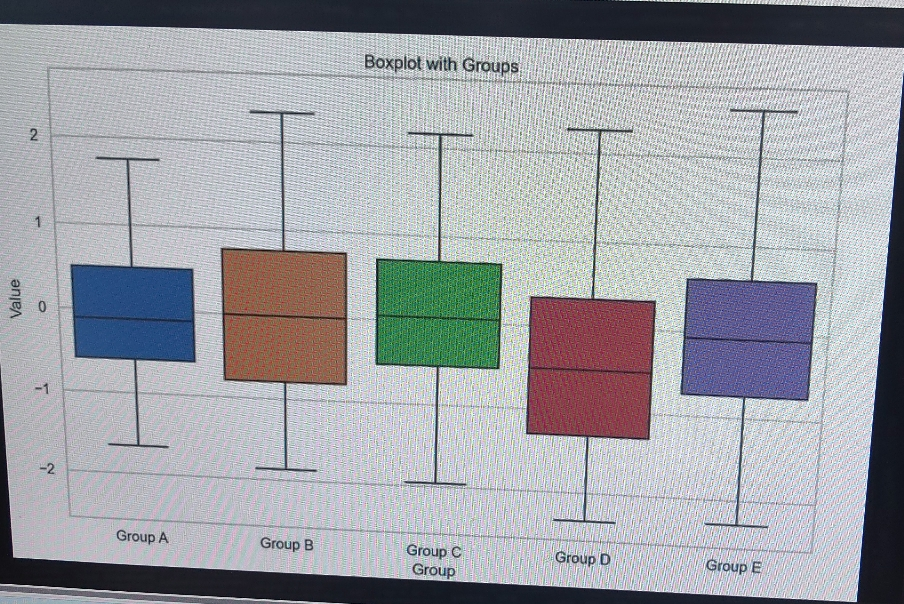}
        \par\vspace{0.3em}{\small (e) \textbf{Box plot} -- photo (printed).}
    \end{minipage}\hfill
    \begin{minipage}[b]{0.60\textwidth}
        \vspace{0pt}%
        \begin{sample_anno_box}{Markdown table}
            \#\#\# Boxplot with Groups\\
            | Group   | Value-min | Value-Q1 | Value-median | Value-Q3 | Value-max |\\
            | ---     | --- | --- | --- | --- | --- |\\
            | Group A | -1.6  | -0.6  | -0.1  | 0.5  | 1.8 |\\
            | Group B | -1.9  | -0.8  | 0     | 0.8  | 2.3 |\\
            | Group C | -1.96 | -0.5  | 0.05  | 0.7  | 2.2 |\\
            | Group D | -2.4  | -1.5  | -0.7  | 0.1  | 2.1 |\\
            | Group E | -2.3  | -1.0  | -0.4  | 0.4  | 2.2 |
        \end{sample_anno_box}
    \end{minipage}
    \caption{Representative sample: box plot.}
    \label{fig:sample_box}
\end{figure}

\clearpage
\section{Training Details}
\label{supp:sec:training_details}
\vspace{-0.5em}

In this section, we provide additional training details, aiming to present the experimental setup more comprehensively and facilitate reproducibility for the community.

\subsection{Training Setup}
\label{supp:subsec:training_setup}
\vspace{-0.5em}

We summarize the detailed training configurations for the pretraining and supervised fine-tuning (SFT) stages in \cref{tab:training_setup}. Full-parameter fine-tuning is applied in all stages with a maximum context length of 8192 tokens and a maximum image resolution of 2K.

\begin{table*}[h!]
    \centering
    \caption{
        \textbf{Training setup and hyperparameters for pretraining and SFT.}
    }
    \label{tab:training_setup}
    \vspace{-0.6em}
    \small
    \setlength{\tabcolsep}{8pt}
    \renewcommand{\arraystretch}{1.05}
    \begin{tabular}{c l >{\centering\arraybackslash}p{0.18\textwidth} >{\centering\arraybackslash}p{0.18\textwidth}}
        \toprule
         & \textbf{Setting}                     & \textbf{Pretraining}                            & \textbf{SFT}      \\
        \midrule

        \multirow{2}{*}{\rotatebox[origin=c]{90}{\small\textit{ViT}}}
         & {Max Resolution}                     & 2K                                              & 2K                \\
         & {Tokens per Image}                   & 256-1280                                        & 256-1280          \\
        \midrule

        \multirow{3}{*}{\rotatebox[origin=c]{90}{\small\textit{Data}}}
         & {Dataset}                            & Synth + Public                                  & Public + In house \\
         & {Raw Samples}                        & 4M                                              & 80K               \\
         & {Training Samples}                   & 4M                                              & 80K               \\
        \midrule

        \multirow{3}{*}{\rotatebox[origin=c]{90}{\small\textit{Model}}}
         & {Trainable Parameters}               & All                                             & All               \\
         & \multirow{1}{*}{{Training Paradigm}} & Direct Parsing                                  & CAP Reasoning     \\
         & {Sequence Length}                    & 8192                                            & 8192              \\
        \midrule

        \multirow{7}{*}{\rotatebox[origin=c]{90}{\small\textit{Training}}}
         & {Batch Size}                         & 128                                             & 128               \\
         & {Epoch}                              & 1                                               & 1                 \\
         & {Learning Rate}                      & $4\times10^{-5}$                                & $2\times10^{-5}$  \\
         & {min-LR}                             & $2\times10^{-6}$                                & $1\times10^{-6}$  \\
         & {LR scheduler}                       & \multicolumn{2}{c}{Cosine}                                          \\
         & {Optimizer}                          & \multicolumn{2}{c}{Adam}                                            \\
         & {Hardware}                           & \multicolumn{2}{c}{16 $\times$ NVIDIA H20 GPUs}                     \\
        \bottomrule
    \end{tabular}
\end{table*}

\subsection{Training Data}
\label{supp:subsec:training_data}
\vspace{-0.5em}

\mypara{Pretraining data.}
Our pretraining corpus combines two sources.
First, we select $0.8$M samples from ChartGalaxy~\cite{ChartGalaxy_2025} and $1.2$M samples from ChartNet~\cite{ChartNet_2026}, two high-quality synthetic chart corpora, for a total of $2$M filtered image--annotation pairs.
Since both sources cover only numeric chart families and do not include topologically richer diagrammatic charts such as flowcharts and mind maps, we further augment the corpus with another $2$M image--text pairs synthesized by the data pipeline described in~\cref{sec:data_pipeline}, where we not only fill the missing diagrammatic slice but also substantially expand the diversity of numeric chart coverage.

\mypara{SFT data.}
Our SFT dataset has two parts.
The first is $60$K high-quality human-annotated samples released with ChartNet~\cite{ChartNet_2026}, which remain synthetic in their underlying image source.
To complement this with real-world visual variation, we contribute an additional $20$K in-house human-annotated samples spanning three visual styles: digital renderings, printed scans, and hand-drawn scans.
The rendering perturbations of~\cref{subsec:sft_diversity}, namely font variation and geometric warping, are applied only to the ChartNet images and transform them in place, so the number of training samples is unchanged; the in-house images are used as captured. The contribution of each data source is analyzed in~\cref{supp:subsec:data_attribution}.

\mypara{CAP-based reasoning construction.}
Following~\cref{subsec:cap_reasoning}, we first extract atomic primitives from each $(I, Y)$ pair by prompting an MLLM; the full extraction prompt is shown in~\cref{prompt:chart_2D_extract}.
Notably, the prompt is slightly adapted per chart family to reflect the chart-type-specific schema of~\cref{subsec:cap} (e.g., pie charts replace the $x$/$y$-axis slots with ``names of slices''), but in all cases the extraction is restricted to chart-level semantics while discarding low-level visual details.
Given the extracted atomic primitives $A$, we then assemble a reasoning trace following~\cref{eq: reasoning_construction}, whose templating logic is summarized in~\cref{alg:chart_cot_pipeline}: the atomic fields are serialized into an ordered sequence of semantic analysis steps (chart identification, data subject, structural analysis) and wrapped into the reasoning trace $R$.
Finally, $R$ is combined with the parsing answer $Y$ via~\cref{eq: unified_supervision} to form the unified supervision target $T$ used throughout SFT stage.

\begin{algorithm}[th!]
    \caption{CAP-based Reasoning Trace Construction ($\mathcal{G}$)}
    \begin{algorithmic}[1]
        \Require Dataset $\mathcal{D}_{\text{raw}} = \{(I_i, Y_i, \tau_i, A_i)\}_{i=1}^N$, where $I_i$ denotes the chart image, $Y_i$ the target annotation, $\tau_i$ the chart type, and $A_i$ the extracted CAP atomic semantics (\myeg title, axes)        
        \Ensure Supervised dataset $\mathcal{D}_{\text{sft}} = \{(I_i, T_i)\}$, where $T_i$ is the structured target that encodes both CAP-based reasoning and the final output (see \cref{eq: unified_supervision})
        \Statex

        \Function{ConstructSupervisionTargets}{$\mathcal{D}_{\text{raw}}$}
        \State $\mathcal{D}_{\text{sft}} \gets \emptyset$ \Comment{Output buffer for unified supervision samples}

        \For{\textbf{each} $(I, Y, \tau, A) \in \mathcal{D}_{\text{raw}}$}

        \State \textbf{\# [Shortcut Branch] Types whose structure is already sequential}
        \State \textbf{\# (e.g.\ flowcharts in Mermaid) need no extra reasoning scaffold}
        \If{$\tau \in \mathcal{T}_{\text{short}}$}
            \State $T \gets \text{``\textless answer\textgreater''} \oplus Y \oplus \text{``\textless /answer\textgreater''}$ \Comment{Directly wrap the annotation}
            \State $\mathcal{D}_{\text{sft}} \gets \mathcal{D}_{\text{sft}} \cup \{(I, T)\}$
            \State \textbf{continue} \Comment{Skip reasoning construction for this sample}
        \EndIf

        \State
        \State \textbf{\# [Step 1] Chart Identification --- high-level "what is this chart"}
        \State \textbf{\# Extract chart type, title and caption as the opening statement}
        \State $S_{\text{id}} \gets \textsc{FormatIdentity}(\tau, A.\text{title}, A.\text{caption})$

        \State
        \State \textbf{\# [Step 2] Data Subject --- "what entities are being compared"}
        \State \textbf{\# Different chart types expose different subject fields:}
        \State \textbf{\# \quad 2D chart $\to$ comparison\_items / statistics\_fields}
        \State \textbf{\# \quad radar    $\to$ comparison\_items / dimensions}
        \State \textbf{\# \quad mindmap  $\to$ main\_branches / semantic\_domain}
        \State \textbf{\# \quad pie      $\to$ categories / value\_semantic}
        \State $\mathcal{F}_{\text{subj}} \gets \textsc{SelectSubjectFields}(\tau)$ \Comment{Type-specific field selector}
        \State $S_{\text{subj}} \gets \textsc{FormatFields}(A, \mathcal{F}_{\text{subj}})$ \Comment{Serialise selected fields to text}

        \State
        \State \textbf{\# [Step 3] Structural Analysis --- "how is the chart organized"}
        \State \textbf{\# Captures axes semantics, tick ranges, tree depth, \# sectors, etc.}
        \State $\mathcal{F}_{\text{struct}} \gets \textsc{SelectStructFields}(\tau)$ \Comment{Type-specific structural fields}
        \State $S_{\text{struct}} \gets \textsc{FormatFields}(A, \mathcal{F}_{\text{struct}})$

        \State
        \State \textbf{\# [Step 4] Assemble the interleaved reasoning trace $R$}
        \State \textbf{\# Format: \textless think\textgreater\ [3 sections] \textless /think\textgreater\ \textless answer\textgreater\ $Y$ \textless /answer\textgreater}
        \State $R \gets \text{``[Chart Identification]\textbackslash n''} \oplus S_{\text{id}}   \oplus \text{``\textbackslash n\textbackslash n''}$ \Comment{Section 1}
        \State $R \gets R \oplus \text{``[Data Subject]\textbackslash n''}        \oplus S_{\text{subj}} \oplus \text{``\textbackslash n\textbackslash n''}$ \Comment{Section 2}
        \State $R \gets R \oplus \text{``[Structural Analysis]\textbackslash n''} \oplus S_{\text{struct}}$ \Comment{Section 3}

        \State $T \gets \text{``\textless think\textgreater''}  \oplus R \oplus \text{``\textless /think\textgreater''}$ \Comment{Wrap reasoning trace}
        \State $T \gets T \oplus \text{``\textless answer\textgreater''} \oplus Y \oplus \text{``\textless /answer\textgreater''}$ \Comment{Append ground-truth parse}

        \State $\mathcal{D}_{\text{sft}} \gets \mathcal{D}_{\text{sft}} \cup \{(I, T)\}$
        \EndFor

        \State \Return $\mathcal{D}_{\text{sft}}$
        \EndFunction

        \Statex
        \Function{FormatFields}{$A, \mathcal{F}$}
        \State \textbf{\# Generic serialiser: iterate over the selected fields of $A$ and}
        \State \textbf{\# emit "field: stringified\_value" lines, skipping empty ones.}
        \State $Lines \gets []$
        \For{\textbf{each} field $f \in \mathcal{F}$}
            \State $v \gets A[f]$ \Comment{Fetch value from atomic semantics}
            \If{$v$ is non-empty}
                \State $Lines.\text{Append}(\text{``}f\text{: ''} \oplus \textsc{Stringify}(v))$ \Comment{e.g.\ list $\to$ comma-joined string}
            \EndIf
        \EndFor
        \State \Return $\text{Join}(Lines, \text{``\textbackslash n''})$ \Comment{One field per line}
        \EndFunction

    \end{algorithmic}
    \label{alg:chart_cot_pipeline}
\end{algorithm}

\section{Further Analysis of CAP}
\label{supp:sec:cap_analysis}
\vspace{-0.5em}

This section complements the ablation study in~\cref{tab:ablation_training_stage} with further analyses. We verify the quality of the generated CAP targets in~\cref{supp:subsec:cap_verification}, separate the contribution of CAP from target length and canonicalization in~\cref{supp:subsec:cap_isolation}, attribute the gains on printed and hand-drawn charts to the training data sources in~\cref{supp:subsec:data_attribution}, and examine whether CAP carries over to a larger backbone in~\cref{supp:subsec:cap_scaling}.

\subsection{Human Verification of CAP Targets}
\label{supp:subsec:cap_verification}
\vspace{-0.5em}

As in~\cref{subsec:cap_reasoning}, the CAP reasoning targets are generated by an MLLM, so errors in axes, entity relations, or hierarchies could in principle enter the supervision. A full manual review of all samples is impractical. We therefore randomly sample $400$ CAP targets and have four independent annotators verify them under the same quality-review protocol of the benchmark (\cref{supp:subsec:annotation_protocol}). The verified targets reach an accuracy of about $97\%$. This high reliability is due to two factors: primitives are extracted from human-verified ground-truth pairs, and the fixed CAP schema restricts extraction to a few chart-level fields, leaving little room for inconsistency.

\subsection{Isolating the Contribution of CAP}
\label{supp:subsec:cap_isolation}
\vspace{-0.5em}

The gain of CAP in~\cref{tab:ablation_training_stage} could in principle come from longer targets, a fixed template, or a canonical ordering rather than from the structural primitives themselves. Because CAP is carried by an explicit reasoning trace for numeric charts and by the target structure for flowcharts (\cref{subsec:cap_reasoning}), we design the controls separately for the two families. All control runs start from the same pretrained checkpoint and share the training data, token budget, and hyperparameters of row (e) of~\cref{tab:ablation_training_stage}.

\mypara{Numeric charts.}
We compare the direct SFT target of row (b) and the CAP target of row (e) with three variants of the numeric-chart target. (i) The length-matched rewrite keeps the content and length of the CAP trace but writes it as free-form natural language without the structured layout. (ii) The random-order variant shuffles the three CAP steps (chart identification, data subject, and structural analysis) independently for each sample. (iii) The swap variant keeps chart identification first and only exchanges the data-subject and structural-analysis steps. Results are reported in~\cref{tab:cap_numeric_controls}.

\begin{table}[t!]
    \centering
    \captionsetup{font={small}}
    \caption{
        \textbf{Controls for the numeric-chart CAP target.}
        Numeric-chart mAP$_{\text{high}}$ under three visual scenarios. The first two rows are rows (b) and (e) of~\cref{tab:ablation_training_stage}. The  three controls keep the training data, token budget, and hyperparameters of row (e) and change only the numeric-chart target.
    }
    \label{tab:cap_numeric_controls}
    {
        \renewcommand{\arraystretch}{1.0}
        \setlength{\tabcolsep}{10pt}
        \resizebox{0.8\columnwidth}{!}{
            \begin{tabular}{l ccc}
                \toprule
                \textbf{Numeric-chart target}                                   & \textbf{Digital} & \textbf{Printed} & \textbf{Drawn} \\
                \midrule
                Direct (w/o CAP), row (b) of~\cref{tab:ablation_training_stage} & 63.1             & 57.8             & 44.9           \\
                \rowcolor{table_ours}
                CAP, row (e) of~\cref{tab:ablation_training_stage}              & \textbf{64.1}    & \textbf{60.4}    & \textbf{48.6}  \\
                \midrule
                (i) Length-matched natural-language rewrite                     & 60.1             & 52.2             & 41.8           \\
                (ii) CAP steps in random order                                  & 61.3             & 54.3             & 42.1           \\
                (iii) CAP with steps 2 and 3 swapped                            & 63.5             & 58.2             & 46.9           \\
                \bottomrule
            \end{tabular}
        }
    }
    \vspace{-0.6em}
\end{table}

At identical length and coverage, the natural-language rewrite falls $4.0$, $8.2$, and $6.8$ points below CAP and even below the direct baseline. Longer targets are therefore not what drives the gain. A free-form reasoning path of the same length is hard for a 1B model to learn, whereas the structured layout of CAP is a much easier target to internalize. Random ordering loses $2.8$, $6.1$, and $6.5$ points and also falls below the direct baseline, so a fixed template with an arbitrary ordering is not sufficient either. Swapping only the last two steps costs much less ($0.6$, $2.2$, and $1.7$ points) and stays above the direct baseline, which identifies chart identification as the critical anchor of the trace. The benefit of CAP thus comes from its structural primitives and their human-aligned ordering.

\mypara{Flowcharts.}
Flowchart evaluation is inherently invariant to node and edge ordering because predictions are parsed into node/edge sets and scored via Hungarian matching (detailed in~\cref{supp:subsec:graph_scoring}). Any impact of target formatting thus comes from training rather than the metric. To separate canonicalization from the ordering direction, we train three models that differ only in how the flowchart target is arranged, while the numeric side is fixed to CAP. The three targets encode the same graph. (i) The interleaved target is the raw Mermaid style, in which a node is declared where an edge first refers to it; this is the direct setting of rows (b) and (c). (ii) The edge-first target is fully canonicalized but lists every edge before appending all node declarations. (iii) The node-first target applies the same canonicalization in the opposite direction and is the CAP setting used in the paper.

\begin{table}[h!]
    \centering
    \captionsetup{font={small}}
    \caption{
        \textbf{Ordering ablation for the flowchart target.}
        mAP$_{\text{high}}$ under three visual scenarios. The three SFT targets encode the same graph and differ only in where node declarations are placed; the numeric side is fixed to CAP. The interleaved and node-first rows correspond to the flowchart slice of rows (c) and (e) of~\cref{tab:ablation_training_stage}.
    }
    \label{tab:cap_flowchart_order}
    {
        \renewcommand{\arraystretch}{1.0}
        \setlength{\tabcolsep}{14pt}
        \resizebox{0.9\columnwidth}{!}{
            \begin{tabular}{l cccc}
                \toprule
                \textbf{Flowchart target}                  & \textbf{Digital} & \textbf{Printed} & \textbf{Drawn} & \textbf{Mean} \\
                \midrule
                Pretrain only                              & 41.9             & 40.8             & 30.7           & 37.8          \\
                \midrule
                (i) SFT, interleaved (not canonicalized)   & 56.0             & 43.6             & 44.4           & 48.0          \\
                (ii) SFT, canonicalized, edge-first        & 55.5             & 42.1             & 42.4           & 46.7          \\
                \rowcolor{table_ours}
                (iii) SFT, canonicalized, node-first (CAP) & \textbf{56.8}    & \textbf{48.3}    & \textbf{47.8}  & \textbf{51.0} \\
                \bottomrule
            \end{tabular}
        }
    }
    \vspace{-0.6em}
\end{table}

As shown in~\cref{tab:cap_flowchart_order}, relative to the interleaved target, the edge-first layout loses $0.5$, $1.5$, and $2.0$ points across scenarios. The node-first layout instead gains $3.0$ points on average, yielding a $4.3$-point gap attributable to ordering direction. Canonicalization alone is thus mildly harmful, with all benefits stemming from committing to the node set first. This aligns with CAP principles: nodes are flowchart primitives while edges define their relations, so a node-first target lets the model fix primitives before choosing edges. An edge-first target inverts this dependency, forcing models to emit relations for undeclared nodes. This pattern mirrors the numeric-chart controls where shuffled CAP steps underperformed. The gain is concentrated on photographed charts and minor on digital renderings (consistent with~\cref{subsec:ablation}), matching the pooled diagrammatic columns in~\cref{tab:ablation_training_stage}.

\subsection{Attribution of Real-World Robustness to Training Data}
\label{supp:subsec:data_attribution}
\vspace{-0.5em}

The training data combines synthetic pretraining pairs, the ChartNet SFT set (with synthetic image sources), in-house annotations of real printed and hand-drawn charts, and rendering perturbations (\cref{supp:subsec:training_data}). We examine which component drives gains on printed and hand-drawn charts.

\mypara{Constraints from the existing setup.}
Three facts constrain this attribution. First, row (a) of~\cref{tab:ablation_training_stage} is a synthetic-only control at scale: its entirely synthetic pretraining corpus lifts the four printed and hand-drawn cells by just $4.3$ points (concentrating $70\%$ of its gain on digital renderings), whereas the SFT stage adds $28.6$ points—about $6.7$ times as much. Second, ChartNet SFT images are synthetic, and similar samples were seen in much greater volume during pretraining. Third, rendering perturbations apply exclusively to ChartNet images, yielding only rendered charts. Real printed and hand-drawn photos exist solely in our in-house annotations. Moreover, because ChartNet only covers numeric charts, neither the synthetic set nor its perturbations touch diagrammatic charts.

\mypara{Controlled ablations.}
We decompose the SFT stage into synthetic-only, in-house-only, augmentation-only, and mixed-data controls. All runs start from the pretrained checkpoint of row (a), keep CAP disabled so that data is the only variable, and are trained under the same token budget with the SFT configuration of the main paper. Results are reported in~\cref{tab:data_attribution}, where rows (b$_2$), (b$_3$), (b) versus (b$_1$), and (b) instantiate the four controls, respectively.

\begin{table}[t!]
    \centering
    \captionsetup{font={small}}
    \caption{
        \textbf{Attribution of the SFT-stage gains to data sources.}
        mAP$_{\text{high}}$ under three visual scenarios. Rows (a) and (b) are those of~\cref{tab:ablation_training_stage}; rows (b$_1$)--(b$_3$) are variants of row (b) that change only the SFT data. All SFT runs start from the checkpoint of row (a), disable CAP, and are trained under the same token budget.
    }
    \label{tab:data_attribution}
    {
        \renewcommand{\arraystretch}{1.0}
        \setlength{\tabcolsep}{10pt}
        \resizebox{\columnwidth}{!}{
            \begin{tabular}{l ccc ccc}
                \toprule
                \multirow{2}{*}[-0.8mm]{\textbf{Setting}}  & \multicolumn{3}{c}{\textbf{Numeric Charts}} & \multicolumn{3}{c}{\textbf{Diagrammatic Charts}}                                                                  \\
                \cmidrule(lr){2-4} \cmidrule(lr){5-7}
                                                            & Digital                                     & Printed                                          & Drawn         & Digital       & Printed       & Drawn         \\
                \midrule
                MLLM Backbone                               & 56.3                                        & 49.7                                             & 37.4          & 55.0          & 50.4          & 26.0          \\
                (a) Pretrain (synthetic only)               & 62.5                                        & 50.2                                             & 38.1          & 58.9          & 51.8          & 27.7          \\
                (b) + SFT, ChartNet + in-house (mixed)      & 63.1                                        & \textbf{57.8}                                    & \textbf{44.9} & 59.6          & \textbf{55.8} & \textbf{37.9} \\
                (b$_1$) + SFT, mixed without augmentation   & 62.2                                        & 57.5                                             & 44.2          & 58.1          & 55.4          & 37.3          \\
                (b$_2$) + SFT, ChartNet only                & \textbf{63.9}                               & 51.2                                             & 39.9          & 58.3          & 52.1          & 28.1          \\
                (b$_3$) + SFT, in-house only                & 62.8                                        & 56.3                                             & 44.7          & \textbf{60.2} & 54.8          & 36.4          \\
                \bottomrule
            \end{tabular}
        }
    }
    \vspace{-0.6em}
\end{table}

Summing the four printed and hand-drawn cells, the gain over row (a) is only $3.5$ points for (b$_2$) but $24.4$ points for (b$_3$), seven times as much from the smaller in-house set and a $21\times$ difference once normalized per sample. The in-house real images alone thus account for about $85\%$ of the total gain of $28.6$ points. The perturbations are measured on a different contrast: removing them from the full mixture, (b) versus (b$_1$), costs $2.0$ points on the same cells, a real but secondary effect. The contrast is clearest on flowcharts and mind maps, where (b$_2$) changes the three cells by only $-0.6$, $+0.3$, and $+0.4$ points while (b$_3$) changes them by $+1.3$, $+3.0$, and $+8.7$ points, as expected since ChartNet contains no diagrammatic charts.

The two sources are complementary rather than redundant, which is why the full mixture is used in the main experiments. Row (b$_2$) reaches the best numeric digital score of the table ($63.9$, $+1.4$ over row (a)), whereas (b$_3$) stays nearly flat there ($+0.3$), so the synthetic set contributes rendering-level precision and the in-house set contributes real-world robustness. Relative to (b$_2$), adding the in-house set gives up $0.8$ points on numeric digital renderings but gains $6.6$ and $5.0$ points on printed and hand-drawn numeric charts.

\subsection{Scaling CAP to a Larger Backbone}
\label{supp:subsec:cap_scaling}
\vspace{-0.5em}

To test whether CAP carries over beyond a 1B parser, we adopt Qwen3.5-9B~\cite{Qwen3_5_2026}, a newer and considerably larger general-purpose MLLM, as the backbone. Because it is already strong, we skip the pretraining stage and apply SFT only, so that the contrast between two SFT settings attributes it to CAP. All other settings, including the SFT data, the evaluation protocol, and the metric, follow the main experiments. We compare the backbone with SFT without and with CAP, whose configurations correspond to rows (b) and (e) of~\cref{tab:ablation_training_stage}. Results are reported in~\cref{tab:cap_scaling}.

\begin{table}[t!]
    \centering
    \captionsetup{font={small}}
    \caption{
        \textbf{CAP on larger backbone Qwen3.5-9B.}
        \emph{Top}: average mAP$_{\text{high}}$ over all chart families, reported separately for English and Chinese. \emph{Bottom}: mAP$_{\text{high}}$ for the two chart categories under three visual scenarios.
    }
    \label{tab:cap_scaling}
    {
        \renewcommand{\arraystretch}{1.0}
        \setlength{\tabcolsep}{18pt}
        \resizebox{\columnwidth}{!}{
            \begin{tabular}{l ccc ccc}
                \toprule
                \textbf{Setting}               & \multicolumn{3}{c}{\textbf{Average (EN)}}   & \multicolumn{3}{c}{\textbf{Average (ZH)}}                                                                        \\
                \midrule
                Qwen3.5-9B~\cite{Qwen3_5_2026} & \multicolumn{3}{c}{40.6}                    & \multicolumn{3}{c}{47.6}                                                                                         \\
                + SFT (w/o CAP)                & \multicolumn{3}{c}{55.9}                    & \multicolumn{3}{c}{67.9}                                                                                         \\
                \rowcolor{table_ours}
                + SFT (w/ CAP)                 & \multicolumn{3}{c}{\textbf{58.8}}           & \multicolumn{3}{c}{\textbf{71.0}}                                                                                \\
                \midrule
                \multirow{2}{*}[-0.8mm]
                {\textbf{Setting}}             & \multicolumn{3}{c}{\textbf{Numeric Charts}} & \multicolumn{3}{c}{\textbf{Diagrammatic Charts}}                                                                 \\
                \cmidrule(lr){2-4} \cmidrule(lr){5-7}
                                               & Digital                                     & Printed                                          & Drawn         & Digital       & Printed       & Drawn         \\
                \midrule
                Qwen3.5-9B                     & 48.4                                        & 43.6                                             & 33.9          & 59.7          & 54.3          & 37.9          \\
                + SFT (w/o CAP)                & 70.6                                        & 65.5                                             & 55.2          & 66.2          & 58.6          & 43.9          \\
                \rowcolor{table_ours}
                + SFT (w/ CAP)                 & \textbf{72.1}                               & \textbf{67.7}                                    & \textbf{60.8} & \textbf{67.4} & \textbf{61.0} & \textbf{48.2} \\
                \bottomrule
            \end{tabular}
        }
    }
\end{table}

\mypara{A stronger backbone learns chart parsing more effectively.}
The untrained Qwen3.5-9B scores below the HunyuanOCR backbone used in the paper ($40.6$/$47.6$ versus $41.0$/$51.8$ on English/Chinese), which is expected since HunyuanOCR is specialized for OCR. With the same SFT data, however, Qwen3.5-9B with CAP reaches $58.8$/$71.0$, well above the $49.5$/$61.3$ of our 1B model in~\cref{tab:ChartArena_main_results}. We attribute this to the richer world knowledge and stronger general understanding of the larger backbone.

\mypara{CAP remains beneficial at the 9B scale.}
Adding CAP on top of the same SFT recipe lifts the average by $2.9$ points in English and $3.1$ points in Chinese. The benefit of CAP does not vanish once the backbone becomes larger and more capable, indicating CAP as a scale-independent mechanism.

\mypara{The gain concentrates on visually harder scenarios.}
The improvement grows as visual conditions degrade. On numeric charts the CAP gain goes from $1.5$ points on digital renderings to $2.2$ on printed photos and $5.6$ on hand-drawn photos, and on diagrammatic charts it goes from $1.2$ to $2.4$ and then $4.3$ points. This trend is consistent with~\cref{subsec:ablation} and confirms that making chart structure explicit mainly improves robustness under visually uncertain real-world inputs.

\clearpage
\section{Evaluation Details}
\label{supp:sec:evaluation_details}
\vspace{-0.5em}

This section provides a detailed description of the unified evaluation protocol in~\cref{subsec:unified_eval_protocol}, and explains how ChartArena converts heterogeneous model outputs into a single, comparable score. We first describe how predictions are routed to one of two semantic views (\cref{supp:subsec:eval_routing}). We then formalize the two scoring backends on these views: a triple-based IoU for numeric charts (\cref{supp:subsec:triple_scoring}) and a graph-matching score or tree-based score for diagrammatic charts (\cref{supp:subsec:graph_scoring}). Finally, we describe how per-sample similarities are aggregated into the reported \emph{EM} and \emph{mAP} (\cref{supp:subsec:aggregation}).

\subsection{Format-Agnostic Routing and Normalization}
\label{supp:subsec:eval_routing}
\vspace{-0.5em}

For each evaluation sample, we start from the raw model output $\hat{Y}$ and the reference annotation $Y$. The evaluator first routes $\hat{Y}$ into a canonical representation based on two pieces of information:
(i) the declared surface format of $\hat{Y}$, such as Markdown, CSV, JSON, code, or SVG; and
(ii) the structural type of the reference $Y$, namely \emph{numeric triples}, \emph{hierarchical trees}, or \emph{directed graphs}.
The high-level routing procedure is shown in~\cref{alg:eval_routing}.

\begin{algorithm}[t]
    \small
    \caption{Routing a prediction to its canonical evaluation view.}
    \label{alg:eval_routing}
    \begin{algorithmic}[1]
        \Require raw prediction $\hat{Y}$, reference $Y$, declared task $\tau$
        \Ensure canonical representation $X_{\hat{Y}}$ and view $v \in \{\text{triple}, \text{tree}, \text{graph}\}$

        \Statex
        \Function{RouteToCanonicalView}{$\hat{Y}, Y, \tau$}

        \State \textbf{\# [Step 1] Graph-based formats (primary path for diagrammatic charts)}
        \State \textbf{\# Includes Mermaid, DOT, PlantUML, D2, Diagrams, Cytoscape, etc.}
        \If{$\tau \in \mathcal{T}_{\text{graph}}$}
            \State $X \gets \textsc{ParseGraph}(\hat{Y}, \tau)$ \Comment{Dedicated DSL parser; failure $\Rightarrow$ empty graph}
            \State \Return $(X, \text{graph})$
        \EndIf

        \State
        \State \textbf{\# [Step 2] Tree-based formats (mind maps)}
        \State \textbf{\# Identified by reference structure rather than surface format}
        \If{$Y$ is a Markdown bullet list}
            \State $X \gets \textsc{ToMarkdownTree}(\hat{Y}, \tau)$
            \State \Return $(X, \text{tree})$
        \EndIf

        \State
        \State \textbf{\# [Step 3] Graph fallback (flowcharts expressed in Mermaid)}
        \If{$Y$ is Mermaid code}
            \State $X \gets \textsc{ToMermaid}(\hat{Y}, \tau)$
            \State \Return $(X, \text{graph})$
        \EndIf

        \State
        \State \textbf{\# [Step 4] Numeric charts (default branch)}
        \State \textbf{\# Covers bar, line, pie, radar, box plot, and combination charts}
        \State $X \gets \textsc{ToInternalCSV}(\hat{Y}, \tau)$
        \State \Return $(X, \text{triple})$

        \EndFunction
    \end{algorithmic}
\end{algorithm}

\mypara{Per-format adapters.}
Each surface format has a deterministic adapter that extracts the semantic content from the prediction and discards presentation details.
For numeric charts, \textsc{ToInternalCSV} converts the prediction into a tabular view.
For mind maps, \textsc{ToMarkdownTree} converts the prediction into a tree-structured Markdown view.
For diagrammatic charts, the corresponding parser converts the prediction into a graph representation.
These adapters do not attempt to correct semantic mistakes in $\hat{Y}$.
Their only goal is to map different output languages into a shared canonical form.

\mypara{Light-touch normalization on the canonical view.}
After routing, we apply only semantics-preserving normalization.
This includes simple punctuation mapping, such as converting full-width symbols to half-width ones, removing symbols such as \verb|$| and \verb|%| from numeric cells, and rewriting equivalent box-plot headers into a unified suffix, such as mapping \verb|lower quartile| and \verb|Q1| to \verb|-Q1|.
We do not apply any task-specific heuristic correction.
If a prediction cannot be normalized by these rules, we keep it unchanged for scoring.

\mypara{Parse failure.}
If an adapter cannot produce a non-empty canonical representation, the sample is marked as \texttt{parse\_failed}.
Examples include syntactically broken Mermaid code or JSON outputs that contain no usable data payload.
Such samples are assigned zero similarity under all thresholds.

\subsection{Triple-Based Scoring for Numeric Charts}
\label{supp:subsec:triple_scoring}
\vspace{-0.5em}

Our numeric-chart scoring follows the Structuring Chart-oriented Representation Metric (SCRM) proposed in StructChart~\cite{StructChart_2026}. We inherit its structured triples, tolerance-aware IoU matching, separate treatment of text through edit distance and numbers through relative error, three tolerance levels, and threshold-averaged mAP. Our adaptation routes heterogeneous surface formats into a common tabular view before matching. Graph and tree scoring extend the protocol to diagrammatic charts in~\cref{supp:subsec:graph_scoring}.

For numeric charts, both the prediction and the reference are first converted into a tabular CSV form, denoted as $X_{\hat{Y}}$ and $X_Y$.
Here, $X_{\hat{Y}}$ is obtained from the model output $\hat{Y}$ through the routing step, and $X_Y$ is the ground-truth table.

\mypara{Triple construction.}
We convert each table into a set of unordered triples of the form $(e, h, v)$, where $e$ is the entity (e.g.\ row or column category), $h$ is the header (e.g.\ series name), and $v$ is the value.
Given a table with $R$ rows and $C$ columns, this produces up to $R \times (C - 1)$ triples.

To make the representation invariant to table orientation, the pair $(e, h)$ is treated as an order-free tuple.
As a result, transposing the table yields the same set of triples.

\mypara{Pre-processing.}
Before matching, we apply several deterministic normalization steps:
(i) all text fields $e$ and $h$ are lower-cased;
(ii) values $v$ are cast to floating-point numbers when possible;
(iii) common header affixes in the reference (e.g.\ prefixes or suffixes in box plots) are aligned with the prediction;
(iv) triples corresponding to outlier or scatter markers are removed if they are not present in the reference.
These steps standardize representation but do not modify semantic content.

\mypara{Tolerance-aware matching.}
Let $T_{\hat{Y}}$ and $T_Y$ denote the triple sets from prediction and reference.
Following recent studies~\cite{ChartX_ChartVLM_2025, OneChart_2024, StructChart_2026}, we define a tolerance-aware matching rule between two triples $(e_1, h_1, v_1)$ and $(e_2, h_2, v_2)$.

First, their text keys are compared by concatenating the entity and the header, and measuring the Levenshtein distance~\cite{Levenshtein_1966}:

\vspace{-1.0em}
\begin{equation}
    \mathrm{Lev}(e_1 h_1, e_2 h_2) \le \epsilon_{\text{text}}.
\end{equation}
\par\nobreak\vspace{-0.5em}

Then, their values are compared as follows: if both $v_1$ and $v_2$ are non-numeric, we apply the same Levenshtein threshold; if both are numeric, we require the relative error to satisfy

\vspace{-1.0em}
\begin{equation}
    \frac{|v_1 - v_2|}{|v_2| + \delta} \le \epsilon_{\text{num}},
\end{equation}
\par\nobreak\vspace{-0.5em}

where $\delta$ is a small constant to avoid division by zero.

A pair of triples is considered a match only if both the text and value conditions are satisfied.

\mypara{Final score.}
Using the above matching rule, we compute a tolerance-aware intersection $T_{\cap}$ and union $T_{\cup}$ between the two triple sets.
The similarity score is then defined as

\vspace{-1.0em}
\begin{equation}
    s = \frac{|T_{\cap}|}{|T_{\cup}|},
\end{equation}
\par\nobreak\vspace{-0.5em}

which corresponds to an IoU-style metric.
If $T_{\cup} = \emptyset$, we define $s = 0$.

\mypara{Tolerance levels.}
The three evaluation levels (\emph{strict}, \emph{slight}, and \emph{high}) correspond to different settings of $(\epsilon_{\text{text}}, \epsilon_{\text{num}})$, as listed in~\cref{tab:tolerance}.

\begin{table}[t]
    \centering
    \caption{\textbf{Tolerance parameters across the three views.} $\epsilon_{\text{text}}$ is the maximum Levenshtein edit distance on the concatenated text keys; $\epsilon_{\text{num}}$ is the relative numeric tolerance; for the tree view, $\theta_{\text{path}}$ is the minimum Levenshtein ratio between two root-to-node paths; for the graph view, $\theta_{\text{item}}$ is the minimum per-node / per-edge similarity allowed to enter the Hungarian matching.}
    \label{tab:tolerance}
    \small
    \setlength{\tabcolsep}{8pt}
    \renewcommand{\arraystretch}{1.0}
    \begin{tabular}{l c c c c}
        \toprule
        \textbf{Tolerance}                       &
        \textbf{Triple $\epsilon_{\text{text}}$} &
        \textbf{Triple $\epsilon_{\text{num}}$}  &
        \textbf{Tree $\theta_{\text{path}}$}     &
        \textbf{Graph $\theta_{\text{item}}$}
        \\
        \midrule
        strict                                   & $0$ & $0$    & $1.00$ & $1.00$ \\
        slight                                   & $2$ & $0.05$ & $0.85$ & $0.85$ \\
        high                                     & $5$ & $0.10$ & $0.60$ & $0.60$ \\
        \bottomrule
    \end{tabular}
\end{table}

\subsection{Graph- and Tree-Based Scoring for Diagrammatic Charts}
\label{supp:subsec:graph_scoring}
\vspace{-0.5em}

For diagrammatic charts, both the prediction $\hat{Y}$ and the reference $Y$ are converted into structured representations.
Depending on the chart type, we use either a graph representation for flowcharts and similar diagrams or a tree representation for mind maps.

\mypara{Common graph representation.}
All supported diagram DSLs (\myeg Mermaid, Graphviz DOT, PlantUML, D2, Cytoscape JSON, and Python-based diagram libraries) are parsed into a unified graph intermediate representation

\vspace{-1.0em}
\begin{equation}
    \mathcal{G} = (V, \mathcal{L}_V, E),
\end{equation}
\par\nobreak\vspace{-0.5em}

where $V$ is the set of nodes, $\mathcal{L}_V: V \rightarrow \Sigma^\star$ maps each node to its text label, and
$E \subseteq V \times V \times \Sigma^\star$ is the set of directed edges with optional labels.

For Python-style outputs, parsing is performed via static AST analysis without executing the code.
This ensures safety and avoids non-determinism.

\mypara{Graph matching.}
Given a predicted graph $\mathcal{G}_{\hat{Y}}$ and a reference graph $\mathcal{G}_Y$, we compute similarity by matching nodes and edges separately.

Node similarity is defined using the Levenshtein ratio between node labels.
Edge similarity is defined as the average of three components: source node, target node, and edge label.
If both edge labels are empty, they are treated as a perfect match.

We construct two similarity matrices: one for nodes and one for edges.
Each matrix is then matched using the Hungarian algorithm, which finds the optimal one-to-one assignment.

After matching, only pairs with similarity above a threshold $\theta_{\text{item}}$ are kept.
The node and edge scores are computed as the average similarity over the matched pairs, normalized by the larger set size:

\vspace{-1.0em}
\begin{equation}
    \mathrm{Match}_V = \frac{1}{\max(|V_1|, |V_2|)} \sum S^V_{ij},
    \quad
    \mathrm{Match}_E = \frac{1}{\max(|E_1|, |E_2|)} \sum S^E_{ij}.
\end{equation}
\par\nobreak\vspace{-0.5em}

The final graph similarity is a weighted sum:

\vspace{-1.0em}
\begin{equation}
    s = w_E \cdot \mathrm{Match}_E + w_V \cdot \mathrm{Match}_V,
    \quad (w_E, w_V) = (0.6, 0.4),
\end{equation}
\par\nobreak\vspace{-0.5em}

which places more emphasis on edge correctness, as topology errors are more critical in practice.

\mypara{Tree scoring for mind maps.}
For mind maps, we use a simpler tree-based formulation.
Each Markdown bullet list is converted into a set of root-to-node paths.
Specifically, every prefix of every leaf path is treated as an individual path.

Two paths are compared by computing the Levenshtein ratio between their string representations (joined by \verb|" -> "|).
We then construct a similarity matrix between the predicted and reference path sets and apply Hungarian matching.

The final score is obtained by averaging the similarities of matched path pairs that exceed a threshold $\theta_{\text{path}}$.
This formulation rewards partial structural correctness, such as correctly identifying top-level branches even if some leaf nodes are incorrect.

\subsection{Aggregation into Exact Match and mAP}
\label{supp:subsec:aggregation}
\vspace{-0.5em}

For each evaluation sample $i$, the previous sections produce a similarity score
$s_i^{(t)} \in [0,1]$ under each tolerance level $t \in \{\text{strict}, \text{slight}, \text{high}\}$.
These similarities are the only inputs to the final metrics.

\mypara{Exact Match (EM).}
Exact Match measures the fraction of samples that are perfectly recovered under the strict setting:

\vspace{-1.0em}
\begin{equation}
    \mathrm{EM} = \frac{1}{N} \sum_{i=1}^N \mathbf{1}\!\left[s_i^{(\text{strict})} = 1\right].
\end{equation}
\par\nobreak\vspace{-0.5em}

A sample contributes to EM only if its prediction matches the reference exactly, without any tolerance.

\mypara{AP at a fixed threshold.}
To capture partial correctness, we evaluate whether each sample exceeds a similarity threshold $\theta$. For any threshold $\theta \in [0,1]$ and tolerance level $t$, we define
\vspace{-1.0em}
\begin{equation}
    \mathrm{AP}@\theta^{(t)} = \frac{1}{N} \sum_{i=1}^N \mathbf{1}\!\left[s_i^{(t)} \ge \theta\right],
\end{equation}
\par\nobreak\vspace{-0.5em}
which is the fraction of samples whose quality is above the bar $\theta$.

\mypara{Mean Average Precision (mAP).}
Instead of fixing a single $\theta$, our primary evaluator averages $\mathrm{AP}@\theta$ over a dense grid of thresholds to obtain a threshold-agnostic measure. For each tolerance level $t$, we compute
\vspace{-1.0em}
\begin{equation}
    \mathrm{mAP}^{(t)} = \frac{1}{10} \sum_{\theta \in \{0.50, 0.55, \dots, 0.95\}} \mathrm{AP}@\theta^{(t)}.
    \label{eq:map_definition}
\end{equation}
\par\nobreak\vspace{-0.5em}
This follows standard practice in detection benchmarks and reflects overall performance across quality levels; it is the metric backing all headline numbers in this paper (main results, ablations, and the aggregated tables in the appendix all report mAP$_{\text{high}}$, which is~\cref{eq:map_definition} with $t=\text{high}$).

\mypara{Reporting single-threshold slices for diagnostics.}
For finer-grained diagnostics we also report $\mathrm{AP}@\theta^{(t)}$ at three representative thresholds, $\theta \in \{0.5, 0.75, 0.90\}$. These are simply instantiations of the AP definition at three fixed cut-offs; they are not an alternative aggregation of mAP but three points sampled from the same $\mathrm{AP}@\theta$ curve. In particular, the ten thresholds used in~\cref{eq:map_definition} for aggregation and the three thresholds used here for point reporting serve different purposes and are consistent by construction; the paper does not aggregate mAP over $\{0.5,0.75,0.90\}$, nor does it report AP at all ten grid thresholds.

\mypara{Final reporting.}
For each tolerance level $t$, the evaluator therefore emits (i) $\mathrm{EM}$, (ii) the aggregate $\mathrm{mAP}^{(t)}$ from~\cref{eq:map_definition}, and (iii) three point-wise $\mathrm{AP}@\theta^{(t)}$ values at $\theta \in \{0.5, 0.75, 0.90\}$. All headline comparisons in the main text use $\mathrm{mAP}^{(\text{high})}$; EM and the AP slices are used only as complementary diagnostics.

\mypara{Interpretation.}
Reporting both EM and mAP provides complementary signals, where EM reflects strict, exact recovery, while mAP captures graded similarity.
For example, a model may achieve high mAP but low EM if it produces structurally correct outputs with small numerical errors. Conversely, a model may have non-trivial EM but low mAP if it occasionally outputs perfectly correct results but fails on most samples.

\subsection{Evaluation Setup}
\label{supp:subsec:evaluation_setup}
\vspace{-0.5em}

In this section, we introduce the detailed evaluation setup used in ChartArena. Our evaluation pipeline is strictly aligned with prior works~\cite{ChartX_ChartVLM_2025, OneChart_2024, StructChart_2026} to ensure fairness and reproducibility.

\mypara{Inference prompt.}
For document parsing MLLMs and expert chart understanding models, we use their officially recommended prompts for inference and configure the corresponding output formats according to each evaluation target. In particular, for our training backbone, we directly adopt the official parsing prompt released by HunyuanOCR\footnote{\url{https://github.com/Tencent-Hunyuan/HunyuanOCR}} as the direct parsing prompt. For CAP reasoning, we additionally append the following instruction to the original prompt:
\textit{``Please first analyze the chart information step by step, and then provide the final parsing result.''}

For general-purpose MLLMs, we use a \emph{unified prompt template} per chart family, shown in~\cref{prompt:eval_numeric_prompt,prompt:eval_logic_prompt,prompt:eval_flow_prompt}. The template is not tuned per model or on the test set. Instead, it is consolidated from two sources, namely the parsing prompts used in prior chart-parsing works~\cite{ChartX_ChartVLM_2025, OneChart_2024, StructChart_2026} and the officially recommended prompts released by each model vendor for structured extraction, and then \emph{frozen} before benchmark evaluation. The same frozen template is applied to every general-purpose MLLM.

\mypara{Cross-model inference protocol.}
To make the cross-model comparison as transparent as possible, we detail the exact inference protocol used for every model in~\cref{tab:ChartArena_main_results}:
\vspace{-0.5em}
\begin{itemize}[
        label=\raisebox{0.4ex}{\tiny$\bullet$},
        leftmargin=1em,
        itemsep=2pt,
        parsep=0pt,
        topsep=0pt,
        partopsep=0pt
    ]
    \item \textbf{Single-pass direct parsing.} Every score corresponds to a single forward pass per image. We do not use self-refinement, multi-turn feedback, retrieval, or tool use. The model receives the image and the frozen prompt and emits one response, from which the structured output is parsed.
    \item \textbf{Decoding parameters.} We call each model through its native OpenAI-compatible or vendor API with default decoding: temperature is left unset, no explicit \texttt{top\_p} or \texttt{max\_tokens} override is applied, and no logit bias is used. We keep decoding uniform rather than per-model tuned to avoid confounding language ability with generation-parameter engineering.
    \item \textbf{Retry policy.} A sample is retried when the API returns a transport-level error (timeout, HTTP $5\text{xx}$, or empty content). Retries reuse the same prompt and the same image.
    \item \textbf{Call and token budget.} For each of the $2{,}400$ benchmark images we issue \emph{one} inference call under the above protocol, yielding $2{,}400$ calls per general-purpose model in the main table. No token budget or cost cap is imposed at inference time.
    \item \textbf{Scope of the fairness claim.} We report cross-model numbers strictly under this fixed, single-pass, frozen-prompt setting, and interpret the resulting comparison as reflecting each model's ability to directly parse charts. Systems built on top of an MLLM with heavier inference-time machinery (self-refinement, tool use, agentic pipelines) are outside the scope of this comparison.
\end{itemize}

\begin{my_template_box}{Numeric Chart Parsing Prompt}
    \small
    Please parse the chart content in the image and extract the data into a structured Markdown \textbf{table} format.

    \vspace{4pt}
    \textbf{Requirements:}
    \begin{enumerate}
        \item Focus only on the chart itself and ignore unrelated elements such as decorations, backgrounds, logos, and watermarks;
        \item If both category labels and numerical units are present (e.g., axis labels), merge them into the table header using the format ``category-unit'';
        \item Preserve all category labels exactly as they appear in the chart without translation or rewriting;
        \item Preserve the original semantics and numerical precision of all values.
    \end{enumerate}
    \label{prompt:eval_numeric_prompt}
\end{my_template_box}

\begin{my_template_box}{Mind Map Parsing Prompt}
    \small
    Please parse the chart content in the image and extract the data into a structured Markdown \textbf{multi-level unordered list} format.

    \vspace{4pt}
    \textbf{Requirements:}
    \begin{enumerate}
        \item Use unordered lists beginning with `-', where each node text is represented as a list item;
        \item Determine the hierarchy according to the connection relationships between nodes, where parent nodes correspond to higher-level list items and child nodes correspond to nested list items;
        \item Fully extract all text contained in each node or box while preserving the original language and punctuation.
    \end{enumerate}
    \label{prompt:eval_logic_prompt}
\end{my_template_box}

\begin{my_template_box}{Flowchart Parsing Prompt}
    \small
    Please carefully analyze the following \textbf{flowchart} image and fully transcribe it into Mermaid flowchart code.

    \vspace{4pt}
    \textbf{Requirements:}
    \begin{enumerate}
        \item Use Mermaid \texttt{flowchart} or \texttt{graph} syntax (preferably \texttt{flowchart TD} or \texttt{flowchart LR} according to the actual direction of the diagram);
        \item Strictly preserve all node text, including the original language and punctuation, without translation, rewriting, or simplification;
        \item Correctly reconstruct all node connections and edge directions. If conditional branches exist, preserve branch labels such as ``Yes/No'';
        \item Ignore decorations, backgrounds, logos, watermarks, and other irrelevant visual elements;
        \item The output must strictly contain two stages:
              \begin{itemize}
                  \item First, declare all nodes line by line without edges;
                  \item Then, declare all relationships line by line without repeating node text.
              \end{itemize}
    \end{enumerate}
    \label{prompt:eval_flow_prompt}
\end{my_template_box}
\clearpage
\section{Further Evaluation Results}
\label{supp:sec:further_evaluation_results}
\vspace{-0.5em}

\subsection{Detailed Analysis under Different Visual Scenarios}
\label{supp:subsec:appendix_visual_scenarios}
\vspace{-0.5em}

In this section, we further analyze model performance under three visual scenarios in~\cref{tab:two_chart_three_visuals}, including digital rendering, printed photo, and hand-drawn photo.

\mypara{Impact of visual scenarios.}
Compared with digitally rendered charts, printed photos introduce additional visual difficulty caused by camera noise, illumination variation, blur, perspective distortion, and printing artifacts, leading to a consistent performance drop across nearly all models and chart categories. The hand-drawn photo scenario is substantially more challenging, since it additionally contains irregular strokes, imperfect structures, inconsistent layouts, and ambiguous visual boundaries. As a result, models generally exhibit a much larger degradation under hand-drawn settings, revealing limited robustness to severe visual uncertainty and distribution shift.

\mypara{Numeric charts vs. diagrammatic charts.}
Diagrammatic charts suffer much heavier degradation than numeric charts when moving from digital renderings to printed or hand-drawn settings. While numeric charts mostly depend on extracting geometric and numerical values, diagrammatic charts require recovering complex topological structures, like node-link relationships, directional arrows, and hierarchical layouts. Because these structures are fragile under real-world visual perturbations, robustly reconstructing diagrammatic charts is significantly harder for current models.

\mypara{Analysis across model categories.}
Across categories, the general trends align with \cref{tab:ChartArena_main_results}. Among general-purpose MLLMs, Gemini 3.1 Pro achieves top performance and overall robustness. On numeric charts, Seed2.0 Pro ranks second for digital and printed formats, while Kimi K2.5 takes second on hand-drawn photos; Seed2.0 Pro also leads the runner-up spot for diagrammatic charts.

Among expert models, our approach outperforms competitors across nearly all scenarios. Its strong performance under printed and hand-drawn conditions demonstrates that our training strategy improves both structural comprehension and resilience to real-world perturbations—underscoring that explicitly modeling chart semantics is essential for robust understanding beyond clean synthetic data.

\mypara{Document parsing models on numeric and diagrammatic charts.}
Document parsing MLLMs handle several numeric chart families reasonably well but remain weaker on diagrammatic structures. In~\cref{tab:ChartArena_main_results}, HunyuanOCR scores 70.8 EN / 72.5 ZH on pie charts, compared with 38.3 EN / 35.5 ZH on flowcharts and 55.0 EN / 46.6 ZH on mind maps. On the two diagrammatic families, it trails Gemini 3.1 Pro by 27.3 EN / 27.6 ZH and 31.8 EN / 38.6 ZH points, respectively. The pattern also appears in dots.mocr, whose flowchart and mind-map scores range from 19.6 to 28.7, while PaddleOCR-VL does not support these families in this evaluation. These results indicate that numeric extraction capability does not necessarily transfer to recovering node relations and hierarchies.

\begin{table*}[h!]
  \centering
  \captionsetup{font={small}}
  \caption{
    \textbf{Main results on two chart categories under three visual scenarios.}
    We report the average scores for numeric charts and diagrammatic charts under
    digital rendering, printed photo, and hand-drawn photo scenarios.
    The red subscript shows the difference from the first visual scenario within each chart category.
  }
  \vspace{-0.6em}
  {
    \renewcommand{\arraystretch}{1.0}
    \setlength{\tabcolsep}{8pt}
    \resizebox{\textwidth}{!}{%
      \begin{tabular}{clcccccc}
        \toprule
        \multirow{2}{*}[-2.8mm]
        {\textbf{\makecell{Model \nextline Type}}}  &
        \multirow{2}{*}[-2.8mm]
        {\textbf{Model}}                            &
        \multicolumn{3}{c}{\textbf{Numeric Charts}} &
        \multicolumn{3}{c}{\textbf{Diagrammatic charts}}
        \\
        \cmidrule(lr){3-5}
        \cmidrule(lr){6-8}
                                                    &
                                                    &
        \makecell{Digital \nextline Rendering}      &
        \makecell{Printed \nextline Photo}          &
        \makecell{Hand-drawn \nextline Photo}       &
        \makecell{Digital \nextline Rendering}      &
        \makecell{Printed \nextline Photo}          &
        \makecell{Hand-drawn \nextline Photo}
        \\
        \midrule
        \multirow{16}{*}
        {\makecell{General \nextline Purpose \nextline MLLMs}}
                                                    & \logo{openai}GPT-4o~\cite{GPT_4o_2024}                                                                   & 36.0          & \sd{34.3}{-1.7}          & \sd{30.5}{-5.5}           & 48.9          & \sd{45.0}{-3.9}          & \sd{30.9}{-18.0}          \\
                                                    & \logo{openai}GPT-5~\cite{GPT_5_2025}                                                                     & 45.2          & \sd{43.6}{-1.6}          & \sd{43.1}{-2.1}           & 60.2          & \sd{52.2}{-8.0}          & \sd{40.9}{-19.3}          \\
                                                    & \logo{intern}InternVL3.5-8B~\cite{InternVL3_5_2025}                                                      & 39.3          & \sd{34.1}{-5.2}          & \sd{31.7}{-7.6}           & 40.5          & \sd{38.1}{-2.4}          & \sd{24.0}{-16.5}          \\
                                                    & \logo{intern}InternVL3.5-241B-A28B~\cite{InternVL3_5_2025}                                               & 45.1          & \sd{40.7}{-4.4}          & \sd{40.5}{-4.6}           & 55.6          & \sd{49.6}{-6.0}          & \sd{36.4}{-19.2}          \\
                                                    & \logo{qwen}Qwen2.5-VL-7B-Ins.~\cite{Qwen2_5_VL_2025}                                                     & 31.1          & \sd{27.6}{-3.5}          & \sd{22.4}{-8.7}           & 37.5          & \sd{36.6}{-0.9}          & \sd{22.0}{-15.5}          \\
                                                    & \logo{qwen}Qwen2.5-VL-72B-Ins.~\cite{Qwen2_5_VL_2025}                                                    & 43.0          & \sd{39.4}{-3.6}          & \sd{37.8}{-5.2}           & 59.9          & \sd{58.7}{-1.2}          & \sd{40.7}{-19.2}          \\
                                                    & \logo{qwen}Qwen3-VL-8B-Ins.~\cite{Qwen3_VL_2025}                                                         & 49.5          & \sd{48.3}{-1.2}          & \sd{46.6}{-2.9}           & 63.1          & \sd{61.4}{-1.7}          & \sd{47.5}{-15.6}          \\
                                                    & \logo{qwen}Qwen3-VL-235B-A22B-Ins.~\cite{Qwen3_VL_2025}                                                  & 51.9          & \sd{49.7}{-2.2}          & \sd{46.3}{-5.6}           & 69.4          & \sd{65.9}{-3.5}          & \sd{47.4}{-22.0}          \\
                                                    & \logo{qwen}Qwen3.5-35B-A3B (think)~\cite{Qwen3_5_2026}                                                   & 61.8          & \sd{57.0}{-4.8}          & \sd{55.8}{-6.0}           & 74.9          & \sd{69.1}{-5.8}          & \sd{53.4}{-21.5}          \\
                                                    & \logo{glmv}GLM-4.5V~\cite{GLM_4_5V_2025}                                                                 & 52.8          & \sd{47.9}{-4.9}          & \sd{46.3}{-6.5}           & 56.8          & \sd{52.1}{-4.7}          & \sd{37.0}{-19.8}          \\
                                                    & \logo{bytedance}Seed1.8 (no-think)~\cite{Seed_1_8_2026}                                                  & 47.8          & \sd{46.8}{-1.0}          & \sd{44.6}{-3.2}           & 66.5          & \sd{64.6}{-1.9}          & \sd{51.7}{-14.8}          \\
                                                    & \logo{bytedance}Seed2.0 Pro (no-think)~\cite{Seed_2_0_2026}                                              & \ul{63.7}     & \sd{\ul{59.0}}{-4.7}     & \sd{55.5}{-8.2}           & \ul{81.8}     & \sd{\ul{75.7}}{-6.1}     & \sd{\ul{63.3}}{-18.5}     \\
                                                    & \logo{kimi}Kimi K2.5 (no-think)~\cite{Kimi_K2_5_2026}                                                    & 61.6          & \sd{58.2}{-3.4}          & \sd{\ul{56.6}}{-5.0}      & 76.0          & \sd{73.5}{-2.5}          & \sd{58.9}{-17.1}          \\
                                                    & \logo{xiaomimimo}MiMo-V2-Omni~\cite{Mimo_V2_Omni_2026}                                                   & 49.3          & \sd{45.6}{-3.7}          & \sd{44.7}{-4.6}           & 71.5          & \sd{65.0}{-6.5}          & \sd{50.6}{-20.9}          \\
                                                    & \logo{gemini}Gemini 2.5 Pro~\cite{Gemini_2_5_2025}                                                       & 54.4          & \sd{51.1}{-3.3}          & \sd{50.0}{-4.4}           & 70.5          & \sd{67.2}{-3.3}          & \sd{56.0}{-14.5}          \\
                                                    & \logo{gemini}Gemini 3.1 Pro~\cite{Gemini_3_1_2026}                                                       & \textbf{67.0} & \sd{\textbf{62.3}}{-4.7} & \sd{\textbf{60.4}}{-6.6}  & \textbf{83.0} & \sd{\textbf{78.3}}{-4.7} & \sd{\textbf{64.2}}{-18.8} \\
        \midrule
        \multirow{3}{*}
        {\makecell{Doc \nextline Parsing \nextline MLLMs}}
                                                    & \logo{xiaohongshu}dots.mocr (3B)~\cite{Dots_mOCR_2026}                                                   & 45.1          & \sd{41.0}{-4.1}          & \sd{\ul{32.3}}{-12.8}     & \ul{27.2}     & \sd{\ul{23.6}}{-3.6}     & \sd{\ul{18.2}}{-9.0}      \\
                                                    & \logo{paddle}PaddleOCR-VL (1B)~\cite{PaddleOCR_VL_2025}                                                  & \ul{50.2}     & \sd{\ul{42.3}}{-7.9}     & \sd{25.7}{-24.5}          & --            & --                       & --                        \\
                                                    & \logo{hunyuan}HunyuanOCR (1B)~\cite{HunyuanOCR_2025}                                                     & \textbf{56.3} & \sd{\textbf{49.7}}{-6.6} & \sd{\textbf{37.4}}{-18.9} & \textbf{55.0} & \sd{\textbf{50.4}}{-4.6} & \sd{\textbf{26.0}}{-29.0} \\
        \midrule
        \multirow{8}{*}
        {\makecell{Chart \nextline Experts}}
                                                    & \logo{opengvlab}ChartAst (13B)~\cite{ChartAssistant_2024}                                                & 2.6           & \sd{2.1}{-0.5}           & \sd{1.0}{-1.6}            & --            & --                       & --                        \\
                                                    & \logo{internscience}ChartVLM (8.3B)~\cite{ChartX_ChartVLM_2025}                                          & 9.1           & \sd{5.6}{-3.5}           & \sd{2.9}{-6.2}            & --            & --                       & --                        \\
                                                    & \logo{mPLUG}TinyChart (3B)~\cite{TinyChart_2024}                                                         & 4.8           & \sd{3.7}{-1.1}           & \sd{3.6}{-1.2}            & --            & --                       & --                        \\
                                                    & \logo{chartmoe}ChartMoE (8B)~\cite{ChartMoE_2024}                                                        & 20.7          & \sd{15.8}{-4.9}          & \sd{12.2}{-8.5}           & 2.9           & \sd{2.5}{-0.4}           & \sd{1.7}{-1.2}            \\
                                                    & \logo{thunlp}ChartCoder (7B)~\cite{ChartCoder_2025}                                                      & 17.7          & \sd{13.8}{-3.9}          & \sd{12.0}{-5.7}           & 2.5           & \sd{1.8}{-0.7}           & \sd{0.6}{-1.9}            \\
                                                    & \logo{shanghaiailab}RRVF (7B)~\cite{RRVF_2025}                                                           & \ul{46.2}     & \sd{\ul{38.3}}{-7.9}     & \sd{\ul{36.1}}{-10.1}     & \ul{59.5}     & \sd{\textbf{59.0}}{-0.5} & \sd{\ul{38.5}}{-21.0}     \\
                                                    & \logo{meituan}MSRL (7B)~\cite{MSRL_2025}                                                                 & 36.6          & \sd{35.6}{-1.0}          & \sd{31.6}{-5.0}           & 24.4          & \sd{21.4}{-3.0}          & \sd{18.6}{-5.8}           \\
                                                    & \multicolumn{1}{>{\columncolor{table_ours}[\tabcolsep][2\tabcolsep]}l}{\logo{hunyuan}\textbf{Ours} (1B)}
                                                    & \multicolumn{1}{>{\columncolor{table_ours}[0pt][2\tabcolsep]}c}{\textbf{64.1}}
                                                    & \multicolumn{1}{>{\columncolor{table_ours}[0pt][2\tabcolsep]}c}{\sd{\textbf{60.4}}{-3.7}}
                                                    & \multicolumn{1}{>{\columncolor{table_ours}[0pt][2\tabcolsep]}c}{\sd{\textbf{48.6}}{-15.5}}
                                                    & \multicolumn{1}{>{\columncolor{table_ours}[0pt][2\tabcolsep]}c}{\textbf{62.5}}
                                                    & \multicolumn{1}{>{\columncolor{table_ours}[0pt][2\tabcolsep]}c}{\sd{\ul{58.1}}{-4.4}}
                                                    & \multicolumn{1}{>{\columncolor{table_ours}[0pt][\tabcolsep]}c}{\sd{\textbf{40.9}}{-21.6}}                                                                                                                                                              \\
        \bottomrule
      \end{tabular}
    }
  }
  \label{tab:two_chart_three_visuals}
\end{table*}

\mypara{Remaining robustness gap on hand-drawn numeric charts.}
Zooming in on the digital-to-hand-drawn performance drop for numeric charts clarifies CAP's exact contribution. Compared to the HunyuanOCR backbone, which drops $-18.9$ points ($56.3\rightarrow 37.4$), our SFT-with-CAP model (row e, \cref{tab:ablation_training_stage}) reduces the gap to $-15.5$ ($64.1\rightarrow 48.6$)—effectively absorbing about $18\%$ of the loss. By forcing the model to infer atomic primitives (axes, entities, series, units) before parsing, CAP decouples numeric extraction from pixel-level recognition, improving transfer to hand-drawn images.

However, our remaining drop ($-15.5$) still lags behind frontier MLLMs like Gemini 3.1 Pro ($-6.6$, $67.0\rightarrow 60.4$). We attribute this residual gap to visual prior capacity tied to model scale rather than structural reasoning limits: parsing irregular strokes, uneven lighting, and non-uniform layouts is a general vision problem whose ceiling depends on the encoder's pre-trained prior. A ~1B specialized backbone simply cannot match the pixel-level robustness of massive MLLMs.

Ultimately, structural reasoning and parameter scale offer complementary paths to robustness: CAP offsets backbone-level degradation, while closing the remaining gap will require scaling the visual encoder or adding targeted hand-drawn augmentations.

\subsection{Analysis of the English--Chinese Performance Gap}
\label{supp:subsec:language_gap}
\vspace{-0.5em}

A recurring pattern in~\cref{tab:ChartArena_main_results} is that most general-purpose models score higher on the Chinese subset than on the English one, while several chart-specialized models show the opposite trend. A natural question is whether this reflects genuine differences in language ability, or differences in the intrinsic difficulty of the two subsets. We investigate this by measuring the structural complexity of each subset and comparing it against the observed score gap.

\mypara{Measuring subset difficulty.}
For every ground-truth annotation we compute an objective structural-complexity measure that matches the evaluation backend: the number of triples ($\text{rows}\times(\text{cols}-1)$) for numeric charts, and the number of nodes and edges for diagrammatic charts. For each chart type we then compute the relative difficulty gap between the two subsets, $(\mathrm{ZH}-\mathrm{EN})/\mathrm{EN}$, where a positive value means the Chinese subset is structurally harder. As summarized in~\cref{tab:lang_difficulty}, the two subsets are not balanced in difficulty and, importantly, the imbalance runs in opposite directions for the two chart families. The Chinese numeric charts are markedly simpler, with $53.9\%$ fewer triples on bar charts and $49.4\%$ fewer on combination charts, whereas the Chinese diagrammatic charts are more complex, with $18.3\%$ more edges on flowcharts and $30.9\%$ more nodes on mind maps.

\begin{table}[h!]
    \centering
    \captionsetup{font={small}}
    \caption{
        \textbf{Language score gap versus subset difficulty, per chart type.}
        \emph{Difficulty gap} is the relative structural-complexity difference, $(\mathrm{ZH}-\mathrm{EN})/\mathrm{EN}$ (measured by \#triples for numeric charts, \#nodes/\#edges for diagrammatic charts; positive means Chinese is harder). \emph{Score gap} is $\mathrm{ZH}-\mathrm{EN}$ in mAP$_{\text{high}}$ points averaged per model group (positive means Chinese scores higher). These quantities correlate strongly and negatively ($r=-0.80$ for general-purpose models): models score higher on the structurally easier language subset.
    }
    \label{tab:lang_difficulty}
    {
        \renewcommand{\arraystretch}{1.0}
        \setlength{\tabcolsep}{22pt}
        \resizebox{\columnwidth}{!}{
            \begin{tabular}{l c ccc}
                \toprule
                \multirow{2}{*}[-0.8mm]
                {\textbf{Chart type}}                                                                       &
                \multirow{2}{*}[-0.8mm]
                {\textbf{\makecell{Difficulty gap \nextline $(\mathrm{ZH}-\mathrm{EN})/\mathrm{EN}$ (\%)}}} &
                \multicolumn{3}{c}
                {\textbf{Score gap $(\mathrm{ZH}-\mathrm{EN})$}}                                                                               \\
                \cmidrule(lr){3-5}
                                                                                                            &       & General & Doc   & Expert \\
                \midrule
                Bar                                                                                         & -53.9 & +25.8   & +18.9 & +5.4   \\
                Line                                                                                        & +3.2  & +21.5   & +15.4 & +0.6   \\
                Pie                                                                                         & -4.0  & +2.2    & +9.6  & +10.2  \\
                Radar                                                                                       & -5.0  & +25.0   & +19.8 & +12.1  \\
                Box plot                                                                                    & -11.9 & +3.8    & +0.3  & +1.1   \\
                Comb. chart                                                                                 & -49.4 & +32.2   & +27.0 & +12.6  \\
                \midrule
                Flowchart                                                                                   & +18.3 & -7.2    & -4.2  & -7.4   \\
                Mind map                                                                                    & +30.9 & -13.0   & -8.8  & -8.4   \\
                \bottomrule
            \end{tabular}
        }
    }
\end{table}

\begin{figure}[t!]
    \centering
    \includegraphics[width=0.62\linewidth]{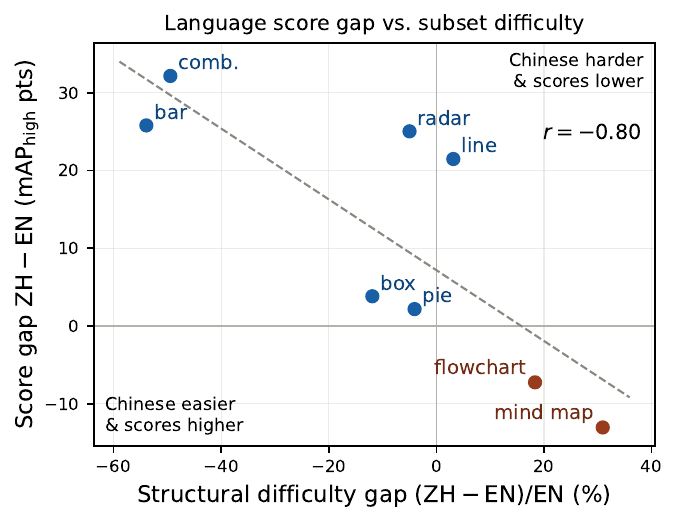}
    \vspace{-0.4em}
    \captionsetup{font={small}}
    \caption{
        \textbf{Language score gap tracks subset difficulty.}
        Each point represents a chart type. The $x$-axis shows the structural difficulty gap, $(\mathrm{ZH}-\mathrm{EN})/\mathrm{EN}$ (positive = Chinese is harder), while the $y$-axis shows the performance gap, $\mathrm{ZH}-\mathrm{EN}$ in mAP$_{\mathrm{high}}$ (positive = Chinese scores higher). Numeric charts (blue) cluster where Chinese is easier and scores higher; diagrammatic charts (orange) drop where Chinese is harder and scores lower. The strong negative correlation ($r = -0.80$) confirms this gap is mostly driven by subset difficulty asymmetry.
    }
    \label{fig:lang_difficulty_scatter}
    \vspace{-1.0em}
\end{figure}

\mypara{The score gap tracks the difficulty gap rather than language ability.}
When we place the score gap next to the difficulty gap, the two align almost perfectly in direction. Numeric charts, where the Chinese subset is easier, are exactly where Chinese scores are higher (a general-purpose gap of $+18.4$ points on average), while diagrammatic charts, where the Chinese subset is harder, are where Chinese scores are lower ($-10.1$ points on average). Across the eight chart types, the general-purpose score gap and the structural difficulty gap are strongly negatively correlated, with a Pearson correlation of $r=-0.80$ (\cref{fig:lang_difficulty_scatter}): models score higher precisely on whichever language subset happens to be structurally easier. This indicates that the English--Chinese gap is largely an artifact of the difficulty asymmetry between the two subsets rather than a pure language-ability effect.

\mypara{Why specialized models differ.}
This view also explains the contrasting trends among specialized models. Because general-purpose models cover all chart families, they fully inherit a large performance advantage on easier Chinese numeric charts, which dominate the benchmark. In contrast, chart-specialized models are mostly restricted to numeric charts and trained on English-centric corpora; their margin on the Chinese numeric subset is much smaller ($+7.0$ points vs. $+18.4$ for general models), which can invert the residual gap when accounting for weaker Chinese language priors. The apparent disagreement between the two model groups thus stems from a single underlying cause: the difficulty asymmetry of the subsets, modulated by each model's language prior.

\subsection{Fine-grained Analysis on Hard Samples}
\label{supp:subsec:hard_samples}
\vspace{-0.5em}

Aggregate benchmark scores may understate the benefits of CAP reasoning because a large fraction of samples are already well-solved after SFT, yielding identical scores for both settings. To provide a finer-grained view, we isolate a \emph{hard-sample subset} of ChartArena comprising $1{,}055$ samples whose per-sample mAP$_{\text{high}}$ falls below $0.5$ under our SFT baseline \emph{without} CAP (row (b) of~\cref{tab:ablation_training_stage}). This subset removes saturated samples to isolate cases where CAP reasoning can genuinely make a difference. Following the main evaluation protocol, we assess the SFT-with-CAP setting (row (e)) on both the full and hard subsets, reporting per-sample average mAP$_{\text{high}}$ in~\cref{tab:hard_sample_analysis}.

\mypara{CAP gain grows by roughly $7\times$ on the hard subset.}
While CAP yields a modest $+2.0$-point average gain across the full benchmark, it boosts performance on the hard subset from $7.1$ to $21.5$, a net jump of $+14.3$ points. This trend holds across all chart families: numeric charts rise from $+1.1$ to $+14.7$, mind maps from $+2.8$ to $+13.6$, and flowcharts from $+6.4$ to $+13.3$. The modest overall gain simply reflects the SFT baseline saturating on easy samples; once evaluated where headroom remains, CAP reasoning provides substantial, uniform improvements across families.

\mypara{Implication.}
This decomposition reveals that the value of CAP concentrates on structurally challenging cases rather than spreading uniformly, explaining why aggregate margins appear modest: they are diluted by easy samples where baseline parsers already succeed. This finer-grained view confirms that lightweight structural reasoning is key to closing the remaining gap on complex real-world charts.

\begin{table}[t!]
    \centering
    \captionsetup{font={small}}
    \caption{
        \textbf{Fine-grained analysis on hard samples.}
        Aggregate scores can mask CAP's benefits since easy samples are well solved after SFT. Isolating the \emph{hard-sample subset} of ChartArena ($1{,}055$ samples with mAP$_{\text{high}} < 0.5$ under standard SFT), adding CAP reasoning boosts the baseline from $7.1$ to $21.5$, a $+14.3$-point net gain compared to just $+2.0$ points on the full benchmark. This gain is roughly $7\times$ larger, with every chart family improving more on hard samples (ranging from $\sim 2\times$ for flowcharts to over $13\times$ for numeric charts).
    }
    \label{tab:hard_sample_analysis}
    {
        \renewcommand{\arraystretch}{0.95}
        \setlength{\tabcolsep}{16pt}
        \resizebox{\columnwidth}{!}{
            \begin{tabular}{l c ccc}
                \toprule
                \multirow{2}{*}[-0.8mm]
                {\textbf{Subset}}                      &
                \multirow{2}{*}[-0.8mm]
                {\textbf{\#Samples}}                   &
                \multicolumn{3}{c}{\textbf{mAP$_{\text{high}}$ ($\times100$)}}
                \\
                \cmidrule(lr){3-5}
                                                       &             & SFT w/o CAP & SFT w/ CAP & $\Delta$ (CAP gain) \\
                \midrule
                \multirow{2}{*}{All}                   & 2{,}400     & 53.42       & 55.40      & $+2.0$              \\
                                                       & Hard (1055) & 7.12        & 21.45      & \textbf{$+14.3$}    \\
                \midrule
                \multirow{2}{*}{Numeric charts}        & Full        & 55.75       & 56.86      & $+1.1$              \\
                                                       & Hard (762)  & 6.06        & 20.75      & \textbf{$+14.7$}    \\
                \midrule
                \multirow{2}{*}{Diagrammatic charts}   & Full        & 46.27       & 50.91      & $+4.6$              \\
                                                       & Hard (293)  & 9.86        & 23.28      & \textbf{$+13.4$}    \\
                \midrule
                \hspace{4mm}\multirow{2}{*}{Mind map}  & Full        & 53.96       & 56.74      & $+2.8$              \\
                                                       & Hard (124)  & 10.16       & 23.71      & \textbf{$+13.6$}    \\
                \midrule
                \hspace{4mm}\multirow{2}{*}{Flowchart} & Full        & 38.97       & 45.37      & $+6.4$              \\
                                                       & Hard (169)  & 9.64        & 22.96      & \textbf{$+13.3$}    \\
                \bottomrule
            \end{tabular}
        }
    }
    \vspace{-1.0em}
\end{table}

\subsection{Consistency Across Evaluation Metrics}
\label{supp:subsec:metric_consistency}
\vspace{-0.5em}

\begin{figure}[t!]
    \centering
    \begin{subfigure}{0.245\linewidth}
        \includegraphics[width=\linewidth]{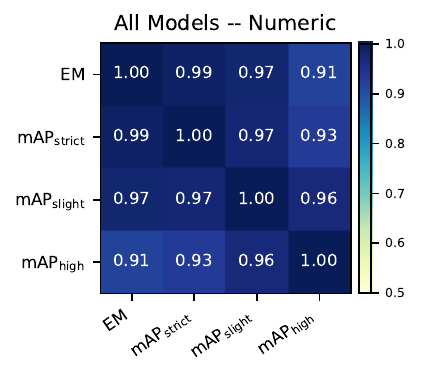}
        \vspace{-1.6em}
        \captionsetup{font={scriptsize}}
        \caption{All models, numeric}
        \label{fig:heatmap_all_numeric}
    \end{subfigure}
    \hfill
    \begin{subfigure}{0.245\linewidth}
        \includegraphics[width=\linewidth]{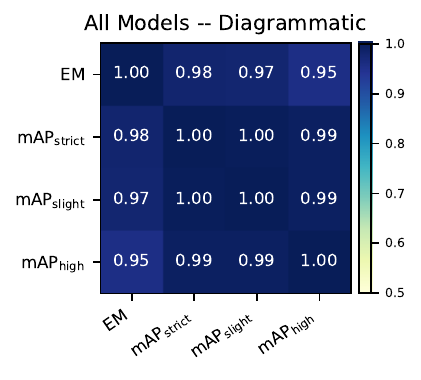}
        \vspace{-1.6em}
        \captionsetup{font={scriptsize}}
        \caption{All models, diagrammatic}
        \label{fig:heatmap_all_diagram}
    \end{subfigure}
    \hfill
    \begin{subfigure}{0.245\linewidth}
        \includegraphics[width=\linewidth]{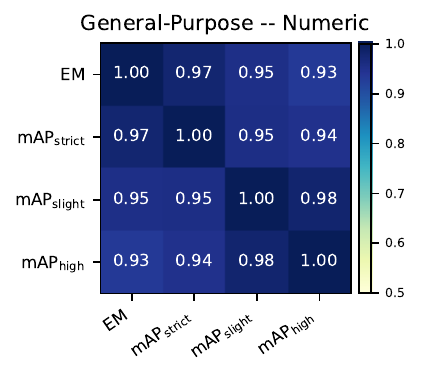}
        \vspace{-1.6em}
        \captionsetup{font={scriptsize}}
        \caption{General-purpose, numeric}
        \label{fig:heatmap_gp_numeric}
    \end{subfigure}
    \hfill
    \begin{subfigure}{0.245\linewidth}
        \includegraphics[width=\linewidth]{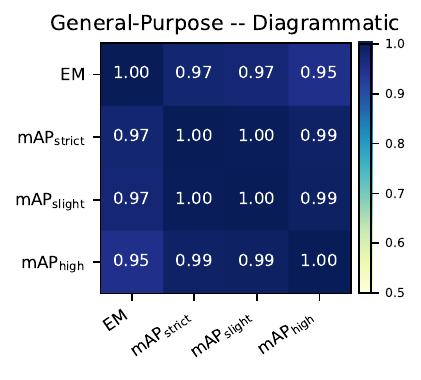}
        \vspace{-1.6em}
        \captionsetup{font={scriptsize}}
        \caption{General-purpose, diagrammatic}
        \label{fig:heatmap_gp_diagram}
    \end{subfigure}
    \vspace{-1.6em}
    \captionsetup{font={small}}
    \caption{
        \textbf{Cross-metric rank-correlation heatmaps.}
        Each panel displays the $4\times4$ Spearman rank correlation among the four evaluation metrics (EM, mAP$_{\text{strict}}$, mAP$_{\text{slight}}$, and mAP$_{\text{high}}$) computed across all models or the general-purpose subset for numeric and diagrammatic charts separately. All off-diagonal correlations are high ($\ge0.90$), confirming that model rankings are largely invariant to metric choice.
    }
    \label{fig:metric_heatmaps}
\end{figure}

The main paper adopts mAP$_{\text{high}}$ as the headline metric. Here we examine whether this choice biases the conclusions by reporting all four metrics in~\cref{tab:two_chart_four_metrics}, and by measuring the pairwise rank correlations among these metrics in~\cref{fig:metric_heatmaps}. This analysis directly addresses whether our unified evaluation is robust to the specific tolerance setting used for aggregation.

\mypara{Model ordering is stable across metrics.}
The relative ordering of models remains highly consistent across metrics. Every metric correlates strongly with mAP$_{\text{high}}$, showing Spearman correlations of $0.91$ (EM), $0.93$ (mAP$_{\text{strict}}$), and $0.96$ (mAP$_{\text{slight}}$) on numeric charts, and $0.95$, $0.99$, and $0.99$ on diagrammatic charts. Agreement is similarly tight within general-purpose models, with all correlations at or above $0.90$. The top of the leaderboard is virtually metric-invariant: Gemini 3.1 Pro ranks first across all metrics for both categories, and Seed2.0 Pro remains the leading general-purpose runner-up under tolerance-based metrics. This confirms that our core conclusions are independent of mAP$_{\text{high}}$, and that alternative, cheaper-to-report metrics convey the same qualitative findings.

\mypara{EM decays faster on diagrammatic charts, while graded metrics decay faster on numeric charts.}
While model rankings remain stable, metric ratios highlight a clear asymmetry between chart categories. Using metric-to-$\mathrm{mAP}_{\mathrm{high}}$ ratios to measure sensitivity to stricter criteria, exact match (EM) retains $0.24$ of $\mathrm{mAP}_{\mathrm{high}}$ on numeric charts but only $0.14$ on diagrammatic charts. EM penalizes diagrammatic charts far more heavily because a complete flowchart or mind map (including every node label and edge direction) must be recovered perfectly.

Conversely, relaxed metrics decay much more slowly on diagrammatic charts: the $\mathrm{mAP}_{\mathrm{slight}}$-to-$\mathrm{mAP}_{\mathrm{high}}$ ratio is $0.74$ compared to $0.55$ for numeric charts, a pattern that also holds for $\mathrm{mAP}_{\mathrm{strict}}$ ($0.41$ vs. $0.30$). This shift stems from fundamentally different failure modes. Numeric values are scored element-wise, so tiny per-cell inaccuracies quickly compound under strict tolerances. Graph matching, on the other hand, awards partial credit for recovered subgraphs, letting scores degrade far more gracefully as long as the overall skeleton is intact.

\mypara{Implications for metric choice.}
Together, these results justify using mAP$_{\text{high}}$ as the main metric: it preserves model rankings while offering a wider dynamic range to better separate models. Meanwhile, divergent decay patterns highlight the value of strict metrics as a complementary signal, exposing a structural recovery gap that is consistently larger for diagrammatic charts than for numeric ones.

\begin{table*}[t!]
  \centering
  \captionsetup{font={small}}
  \caption{
    \textbf{Results on two chart categories under four evaluation metrics.}
    We report \textbf{EM} and \textbf{mAP} at three tolerance levels (strict, slight, high) for numeric and diagrammatic charts, aggregated over all visual scenarios. Models that lack diagrammatic capability are marked ``--''. Within each model category, \textbf{bold} and \underline{underline} denote the best and second-best results per column.
  }
  \vspace{-0.4em}
  {
    \renewcommand{\arraystretch}{1.0}
    \setlength{\tabcolsep}{6pt}
    \resizebox{\textwidth}{!}{
      \begin{tabular}{cl cccc cccc}
        \toprule
        \multirow{2}{*}[-0.8mm]
        {\textbf{\makecell{Model \nextline Type}}}  &
        \multirow{2}{*}[-0.8mm]
        {\textbf{Model}}                            &
        \multicolumn{4}{c}{\textbf{Numeric Charts}} &
        \multicolumn{4}{c}{\textbf{Diagrammatic Charts}}
        \\
        \cmidrule(lr){3-6}
        \cmidrule(lr){7-10}
                                                    &                                                                 &
        EM                                          & mAP$_{\text{strict}}$                                           & mAP$_{\text{slight}}$ & mAP$_{\text{high}}$ &
        EM                                          & mAP$_{\text{strict}}$                                           & mAP$_{\text{slight}}$ & mAP$_{\text{high}}$
        \\
        \midrule
        \multirow{16}{*}
        {\makecell{General \nextline Purpose \nextline MLLMs}}
                                                    & \logo{openai}GPT-4o~\cite{GPT_4o_2024}                          & 8.5                   & 10.8                & 19.1          & 33.6          & 4.3           & 15.5          & 30.3          & 41.4          \\
                                                    & \logo{openai}GPT-5~\cite{GPT_5_2025}                            & 11.8                  & 14.6                & 24.9          & 44.0          & 7.9           & 22.7          & 41.4          & 50.9          \\
                                                    & \logo{intern}InternVL3.5-8B~\cite{InternVL3_5_2025}             & 8.1                   & 10.5                & 19.0          & 35.0          & 3.2           & 10.2          & 22.1          & 34.1          \\
                                                    & \logo{intern}InternVL3.5-241B-A28B~\cite{InternVL3_5_2025}      & 13.2                  & 16.2                & 26.2          & 42.1          & 6.2           & 17.5          & 34.7          & 47.1          \\
                                                    & \logo{qwen}Qwen2.5-VL-7B-Ins.~\cite{Qwen2_5_VL_2025}            & 6.6                   & 8.3                 & 13.6          & 27.0          & 3.6           & 10.5          & 20.1          & 32.0          \\
                                                    & \logo{qwen}Qwen2.5-VL-72B-Ins.~\cite{Qwen2_5_VL_2025}           & 8.6                   & 10.6                & 22.2          & 40.0          & 5.6           & 20.4          & 39.7          & 53.1          \\
                                                    & \logo{qwen}Qwen3-VL-8B-Ins.~\cite{Qwen3_VL_2025}                & 13.8                  & 16.7                & 25.8          & 40.7          & 6.6           & 21.9          & 39.7          & 53.1          \\
                                                    & \logo{qwen}Qwen3-VL-235B-A22B-Ins.~\cite{Qwen3_VL_2025}         & \ul{17.9}             & 21.2                & 33.6          & 49.3          & 9.1           & 27.6          & 47.6          & 60.9          \\
                                                    & \logo{qwen}Qwen3.5-35B-A3B (think)~\cite{Qwen3_5_2026}          & 17.1                  & 21.1                & 37.5          & 57.6          & 9.7           & 30.9          & 54.5          & 66.2          \\
                                                    & \logo{glmv}GLM-4.5V~\cite{GLM_4_5V_2025}                        & 14.1                  & 17.9                & 31.9          & 49.0          & 6.8           & 20.1          & 37.8          & 49.6          \\
                                                    & \logo{bytedance}Seed1.8 (no-think)~\cite{Seed_1_8_2026}         & 14.5                  & 17.5                & 28.8          & 46.4          & 9.3           & 28.0          & 49.7          & 60.9          \\
                                                    & \logo{bytedance}Seed2.0 Pro (no-think)~\cite{Seed_2_0_2026}     & 17.8                  & 21.0                & \ul{38.0}     & 57.7          & \ul{14.2}     & \ul{41.3}     & \ul{64.8}     & \ul{71.9}     \\
                                                    & \logo{kimi}Kimi K2.5 (no-think)~\cite{Kimi_K2_5_2026}           & 17.6                  & \ul{21.6}           & 37.4          & \ul{58.8}     & 10.9          & 36.1          & 59.9          & 69.5          \\
                                                    & \logo{xiaomimimo}MiMo-V2-Omni~\cite{Mimo_V2_Omni_2026}          & 13.5                  & 16.4                & 28.6          & 46.5          & 9.0           & 28.1          & 51.4          & 62.3          \\
                                                    & \logo{gemini}Gemini 2.5 Pro~\cite{Gemini_2_5_2025}              & 17.6                  & 20.9                & 34.5          & 51.8          & 11.7          & 31.2          & 54.8          & 64.7          \\
                                                    & \logo{gemini}Gemini 3.1 Pro~\cite{Gemini_3_1_2026}              & \textbf{24.5}         & \textbf{28.7}       & \textbf{45.4} & \textbf{63.2} & \textbf{15.8} & \textbf{42.8} & \textbf{67.6} & \textbf{75.2} \\
        \midrule
        \multirow{3}{*}
        {\makecell{Document \nextline Parsing \nextline MLLMs}}
                                                    & \logo{xiaohongshu}dots.mocr (3B)~\cite{Dots_mOCR_2026}          & \ul{6.9}              & \textbf{10.6}       & \ul{21.0}     & \ul{41.6}     & \ul{1.5}      & \ul{6.1}      & \ul{9.9}      & \ul{23.8}     \\
                                                    & \logo{paddle}PaddleOCR-VL (1B)~\cite{PaddleOCR_VL_2025}         & 6.3                   & 9.1                 & 17.8          & 38.3          & --            & --            & --            & --            \\
                                                    & \logo{hunyuan}HunyuanOCR (1B)~\cite{HunyuanOCR_2025}            & \textbf{7.4}          & \ul{10.3}           & \textbf{24.5} & \textbf{47.3} & \textbf{7.1}  & \textbf{19.3} & \textbf{34.2} & \textbf{43.8} \\
        \midrule
        \multirow{7}{*}
        {\makecell{Expert Chart \nextline Understanding \nextline Models}}
                                                    & \logo{opengvlab}ChartAst (13B)~\cite{ChartAssistant_2024}       & 0.0                   & 0.0                 & 0.6           & 1.9           & --            & --            & --            & --            \\
                                                    & \logo{internscience}ChartVLM (8.3B)~\cite{ChartX_ChartVLM_2025} & 0.2                   & 0.6                 & 1.8           & 5.8           & --            & --            & --            & --            \\
                                                    & \logo{mPLUG}TinyChart (3B)~\cite{TinyChart_2024}                & 0.1                   & 0.3                 & 0.9           & 3.9           & --            & --            & --            & --            \\
                                                    & \logo{chartmoe}ChartMoE (8B)~\cite{ChartMoE_2024}               & 0.7                   & 1.8                 & 6.6           & 16.0          & 0.1           & 0.4           & 0.8           & 2.3           \\
                                                    & \logo{thunlp}ChartCoder (7B)~\cite{ChartCoder_2025}             & 1.6                   & 2.7                 & 5.8           & 14.5          & 0.0           & 0.1           & 0.2           & 1.6           \\
                                                    & \logo{shanghaiailab}RRVF (7B)~\cite{RRVF_2025}                  & \textbf{7.8}          & \textbf{10.8}       & \textbf{18.7} & \textbf{41.3} & \textbf{5.9}  & \textbf{18.4} & \textbf{32.7} & \textbf{50.3} \\
                                                    & \logo{meituan}MSRL (7B)~\cite{MSRL_2025}                        & \ul{2.6}              & \ul{4.2}            & \ul{9.8}      & \ul{33.9}     & \ul{1.2}      & \ul{4.2}      & \ul{7.8}      & \ul{21.3}     \\
        \bottomrule
      \end{tabular}
    }
  }
  \vspace{-0.4em}
  \label{tab:two_chart_four_metrics}
\end{table*}

\subsection{Node/Edge Weight Sensitivity in Flowchart Scoring}
\label{supp:subsec:weight_sensitivity}
\vspace{-0.5em}

\begin{figure}[t!]
    \centering
    \begin{subfigure}{0.42\linewidth}
        \includegraphics[width=\linewidth]{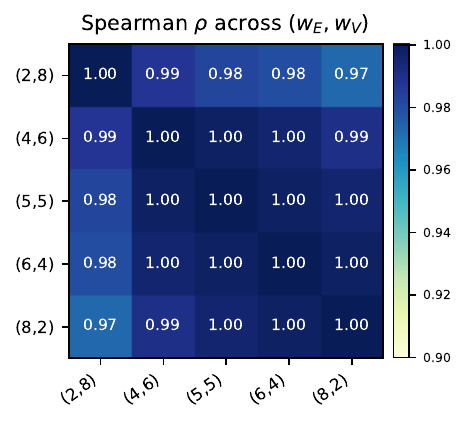}
        \vspace{-1.6em}
        \captionsetup{font={small}}
        \caption{Rank correlation across weightings}
        \label{fig:weight_heatmap}
    \end{subfigure}
    \hfill
    \begin{subfigure}{0.50\linewidth}
        \includegraphics[width=\linewidth]{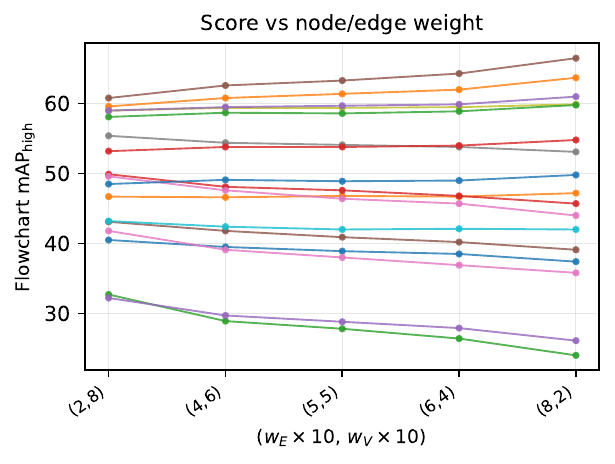}
        \vspace{-1.6em}
        \captionsetup{font={small}}
        \caption{mAP$_{\mathrm{high}}$ vs. node/edge weight}
        \label{fig:weight_lines}
    \end{subfigure}
    \caption{
        \textbf{Node/edge weight sensitivity for flowchart scoring.}
        (a) $5\times5$ Spearman rank correlation among five $(w_E, w_V)$ settings over the same 17 models as~\cref{tab:weight_sensitivity}; all off-diagonal values are $\ge0.97$. (b) Flowchart mAP$_{\mathrm{high}}$ as a function of the weighting for the same models; the curves rarely cross, confirming a stable ordering, while stronger models improve as more weight shifts to edges.
    }
    \label{fig:weight_sensitivity}
    \vspace{-1.0em}
\end{figure}

The flowchart score in~\cref{supp:subsec:graph_scoring} combines a node-matching term and an edge-matching term through a weighted sum with the default $(w_E, w_V) = (0.6, 0.4)$. To verify that this choice does not bias the reported comparisons, we re-evaluate all flowchart-capable models under five weightings, $(w_E, w_V) \in \{(0.2,0.8), (0.4,0.6), (0.5,0.5), (0.6,0.4), (0.8,0.2)\}$, keeping the rest of the protocol fixed and scoring every model on its Mermaid output. The per-weighting scores and their rank correlations with the default setting are reported in~\cref{tab:weight_sensitivity}, and visualized in~\cref{fig:weight_sensitivity}.

\mypara{Model ranking is stable across all node/edge weightings.}
The relative ordering of models barely changes as the weight moves from node-dominated to edge-dominated matching. Every column correlates strongly with the default $(0.6, 0.4)$ ranking, with Spearman $\rho \ge 0.98$ and Kendall $\tau \ge 0.91$ even at the two extremes, and $\rho \ge 0.99$ for all intermediate settings; the pairwise Spearman correlation among any two weightings never drops below $0.97$. Gemini 3.1 Pro remains the single best model under all five weightings, and the strong and weak groups stay clearly separated throughout. The choice of $(w_E, w_V)$ therefore does not alter any conclusion drawn from the flowchart results.

\mypara{The score differences reveal that edges are harder and more discriminative.}
Although rankings remain stable, score magnitudes shift systematically with weighting based on model capability. Moving weight toward edges raises scores for strong models, such as Gemini 3.1 Pro ($60.8 \rightarrow 66.5$) and Seed2.0 Pro ($59.6 \rightarrow 63.7$), while lowering them for weaker ones, like Qwen2.5-VL-7B ($32.7 \rightarrow 24.0$). Across models, edge-induced gains (the difference between $(0.8,0.2)$ and $(0.2,0.8)$ weightings) correlate strongly with average scores ($r = 0.87$). This indicates that recovering edge topology (i.e., directed connectivity in flowcharts) is harder than recognizing isolated nodes, and that edge reconstruction capability is precisely what separates strong parsers from weak ones.

\mypara{This pattern supports weighting edges more heavily by default.}
Because edge correctness is both harder to achieve and more discriminative of true parsing ability, a metric that emphasizes edges is more faithful to the practical difficulty of flowchart parsing, where a single misrouted edge can invalidate the logic of the whole diagram. Our default $(w_E, w_V) = (0.6, 0.4)$ reflects this by placing more weight on edges than on nodes, while remaining well within the stable region identified above. The sensitivity analysis thus confirms both that the reported rankings are robust to the weighting and that the default weighting is a well-motivated design choice rather than an arbitrary one.

\begin{table}[t!]
  \centering
  \captionsetup{font={small}}
  \caption{
    \textbf{Node/edge weight sensitivity on flowcharts.}
    Flowchart mAP$_{\text{high}}$ under five node/edge weightings $(w_E\!\times\!10,\, w_V\!\times\!10)$. The paper default is $(w_E,w_V)=(0.6,0.4)$. The bottom rows report the Spearman and Kendall rank correlation of each column against the default ranking.
  }
  \label{tab:weight_sensitivity}
  {
    \renewcommand{\arraystretch}{1.0}
    \setlength{\tabcolsep}{20pt}
    \resizebox{\columnwidth}{!}{
      \begin{tabular}{l ccccc}
        \toprule
        \textbf{Model}                                             &
        $(2,8)$                                                    &
        $(4,6)$                                                    &
        $(5,5)$                                                    &
        $(6,4)$                                                    &
        $(8,2)$
        \\
        \cmidrule(lr){1-1}
        \cmidrule(lr){2-6}
        \logo{openai}GPT-4o~\cite{GPT_4o_2024}                     & 40.5  & 39.5  & 38.9  & 38.5  & 37.4  \\
        \logo{openai}GPT-5~\cite{GPT_5_2025}                       & 46.7  & 46.6  & 46.8  & 46.7  & 47.2  \\
        \logo{qwen}Qwen2.5-VL-7B-Ins.~\cite{Qwen2_5_VL_2025}       & 32.7  & 28.9  & 27.8  & 26.4  & 24.0  \\
        \logo{qwen}Qwen2.5-VL-72B-Ins.~\cite{Qwen2_5_VL_2025}      & 49.9  & 48.1  & 47.6  & 46.8  & 45.7  \\
        \logo{intern}InternVL3.5-8B~\cite{InternVL3_5_2025}        & 32.2  & 29.7  & 28.8  & 27.9  & 26.1  \\
        \logo{intern}InternVL3.5-241B-A28B~\cite{InternVL3_5_2025} & 43.1  & 41.8  & 40.9  & 40.2  & 39.1  \\
        \logo{qwen}Qwen3-VL-8B-Ins.~\cite{Qwen3_VL_2025}           & 49.6  & 47.6  & 46.4  & 45.7  & 44.0  \\
        \logo{qwen}Qwen3-VL-235B-A22B-Ins.~\cite{Qwen3_VL_2025}    & 55.4  & 54.4  & 54.1  & 53.8  & 53.1  \\
        \logo{qwen}Qwen3.5-35B-A3B~\cite{Qwen3_5_2026}             & 59.0  & 59.4  & 59.4  & 59.5  & 59.9  \\
        \logo{glmv}GLM-4.5V~\cite{GLM_4_5V_2025}                   & 43.2  & 42.4  & 42.0  & 42.1  & 42.0  \\
        \logo{bytedance}Seed1.8~\cite{Seed_1_8_2026}               & 48.5  & 49.1  & 48.9  & 49.0  & 49.8  \\
        \logo{bytedance}Seed2.0 Pro~\cite{Seed_2_0_2026}           & 59.6  & 60.8  & 61.4  & 62.0  & 63.7  \\
        \logo{kimi}Kimi K2.5~\cite{Kimi_K2_5_2026}                 & 58.1  & 58.7  & 58.6  & 58.9  & 59.8  \\
        \logo{xiaomimimo}MiMo-V2-Omni~\cite{Mimo_V2_Omni_2026}     & 53.2  & 53.8  & 53.8  & 54.0  & 54.8  \\
        \logo{gemini}Gemini 2.5 Pro~\cite{Gemini_2_5_2025}         & 59.0  & 59.5  & 59.7  & 59.9  & 61.0  \\
        \logo{gemini}Gemini 3.1 Pro~\cite{Gemini_3_1_2026}         & 60.8  & 62.6  & 63.3  & 64.3  & 66.5  \\
        \logo{hunyuan}HunyuanOCR~\cite{HunyuanOCR_2025}            & 41.8  & 39.1  & 38.0  & 36.9  & 35.8  \\
        \midrule
        Spearman $\rho$ vs.\ $(6,4)$                               & 0.980 & 0.995 & 0.998 & 1.000 & 0.998 \\
        Kendall $\tau$ vs.\ $(6,4)$                                & 0.908 & 0.971 & 0.985 & 1.000 & 0.985 \\
        \bottomrule
      \end{tabular}
    }
  }
  \vspace{-1.0em}
\end{table}

\subsection{Statistical Significance}
\label{supp:subsec:bootstrap}
\vspace{-0.5em}

To quantify how much of the reported differences could be explained by sampling noise, we run a sample-level bootstrap on the per-sample mAP$_{\text{high}}$ scores and report the results in~\cref{tab:bootstrap_ci}.

\mypara{Protocol.}
To ensure a fair comparison, we enforce a single output format per chart family (Markdown for numeric charts and mind maps; Mermaid for flowcharts) across all $2{,}400$ ChartArena samples. We compute $95\%$ confidence intervals via bootstrapping ($B=1{,}000$ with replacement, seed $42$) using the $2.5$th and $97.5$th percentiles. For paired comparisons, we apply identical resampled indices to both models to cancel out sample-difficulty variance.

\mypara{Sampling noise is small relative to the model gaps.}
The 95\% intervals are about $3$ to $4$ points wide, far narrower than the gaps between performance tiers. Gemini 3.1 Pro forms a tier of its own, and its interval does not overlap that of Kimi K2.5. Our model is statistically indistinguishable from Gemini 2.5 Pro, and its interval clearly separates from that of the HunyuanOCR backbone.

\mypara{The core claims hold under paired resampling.}
Our model outperforms the HunyuanOCR backbone by $+8.93$ points ($[+7.31, +10.29]$), while CAP adds $+1.99$ points ($[+0.74, +3.29]$) over the SFT baseline. Because neither interval includes zero, these gains are robust against sampling noise. Meanwhile, the performance gaps relative to Gemini 3.1 Pro and Kimi K2.5 remain significant, aligning with the parameter-scale differences discussed in~\cref{supp:subsec:appendix_visual_scenarios}.

\begin{table*}[t!]
    \centering
    \captionsetup{font={small}}
    \caption{
        \textbf{Capability-conditioned averages and bootstrap 95\% confidence intervals.}
        Capability-conditioned averages exclude chart families marked ``--'' from the denominator. We report mAP$_{\text{high}}$ averaged over all samples, along with $95\%$ intervals from $1{,}000$ sample-level bootstrap resamples. The bottom block presents paired bootstrap intervals on $\Delta$mAP$_{\text{high}}$ using identical resampled indices. Bootstrap point estimates use one fixed output format per chart family (\cref{supp:subsec:bootstrap}). Bold paired intervals exclude zero.
    }
    \label{tab:bootstrap_ci}
    \vspace{-0.4em}
    {
        \renewcommand{\arraystretch}{1.1}
        \setlength{\tabcolsep}{14pt}
        \resizebox{\textwidth}{!}{
            \begin{tabular}{c l cc cc}
                \toprule
                \multirow{2}{*}[-0.8mm]{\textbf{\makecell{Model \nextline Type}}}                                        &
                \multirow{2}{*}[-0.8mm]{\textbf{Model}}                                                                  &
                \multicolumn{2}{c}{\textbf{Average (capability)}}                                                        &
                \multirow{2}{*}[-0.8mm]{\textbf{Bootstrap mAP$_{\text{high}}$}}                                          &
                \multirow{2}{*}[-0.8mm]{\textbf{95\% CI}}                                                                                                                                                                                                                             \\
                \cmidrule(lr){3-4}
                                                                                                                         &
                                                                                                                         &
                EN                                                                                                       &
                ZH                                                                                                       &
                                                                                                                         &
                \\
                \midrule
                \multirow{16}{*}{\makecell{General                                                                                                                                                                                                                                    \\ Purpose \\ MLLMs}} & \logo{openai}GPT-4o~\cite{GPT_4o_2024} & 34.8 & 36.3 & 35.63 & [33.97, 37.32] \\
                                                                                                                         & \logo{openai}GPT-5~\cite{GPT_5_2025}                            & 45.6                                 & 45.7                     & 45.79 & [43.91, 47.60] \\
                                                                                                                         & \logo{intern}InternVL3.5-8B~\cite{InternVL3_5_2025}             & 29.3                                 & 40.2                     & 34.75 & [33.05, 36.46] \\
                                                                                                                         & \logo{intern}InternVL3.5-241B-A28B~\cite{InternVL3_5_2025}      & 38.1                                 & 48.6                     & 43.35 & [41.60, 45.13] \\
                                                                                                                         & \logo{qwen}Qwen2.5-VL-7B-Ins.~\cite{Qwen2_5_VL_2025}            & 23.3                                 & 33.2                     & 28.31 & [26.64, 29.92] \\
                                                                                                                         & \logo{qwen}Qwen2.5-VL-72B-Ins.~\cite{Qwen2_5_VL_2025}           & 36.6                                 & 50.0                     & 43.27 & [41.63, 45.02] \\
                                                                                                                         & \logo{qwen}Qwen3-VL-8B-Ins.~\cite{Qwen3_VL_2025}                & 37.3                                 & 50.4                     & 43.84 & [42.11, 45.69] \\
                                                                                                                         & \logo{qwen}Qwen3-VL-235B-A22B-Ins.~\cite{Qwen3_VL_2025}         & 46.0                                 & 58.4                     & 52.17 & [50.48, 53.98] \\
                                                                                                                         & \logo{qwen}Qwen3.5-35B-A3B (think)~\cite{Qwen3_5_2026}          & 54.1                                 & 65.5                     & 59.80 & [58.08, 61.51] \\
                                                                                                                         & \logo{glmv}GLM-4.5V~\cite{GLM_4_5V_2025}                        & 43.8                                 & 53.9                     & 48.96 & [47.26, 50.80] \\
                                                                                                                         & \logo{bytedance}Seed1.8 (no-think)~\cite{Seed_1_8_2026}         & 42.1                                 & 57.8                     & 50.00 & [48.23, 51.80] \\
                                                                                                                         & \logo{bytedance}Seed2.0 Pro (no-think)~\cite{Seed_2_0_2026}     & 54.0                                 & \ul{70.5}                & 62.25 & [60.55, 63.92] \\
                                                                                                                         & \logo{kimi}Kimi K2.5 (no-think)~\cite{Kimi_K2_5_2026}           & \ul{54.8}                            & 68.1                     & 61.44 & [59.64, 63.12] \\
                                                                                                                         & \logo{xiaomimimo}MiMo-V2-Omni~\cite{Mimo_V2_Omni_2026}          & 43.9                                 & 57.0                     & 50.43 & [48.60, 52.16] \\
                                                                                                                         & \logo{gemini}Gemini 2.5 Pro~\cite{Gemini_2_5_2025}              & 47.7                                 & 62.4                     & 55.06 & [53.34, 56.83] \\
                                                                                                                         & \logo{gemini}Gemini 3.1 Pro~\cite{Gemini_3_1_2026}              & \textbf{59.2}                        & \textbf{73.2}            & 66.15 & [64.59, 67.76] \\
                \midrule
                \multirow{3}{*}{\makecell{Document                                                                                                                                                                                                                                    \\ Parsing \\ MLLMs}} & \logo{xiaohongshu}dots.mocr (3B)~\cite{Dots_mOCR_2026} & \ul{33.1} & 40.7 & -- & -- \\
                                                                                                                         & \logo{paddle}PaddleOCR-VL (1B)~\cite{PaddleOCR_VL_2025}         & 29.9                                 & \ul{46.6}                & --    & --             \\
                                                                                                                         & \logo{hunyuan}HunyuanOCR (1B)~\cite{HunyuanOCR_2025}            & \textbf{41.0}                        & \textbf{51.8}            & 46.51 & [44.71, 48.36] \\
                \midrule
                \multirow{8}{*}{\makecell{Expert Chart                                                                                                                                                                                                                                \\ Understanding \\ Models}} & \logo{opengvlab}ChartAst (13B)~\cite{ChartAssistant_2024} & 1.9 & -- & -- & -- \\
                                                                                                                         & \logo{internscience}ChartVLM (8.3B)~\cite{ChartX_ChartVLM_2025} & 7.1                                  & 4.6                      & --    & --             \\
                                                                                                                         & \logo{mPLUG}TinyChart (3B)~\cite{TinyChart_2024}                & 3.8                                  & 4.0                      & --    & --             \\
                                                                                                                         & \logo{chartmoe}ChartMoE (8B)~\cite{ChartMoE_2024}               & 8.5                                  & 22.1                     & --    & --             \\
                                                                                                                         & \logo{thunlp}ChartCoder (7B)~\cite{ChartCoder_2025}             & 12.6                                 & 13.2                     & --    & --             \\
                                                                                                                         & \logo{shanghaiailab}RRVF (7B)~\cite{RRVF_2025}                  & \ul{36.1}                            & \ul{51.0}                & --    & --             \\
                                                                                                                         & \logo{meituan}MSRL (7B)~\cite{MSRL_2025}                        & 27.1                                 & 34.4                     & --    & --             \\
                \rowcolor{table_ours}
                                                                                                                         & \logo{hunyuan}\textbf{Ours} (1B)                                & \textbf{49.5}                        & \textbf{61.3}            & 55.44 & [53.59, 57.16] \\
                \midrule
                \multicolumn{2}{l}{\textbf{Paired comparison}}                                                           & \multicolumn{2}{c}{}                                            & \textbf{$\Delta$mAP$_{\text{high}}$} & \textbf{95\% CI}                                  \\
                \midrule
                \multicolumn{2}{l}{\textbf{Ours} (1B) $-$ HunyuanOCR (1B)~\cite{HunyuanOCR_2025}}                        & \multicolumn{2}{c}{}                                            & +8.93                                & \textbf{[+7.31, +10.29]}                          \\
                \multicolumn{2}{l}{\textbf{Ours} (1B) $-$ SFT w/o CAP, row (b) of~\cref{tab:ablation_training_stage}}    & \multicolumn{2}{c}{}                                            & +1.99                                & \textbf{[+0.74, +3.29]}                           \\
                \multicolumn{2}{l}{\textbf{Ours} (1B) $-$ Gemini 3.1 Pro~\cite{Gemini_3_1_2026}}                         & \multicolumn{2}{c}{}                                            & -10.71                               & \textbf{[-12.52, -9.07]}                          \\
                \multicolumn{2}{l}{\textbf{Ours} (1B) $-$ Kimi K2.5 (no-think)~\cite{Kimi_K2_5_2026}}                    & \multicolumn{2}{c}{}                                            & -6.00                                & \textbf{[-7.72, -4.24]}                           \\
                \multicolumn{2}{l}{\textbf{Ours} (1B) $-$ Gemini 2.5 Pro~\cite{Gemini_2_5_2025}}                         & \multicolumn{2}{c}{}                                            & +0.37                                & [-1.53, +2.04]                                    \\
                \multicolumn{2}{l}{\textbf{Ours} (1B) $-$ Qwen3-VL-235B-A22B-Ins.~\cite{Qwen3_VL_2025}}                  & \multicolumn{2}{c}{}                                            & +3.27                                & \textbf{[+1.55, +4.93]}                           \\
                \multicolumn{2}{l}{Gemini 3.1 Pro~\cite{Gemini_3_1_2026} $-$ Kimi K2.5 (no-think)~\cite{Kimi_K2_5_2026}} & \multicolumn{2}{c}{}                                            & +4.71                                & \textbf{[+3.39, +6.22]}                           \\
                \bottomrule
            \end{tabular}
        }
    }
    \vspace{-1.2em}
\end{table*}

\subsection{Additional Model Results}
\label{supp:subsec:additional_results}
\vspace{-0.5em}

To improve transparency and broaden comparability across model scales, we evaluate additional models in~\cref{tab:ChartArena_additional_results}.
These models include multiple size variants from InternVL3.5~\cite{InternVL3_5_2025} and Qwen3.5~\cite{Qwen3_5_2026}, as well as Gemma~4-31B-IT~\cite{Gemma_4_2026}, MiniCPM-V~4.5~\cite{MiniCPM_V_4_5_2025}, and the Molmo series~\cite{Molmo_PixMo_2025}.
All deployment and evaluation details follow exactly the same setup as the main experiments described in~\cref{sec:experiments}.

\begin{table*}[t!]
    \centering
    \small
    \caption{
        \textbf{Additional results on ChartArena.}
        We provide additional results across various model groups and parameter scales to ensure transparency and enable broader comparisons.
    }
    \vspace{-0.4em}
    {
        \renewcommand{\arraystretch}{1.1}
        \setlength{\tabcolsep}{4pt}
        \resizebox{\textwidth}{!}{
            \begin{tabular}{clc cc cc cc cc cc cc cc cc cc}
                \toprule
                \multirow{2}{*}[-0.8mm]
                {\textbf{\makecell{Model \nextline Group}}}  &
                \multirow{2}{*}[-0.8mm]
                {\textbf{Model}}                             &
                \multirow{2}{*}[-0.8mm]
                {\textbf{\makecell{Release \nextline Date}}} &
                \multicolumn{2}{c}{\textbf{bar}}             &
                \multicolumn{2}{c}{\textbf{line}}            &
                \multicolumn{2}{c}{\textbf{pie}}             &
                \multicolumn{2}{c}{\textbf{radar}}           &
                \multicolumn{2}{c}{\textbf{box plot}}        &
                \multicolumn{2}{c}{\textbf{comb.}}           &
                \multicolumn{2}{c}{\textbf{flowchart}}       &
                \multicolumn{2}{c}{\textbf{mind map}}        &
                \multicolumn{2}{c}{\textbf{Average}}
                \\
                \cmidrule(lr){4-5}
                \cmidrule(lr){6-7}
                \cmidrule(lr){8-9}
                \cmidrule(lr){10-11}
                \cmidrule(lr){12-13}
                \cmidrule(lr){14-15}
                \cmidrule(lr){16-17}
                \cmidrule(lr){18-19}
                \cmidrule(lr){20-21}
                                                             &
                                                             &
                                                             &
                EN                                           & ZH                                                    &
                EN                                           & ZH                                                    &
                EN                                           & ZH                                                    &
                EN                                           & ZH                                                    &
                EN                                           & ZH                                                    &
                EN                                           & ZH                                                    &
                EN                                           & ZH                                                    &
                EN                                           & ZH                                                    &
                EN                                           & ZH
                \\
                \midrule
                \multirow{4}{*}[-0.8mm]{\makecell{InternVL \nextline 3.5}}
                                                             & \logo{intern}InternVL3.5-1B~\cite{InternVL3_5_2025}   & 2025.08 & \ul{22.9}     & 40.5          & 24.5          & 36.6          & 51.1          & 73.3          & 4.2           & 18.1          & 0.5           & 1.0           & 4.3           & 32.7          & 8.5           & 4.7           & 14.3          & 3.1           & 16.3          & 26.2          \\
                                                             & \logo{intern}InternVL3.5-2B~\cite{InternVL3_5_2025}   & 2025.08 & \textbf{23.1} & \textbf{54.2} & 27.3          & 46.7          & 63.1          & \textbf{79.8} & 7.4           & \textbf{35.3} & \ul{6.2}      & 7.0           & 6.9           & 37.8          & 16.3          & 12.2          & 11.2          & 2.6           & 20.2          & 34.5          \\
                                                             & \logo{intern}InternVL3.5-4B~\cite{InternVL3_5_2025}   & 2025.08 & 20.0          & 50.1          & \ul{31.5}     & \ul{52.5}     & \ul{65.6}     & 66.9          & \ul{11.4}     & 34.3          & \textbf{13.4} & \textbf{10.9} & \textbf{11.4} & \ul{40.0}     & \ul{29.9}     & \textbf{23.8} & \textbf{50.1} & \textbf{33.8} & \ul{29.2}     & \ul{39.0}     \\
                                                             & \logo{intern}InternVL3.5-8B~\cite{InternVL3_5_2025}   & 2025.08 & 22.7          & \ul{52.6}     & \textbf{34.4} & \textbf{53.7} & \textbf{65.8} & \ul{73.8}     & \textbf{14.0} & \ul{34.7}     & 5.6           & \ul{9.5}      & \ul{11.3}     & \textbf{42.1} & \textbf{32.6} & \ul{23.8}     & \ul{48.3}     & \ul{31.8}     & \textbf{29.3} & \textbf{40.2} \\
                \midrule
                \multirow{4}{*}[-0.8mm]{\makecell{Qwen2.5 \nextline VL}}
                                                             & \logo{qwen}Qwen2.5-VL-3B-Ins.~\cite{Qwen2_5_VL_2025}  & 2025.01 & 20.4          & 51.7          & \ul{32.9}     & 50.7          & 62.8          & \ul{78.3}     & 6.0           & 24.9          & 1.6           & 2.7           & \ul{11.7}     & 41.0          & 9.0           & 12.4          & 26.5          & 16.9          & 21.4          & 34.8          \\
                                                             & \logo{qwen}Qwen2.5-VL-7B-Ins.~\cite{Qwen2_5_VL_2025}  & 2025.01 & 15.2          & 36.9          & 17.9          & 39.9          & 63.4          & 73.1          & 8.3           & 19.1          & 0.9           & 2.8           & 6.0           & 40.6          & 29.7          & 23.2          & 45.4          & 29.9          & 23.3          & 33.2          \\
                                                             & \logo{qwen}Qwen2.5-VL-32B-Ins.~\cite{Qwen2_5_VL_2025} & 2025.03 & \textbf{27.4} & \textbf{60.1} & 25.3          & \ul{53.4}     & \ul{72.5}     & \textbf{84.7} & \textbf{14.0} & \textbf{39.2} & \ul{8.9}      & \textbf{16.4} & 10.6          & \ul{47.3}     & \ul{41.7}     & \ul{40.1}     & \ul{54.0}     & \ul{40.6}     & \ul{31.8}     & \ul{47.7}     \\
                                                             & \logo{qwen}Qwen2.5-VL-72B-Ins.~\cite{Qwen2_5_VL_2025} & 2025.01 & \ul{27.1}     & \ul{53.3}     & \textbf{38.2} & \textbf{66.7} & \textbf{73.5} & 77.0          & \ul{10.9}     & \ul{38.5}     & \textbf{15.0} & \ul{15.3}     & \textbf{14.3} & \textbf{50.5} & \textbf{50.1} & \textbf{43.6} & \textbf{63.8} & \textbf{55.0} & \textbf{36.6} & \textbf{50.0} \\
                \midrule
                \multirow{6}{*}[-1.2mm]{\makecell{Qwen3 \nextline VL}}
                                                             & \logo{qwen}Qwen3-VL-2B-Ins.~\cite{Qwen3_VL_2025}      & 2025.10 & 22.1          & 45.9          & 34.1          & 45.2          & 70.4          & 79.1          & 9.2           & 35.3          & 1.9           & 2.3           & 10.8          & 40.0          & 23.3          & 20.1          & 44.1          & 21.0          & 27.0          & 36.1          \\
                                                             & \logo{qwen}Qwen3-VL-2B-Think.~\cite{Qwen3_VL_2025}    & 2025.10 & 21.1          & 45.9          & 29.7          & 48.0          & 69.1          & 65.6          & 8.0           & 32.1          & 3.1           & 1.1           & 7.3           & 39.7          & 30.1          & 19.1          & 43.9          & 18.6          & 26.5          & 33.8          \\
                                                             & \logo{qwen}Qwen3-VL-4B-Ins.~\cite{Qwen3_VL_2025}      & 2025.10 & 25.9          & 52.7          & 26.3          & 57.0          & 75.4          & \textbf{85.5} & 12.3          & 28.5          & 15.3          & 20.0          & 7.8           & \ul{47.1}     & 44.7          & 35.9          & 57.6          & 38.1          & 33.2          & 45.6          \\
                                                             & \logo{qwen}Qwen3-VL-4B-Think.~\cite{Qwen3_VL_2025}    & 2025.10 & 24.1          & 49.8          & 34.3          & 57.9          & 73.5          & 73.7          & 10.4          & 35.5          & \ul{22.1}     & \textbf{29.0} & 11.3          & 44.0          & 39.1          & 29.0          & 62.4          & 35.7          & 34.7          & 44.3          \\
                                                             & \logo{qwen}Qwen3-VL-8B-Ins.~\cite{Qwen3_VL_2025}      & 2025.10 & \ul{27.5}     & \textbf{58.6} & \ul{35.5}     & \ul{61.1}     & \ul{77.3}     & \ul{84.7}     & \textbf{16.8} & \textbf{42.6} & 11.6          & 12.1          & \ul{13.2}     & \textbf{47.9} & \textbf{50.0} & \textbf{41.5} & \textbf{66.4} & \textbf{54.6} & \ul{37.3}     & \textbf{50.4} \\
                                                             & \logo{qwen}Qwen3-VL-8B-Think.~\cite{Qwen3_VL_2025}    & 2025.10 & \textbf{32.1} & \ul{56.5}     & \textbf{42.0} & \textbf{64.9} & \textbf{79.4} & 84.2          & \ul{14.6}     & \ul{39.3}     & \textbf{22.4} & \ul{25.9}     & \textbf{16.0} & 46.2          & \ul{45.7}     & \ul{38.1}     & \ul{64.8}     & \ul{42.3}     & \textbf{39.6} & \ul{49.7}     \\
                \midrule
                \multirow{4}{*}[-0.8mm]{\makecell{Qwen \nextline 3.5}}
                                                             & \logo{qwen}Qwen3.5-0.8B~\cite{Qwen3_5_2026}           & 2026.03 & 16.1          & 26.0          & 9.1           & 21.3          & 35.6          & 49.2          & 3.4           & 19.3          & 0.0           & 0.2           & 1.8           & 10.0          & 6.2           & 4.3           & 5.9           & 0.5           & 9.8           & 16.3          \\
                                                             & \logo{qwen}Qwen3.5-2B~\cite{Qwen3_5_2026}             & 2026.03 & 16.7          & 33.2          & 19.6          & 31.3          & 46.7          & 62.7          & 5.4           & 20.3          & 1.5           & 1.5           & 2.9           & 20.1          & 17.7          & 11.7          & 18.7          & 7.7           & 16.2          & 23.6          \\
                                                             & \logo{qwen}Qwen3.5-4B~\cite{Qwen3_5_2026}             & 2026.03 & \ul{24.2}     & \ul{35.3}     & \ul{35.1}     & \ul{50.1}     & \ul{76.9}     & \textbf{79.6} & \ul{14.1}     & \ul{38.0}     & \ul{9.3}      & \ul{8.1}      & \ul{8.8}      & \ul{46.7}     & \ul{39.3}     & \ul{30.3}     & \ul{60.2}     & \ul{39.6}     & \ul{33.5}     & \ul{41.0}     \\
                                                             & \logo{qwen}Qwen3.5-9B~\cite{Qwen3_5_2026}             & 2026.03 & \textbf{32.5} & \textbf{45.1} & \textbf{45.5} & \textbf{54.1} & \textbf{82.6} & \ul{76.9}     & \textbf{22.0} & \textbf{44.8} & \textbf{15.3} & \textbf{18.1} & \textbf{16.8} & \textbf{49.5} & \textbf{45.5} & \textbf{38.1} & \textbf{64.2} & \textbf{54.5} & \textbf{40.6} & \textbf{47.6} \\
                \midrule
                \multirow{4}{*}[-0.8mm]{\makecell{Other \nextline General \nextline MLLMs}}
                                                             & \logo{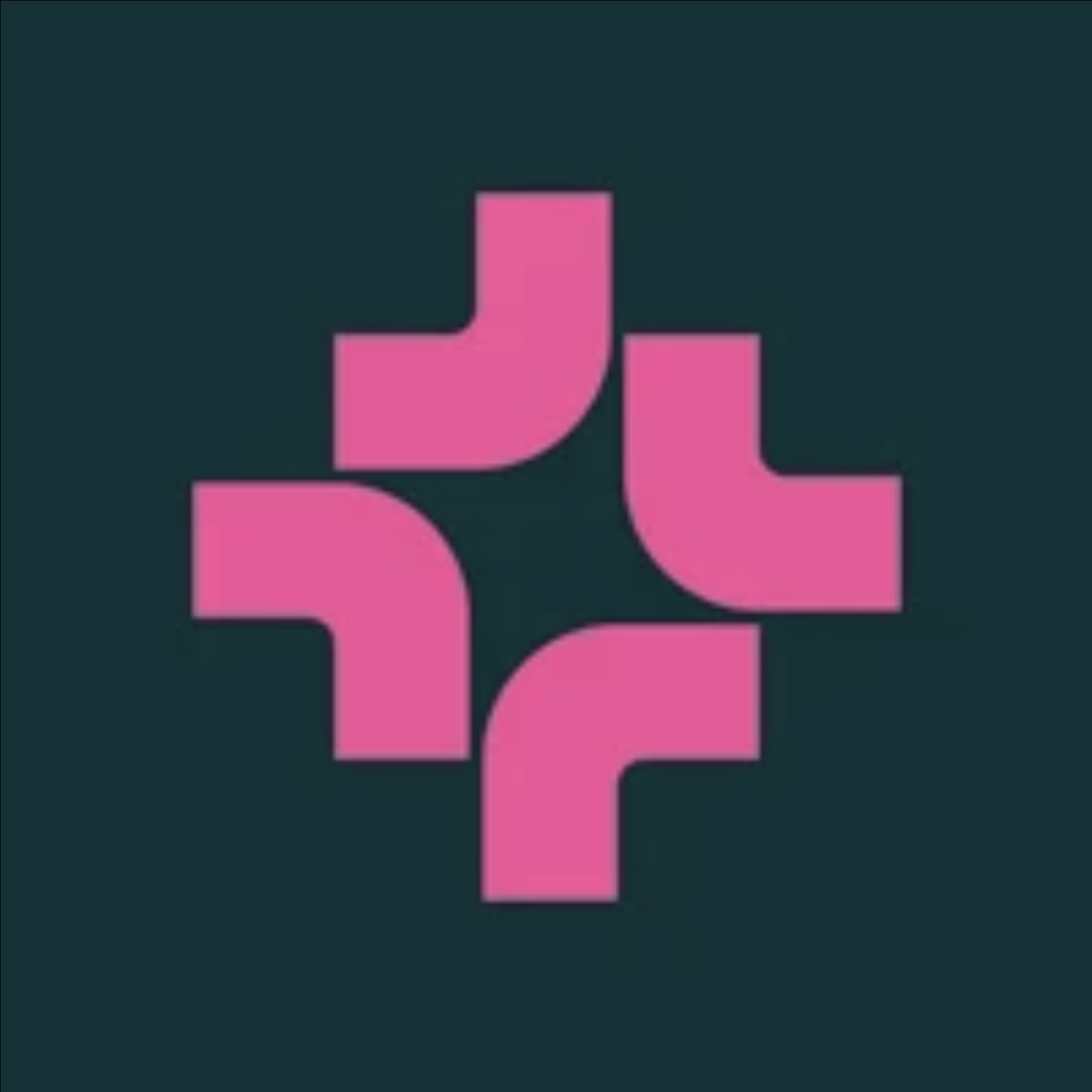}Molmo-7B-D~\cite{Molmo_PixMo_2025}      & 2024.09 & 2.9           & 3.3           & 2.9           & 0.8           & 3.6           & 9.1           & 1.4           & 2.7           & 0.0           & 0.0           & 0.0           & 0.9           & 0.0           & 0.0           & 2.3           & 0.0           & 1.6           & 2.2           \\
                                                             & \logo{AllenAI}Molmo-72B~\cite{Molmo_PixMo_2025}       & 2024.09 & 3.5           & 2.6           & 2.5           & 0.7           & 6.3           & 5.5           & 1.5           & 6.5           & 0.7           & 0.1           & 0.1           & 0.7           & 0.3           & 0.0           & 2.1           & 0.0           & 2.1           & 2.0           \\
                                                             & \logo{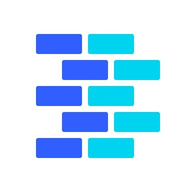}MiniCPM-V~4.5~\cite{MiniCPM_V_4_5_2025} & 2025.08 & \ul{23.3}     & \ul{52.2}     & \ul{34.7}     & \ul{53.5}     & \ul{73.3}     & \ul{79.3}     & \ul{13.0}     & \ul{34.7}     & \textbf{15.3} & \textbf{18.0} & \ul{14.2}     & \ul{41.1}     & \ul{34.3}     & \ul{29.0}     & \ul{57.4}     & \ul{31.2}     & \ul{33.2}     & \ul{42.4}     \\
                                                             & \logo{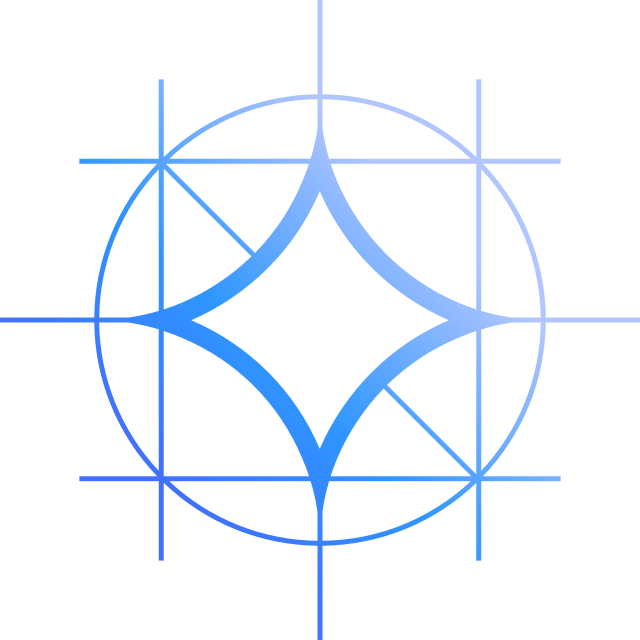}Gemma~4-31B-IT~\cite{Gemma_4_2026}        & 2026.04 & \textbf{43.7} & \textbf{68.7} & \textbf{53.4} & \textbf{76.4} & \textbf{88.9} & \textbf{87.2} & \textbf{20.9} & \textbf{55.5} & \ul{9.3}      & \ul{8.9}      & \textbf{22.7} & \textbf{59.3} & \textbf{59.6} & \textbf{44.0} & \textbf{76.2} & \textbf{61.0} & \textbf{46.8} & \textbf{57.6} \\
                \bottomrule
            \end{tabular}
        }
    }
    \label{tab:ChartArena_additional_results}
\end{table*}

\section{Further Related Work}
\label{supp:sec:further_related_work}

This section extends the discussion in~\cref{sec:related_work} by providing background on large language models, specialized chart parsers, and evaluation conventions in neighboring document parsing tasks.

\subsection{Large Language and Multimodal Models}
\label{supp:subsec:llm_mllm}

The rapid progress of large language models (LLMs)~\citep{Qwen3_2025, GPT_4o_2024, GPT3_2020, Llama_2_2023, Llama_3_2024, Qwen2_5_2024, Gemini_1_5_2024, Claude3_2024, Deepseek_v3_2024, Deepseek_R1_2025} has substantially reshaped the landscape of artificial intelligence. A key driver behind this progress is the combination of large-scale pre-training on web-scale text corpora with subsequent supervised fine-tuning (SFT) and preference alignment~\citep{Uni_DPO_2026, DPO_2023, SimPO_2024, Training_follow_2022}, which together endow LLMs with strong reasoning, instruction-following, and emergent generalization abilities~\cite{GPT3_2020,Deepseek_v3_2024,Deepseek_R1_2025}. These developments have steadily expanded the range of real-world tasks that language models can reliably handle.

Building on this foundation, vision-language models (VLMs) extend such capabilities to the visual domain by aligning image and text representations in a shared semantic space~\citep{CLIP_2021, BLIP2_2023}. More recently, multimodal large language models (MLLMs) integrate a visual encoder with a powerful LLM backbone through cross-modal connectors and visual instruction tuning, achieving strong perception and reasoning over images, documents, and other modalities~\citep{Qwen2_5_VL_2025, HunyuanOCR_2025, Qwen3_VL_2025, InternVL3_5_2025, GLM_4_5V_2025, LLaVA_2023, InstructBLIP_2023, Qwen2_VL_2024}. This convergence of language and vision has turned MLLMs into general-purpose interfaces for visually grounded understanding~\citep{SENTINEL_2025}, and motivates their use as the dominant paradigm for chart parsing studied in this work.

\subsection{Specialized MLLMs for Chart Parsing}
\label{supp:subsec:mllm_chart}

Building on the end-to-end formulation reviewed in~\cref{sec:related_work}, a growing body of specialized chart parsers further adapts general MLLMs to this task, either by instruction tuning on synthetic chart corpora or by introducing chart-specific representations and training objectives~\citep{ChartX_ChartVLM_2025, OneChart_2024, TinyChart_2024, ChartAssistant_2024, MSRL_2025, RRVF_2025, ChartCoder_2025, ChartMoE_2024, ChartLlama_2023}. Despite these advances, most parsers still learn a direct pixel-to-string mapping and emit results in idiosyncratic output formats, which complicates fair comparison and leaves diagrammatic structures such as flowcharts and mind maps largely underexplored. These observations directly motivate the unified benchmark and format-agnostic protocol introduced in this work.

\subsection{Evaluation for Document and Chart Parsing}
\label{supp:subsec:eval_doc_chart}

Standardized evaluation has been a major catalyst of progress in structured document parsing. Closely related tasks have largely converged toward unified evaluation conventions, including table parsing~\cite{shen2023divide, data_fintabnet, data_pubtabnet, StrucTab_2026, CC_OCR_2025} and mathematical formula parsing~\citep{cdm_eval, data_hme, data_unimernet}, where shared output formats and metrics enable direct cross-model comparison. Beyond contemporary documents, parsing historical and ancient materials introduces additional challenges from degraded media, archaic characters, and evolving character forms, and has correspondingly motivated dedicated benchmarks~\citep{Chronicles_OCR_2026, M5HisDoc_2023}.

Chart parsing, by contrast, remains comparatively fragmented, as summarized in~\cref{tab:chart_benchmark_comparison}. Along the chart-type axis, coverage has expanded only incrementally over time. Early benchmarks such as PlotQA-SE~\citep{PlotQA_2020} and ChartQA-SE~\citep{ChartQA_2022} focus on the three most common numeric types (bar, line, and pie), and subsequent efforts add only a few more, such as radar in MMC-Bench~\citep{MMC_Bench_2024} and ExChart-Bench~\citep{ExChart_Bench_2026}, box plots in ChartX-SE~\citep{ChartX_ChartVLM_2025} and VG-DCU~\citep{VG_DCU_2024}, or combination charts in ChartY~\citep{OneChart_2024} and ParseBench~\citep{ParseBench_2026}. Crucially, none of these benchmarks include diagrammatic charts such as flowcharts and mind maps, even though such structures are pervasive in real documents and demand explicit topological reasoning beyond value extraction. The visual-style and language axes are equally narrow: almost all existing benchmarks consist solely of clean digital renderings and overlook real-world conditions such as printed or hand-drawn photos, and only ChartY~\citep{OneChart_2024} offers bilingual content while the rest are English-only. Compounded by inconsistent output formats across methods, these gaps make it difficult to assess chart parsers fairly or to evaluate numeric and diagrammatic charts within a single framework. As shown in~\cref{tab:chart_benchmark_comparison}, ChartArena is the first benchmark to jointly cover all eight chart families, three visual scenarios, and both languages, and is explicitly designed to close this gap under one unified, format-agnostic evaluation protocol.

\section{Broader Impact}
\label{supp:sec:broader_impact}

This work contributes to the advancement of general chart parsing and evaluation. By introducing a unified benchmark and evaluation protocol, we aim to support more reliable assessment of multimodal models on structured visual reasoning tasks. While chart understanding may be applied in domains such as scientific analysis, business intelligence, and document accessibility, we have not identified any broader societal impacts that warrant particular concern at this time.

\section{LLM Usage Statement}
\label{supp:sec:llm_usage}

LLMs and MLLMs play two distinct roles in this work, and we disclose both.

\mypara{As part of the method.}
Model usage that forms part of the research is described in the corresponding method sections. In benchmark construction, multiple MLLMs generate candidate annotations that are merged into a draft and then verified and corrected by human annotators (\cref{subsec:benchmark_construction} and~\cref{supp:subsec:annotation_protocol}); every released label is human-verified. In training data construction, Qwen3-235B-A22B-Instruct synthesizes the semantic corpora, chart templates, and diagrammatic targets of the synthetic pretraining data (\cref{subsec:synthesis_pipeline}), and Qwen3-VL-235B-A22B-Instruct extracts the atomic primitives from which the CAP reasoning traces are built (\cref{subsec:cap_reasoning}); a human audit of these targets is reported in~\cref{supp:subsec:cap_verification}. All MLLMs evaluated on ChartArena are listed with their inference setup in~\cref{supp:subsec:evaluation_setup}.

\mypara{In manuscript preparation.}
LLMs were used only as auxiliary writing tools. Their role was limited to improving language quality, including grammar correction and wording refinement.

\section{Further Case Study}
\label{supp:sec:further_case_study}

\begin{figure}[h!]
    \centering
    \includegraphics[width=\linewidth]{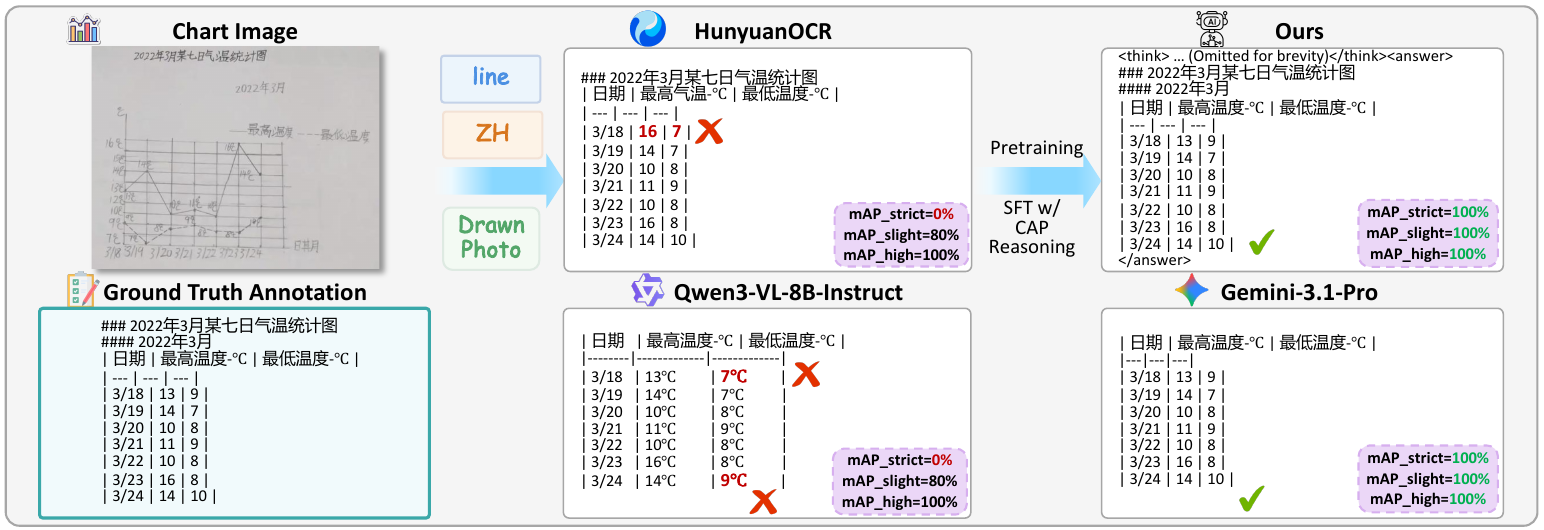}
    \vspace{-1.4em}
    \caption{
        \textbf{Case on hand-drawn line chart.}
        CAP reasoning improves robustness under challenging visual conditions, enabling more faithful numerical recovery.
    }
    \label{fig:supp_case_1}
\end{figure}

\begin{figure}[h!]
    \centering
    \includegraphics[width=\linewidth]{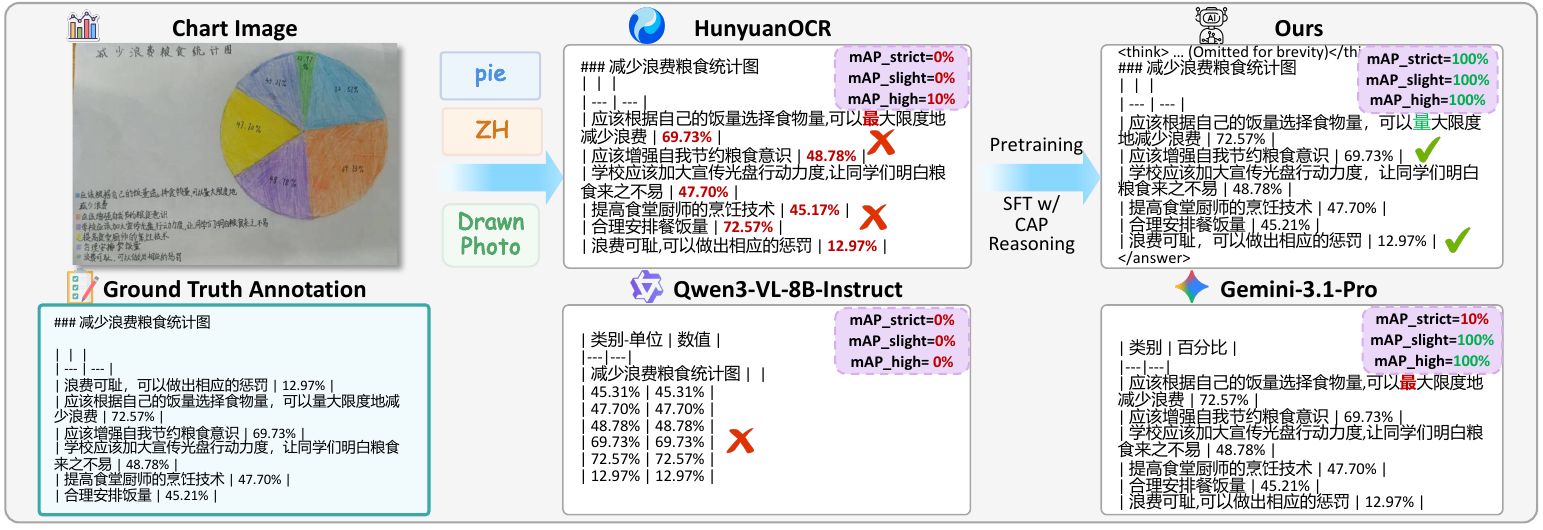}
    \vspace{-1.4em}
    \caption{
        \textbf{Case on hand-drawn pie chart.}
        CAP reasoning improves robustness under challenging visual conditions, enabling more faithful numerical recovery. Notably, the strong baseline Gemini 3.1 Pro~\cite{Gemini_3_1_2026} suffers from hallucination, incorrectly recognizing a visually present character as a more plausible alternative, which subsequently leads to parsing errors.
    }
    \label{fig:supp_case_2}
\end{figure}

We provide additional qualitative comparisons in~\cref{fig:supp_case_1,fig:supp_case_2} to further analyze the behavior of different models under challenging visual scenarios. The examples include hand-drawn photos, introducing substantial visual uncertainty for chart parsing.

\end{document}